\documentclass{article}
\usepackage{iclr2027_conference,times}

\newif\ifdraft
\draftfalse   %

\newif\ifpreprint
\preprinttrue

\usepackage{amsmath,amsfonts,bm}

\def\eqref#1{equation~\ref{#1}}

\def\1{\bm{1}}

\DeclareMathAlphabet{\mathsfit}{\encodingdefault}{\sfdefault}{m}{sl}
\SetMathAlphabet{\mathsfit}{bold}{\encodingdefault}{\sfdefault}{bx}{n}

\usepackage[T1]{fontenc}

\usepackage{microtype}

\usepackage{enumitem}
\setlist[enumerate]{topsep=2pt, itemsep=2pt, parsep=0pt, partopsep=0pt, leftmargin=*}
\setlist[itemize]{topsep=2pt, itemsep=2pt, parsep=0pt, partopsep=0pt, leftmargin=*}

\usepackage{amsmath}
\usepackage{amssymb}
\usepackage{mathtools}

\usepackage{graphicx}
\usepackage{booktabs}
\usepackage{dashrule}
\newcommand{\dashedmidrule}{\noalign{\vskip\aboverulesep\hbox{\hdashrule[0.5ex]{\linewidth}{0.4pt}{2.5pt 2pt}}\vskip\belowrulesep}}
\usepackage{threeparttable}
\usepackage{multirow}
\usepackage{array}
\usepackage{rotating}   %
\usepackage{longtable}  %
\usepackage{pdflscape}  %
\usepackage{placeins}   %

\usepackage{subcaption} %

\usepackage{xcolor}
\usepackage{listings}

\usepackage{tikz}
\definecolor{f1gray}{HTML}{666666}
\definecolor{f1graylight}{HTML}{A8A8A8}
\definecolor{methodsingle}{HTML}{E69F00}
\definecolor{methodaveraged}{HTML}{009E73}
\definecolor{methoddebias}{HTML}{CC79A7}
\definecolor{methodperm}{HTML}{56B4E9}
\definecolor{f1blue}{HTML}{0072B2}
\definecolor{f1churn}{HTML}{D55E00}
\definecolor{f1signal}{HTML}{009E73}

\tikzset{figbase/.style  = {x=1pt, y=1pt, font=\sffamily\fontsize{8}{9.6}\selectfont}}
\tikzset{figbase7/.style = {x=1pt, y=1pt, font=\sffamily\fontsize{7}{8}\selectfont}}

\tikzset{
  figtitle/.style   = {font=\sffamily\fontsize{8}{9.6}\selectfont\bfseries, color=black!85},
  figtitle7/.style  = {font=\sffamily\fontsize{7}{8}\selectfont\bfseries,   color=black!85},
  figblock/.style   = {font=\sffamily\fontsize{8}{9.6}\selectfont\bfseries, color=black!50},
  figblock7/.style  = {font=\sffamily\fontsize{7}{8}\selectfont\bfseries,   color=black!50},
  figop/.style      = {font=\sffamily\fontsize{11}{11}\selectfont, color=black!85},
  figop7/.style     = {font=\sffamily\fontsize{10}{10}\selectfont, color=black!85},
  figmetric/.style  = {color=black!50},
  figlbl/.style     = {color=black!85},
  figtick/.style    = {color=black!50},
  figaxislbl/.style = {color=black!50},
  figonmark/.style  = {color=white},
  figflow/.style    = {line width=0.4pt, color=black!50},
  figrule/.style    = {line width=0.3pt, color=black!25},
  figgrid/.style    = {line width=0.3pt, color=black!12},
  figzero/.style    = {line width=0.5pt, color=black!25},
  fighair/.style    = {line width=0.4pt, color=black!25},
  figbox/.style     = {draw=black!25, line width=0.3pt, fill=white, rounded corners=0.8pt},
  figseries/.style     = {line width=1.0pt, line cap=round, color=f1gray},
  figours/.style       = {line width=1.4pt, line cap=round, color=f1blue},
  figmark/.style       = {fill=f1gray, draw=white, line width=0.5pt},
  figmarkours/.style   = {fill=f1blue, draw=white, line width=0.5pt},
  figmarkopen/.style   = {fill=white, draw=f1gray, line width=0.5pt},
  figband/.style       = {fill=f1gray, fill opacity=0.30, draw=none},
  figbandours/.style   = {fill=f1blue, fill opacity=0.30, draw=none},
  figcell/.style       = {fill=f1gray, fill opacity=0.55, draw=none},
}

\tikzset{
  figtxt/.style      = {inner sep=0pt, outer sep=0pt},
  figlblours/.style  = {color=f1blue},
  fighalo/.style     = {line width=2.4pt, line cap=round, color=white},
  figbandcap/.style  = {line width=1.0pt, line cap=butt, color=f1gray},
  figbandcapours/.style = {line width=1.0pt, line cap=butt, color=f1blue},
  figmarkrule/.style = {line width=0.4pt, color=f1gray},
  figdashrule/.style = {line width=0.3pt, color=black!25, dash pattern=on 0.6pt off 1.1pt},
  figghost/.style    = {line width=0.3pt, color=black!12, dash pattern=on 0.4pt off 0.8pt},
  figemph/.style     = {line width=1.0pt, color=black!85},
  figboxmark/.style  = {draw=black!25, line width=0.3pt, fill=f1gray, rounded corners=0.8pt},
  figboxmarkours/.style = {draw=black!25, line width=0.3pt, fill=f1blue, rounded corners=0.8pt},
  figlevel/.style    = {line width=0.9pt, color=f1blue!55, dash pattern=on 1.7pt off 1.5pt},
}

\newcommand{\figmetrictext}[1]{{\color{black!50}#1}}

\usepackage{url}
\usepackage{hyperref}
\hypersetup{colorlinks=true, linkcolor=blue!55!black, citecolor=blue!55!black,
            urlcolor=blue!55!black, bookmarksnumbered=true}
\makeatletter
\AtBeginDocument{\hypersetup{pdftitle={\@title}}}
\makeatother

\ifdraft
  \usepackage[colorinlistoftodos, textsize=footnotesize]{todonotes}
\else
  \usepackage[disable]{todonotes}
\fi

\usepackage[capitalise]{cleveref}
\input{cleveref-aux-guard.tex}
\crefname{appendix}{App.}{Apps.}
\Crefname{appendix}{App.}{Apps.}
\crefalias{subsection}{section}
\crefformat{section}{\S#2#1#3}
\Crefformat{section}{\S#2#1#3}
\crefrangeformat{section}{\S\S#3#1#4--#5#2#6}
\Crefrangeformat{section}{\S\S#3#1#4--#5#2#6}
\crefmultiformat{section}{\S\S#2#1#3}{ and~#2#1#3}{, #2#1#3}{, and~#2#1#3}
\Crefmultiformat{section}{\S\S#2#1#3}{ and~#2#1#3}{, #2#1#3}{, and~#2#1#3}
\crefrangemultiformat{section}{\S\S#3#1#4--#5#2#6}{ and~#3#1#4--#5#2#6}{, #3#1#4--#5#2#6}{, and~#3#1#4--#5#2#6}
\Crefrangemultiformat{section}{\S\S#3#1#4--#5#2#6}{ and~#3#1#4--#5#2#6}{, #3#1#4--#5#2#6}{, and~#3#1#4--#5#2#6}

\lstdefinestyle{prompt}{%
  basicstyle=\ttfamily\small,
  breaklines=true,
  frame=single,
  framesep=6pt,
  columns=fullflexible,
  keepspaces=true,
  showstringspaces=false,
  aboveskip=8pt, belowskip=8pt,
}
\newcommand{\rothead}[1]{\rotatebox[origin=l]{90}{#1}}

\title{Equal Ranking Quality, Different Decisions: Measuring and Reducing Order Dependence in LLM Scorers}

\ifpreprint\iclrfinalcopy\fi

\ifpreprint
\newcommand{\codeurl}{\href{https://github.com/thomsonreuters/presentation-dependence}%
  {\nolinkurl{github.com/thomsonreuters/presentation-dependence}}}
\newcommand{\codeurlfull}{\href{https://github.com/thomsonreuters/presentation-dependence}%
  {\nolinkurl{github.com/thomsonreuters/presentation-dependence}}}
\else
\newcommand{\codeurl}{\href{https://anonymous.4open.science/r/order-dependence-llm-scorers}%
  {\nolinkurl{anonymous.4open.science}}}
\newcommand{\codeurlfull}{\href{https://anonymous.4open.science/r/order-dependence-llm-scorers}%
  {\nolinkurl{anonymous.4open.science/r/order-dependence-llm-scorers}}}
\fi

\ifpreprint
\author{%
  Markus Frohmann$^{1,2}$ \quad Mahdiyar Ali Akbar Alavi$^{1}$ \quad
  Elizabeth Lingg$^{1}$ \quad Navid Rekabsaz$^{1}$\thanks{The contribution of the
  author was made during his work at Thomson Reuters.} \\[4pt]
  $^{1}$Thomson Reuters Labs \quad
  $^{2}$University of Toronto, Vector Institute \\[3pt]
  Correspondence: \texttt{\small markus.frohmann@mail.utoronto.ca} \\
  \texttt{\small\{mahdiyar.alavi, elizabeth.lingg, navid.rekabsaz\}@thomsonreuters.com}
}
\hypersetup{pdfauthor={Markus Frohmann, Mahdiyar Alavi, Elizabeth Lingg, Navid Rekabsaz}}
\else
\author{Anonymous authors\\Paper under double-blind review}
\fi

\begin{document}
\ifpreprint\vspace*{-18pt}\fi
\maketitle
\ifpreprint\lhead{\ifnum\value{page}=1\relax\footnotesize\scshape Preprint\fi}\fi
\ifpreprint\vspace{-20pt}\else\vspace{-10pt}\fi

\begin{abstract}
    In passage reranking, response ranking and multi-document question answering, LLMs can score several candidate documents or responses together in one prompt, each still receiving its own score.
    Such scorers are selected on ranking quality, but their scores determine a decision: what a score threshold retains, a reader answers, or which chosen/rejected pair enters preference training.
    Because the candidates share that prompt, reordering them changes their scores. The same query over the same candidates should still yield the same decision.
    However, equal ranking quality does not imply equal decisions: on passage reranking, five trained scorers within 0.010 nDCG@10 retain sets that overlap by only 0.66--0.84 when reordered.
    No prompt-time change we test resolves that dependence: the only one that improves ranking quality does not measurably improve decision stability.
    We introduce order-consistency SFT (OC-SFT), which attenuates it in the weights by penalizing disagreement between a candidate's scores across orderings.
    It holds ranking quality and leads every decision-stability measure among trained scorers on all three tasks.
    It is also more stable on 12 base models than order-averaged distillation, which trains on labels averaged across permutations.
    One OC-SFT permutation retains sets that overlap more than ten averaged off-the-shelf permutations.
    A comparison of such scorers should therefore report what a threshold retains and a reader answers, not ranking quality alone.
    Code is available at \codeurl.\looseness=-1
\end{abstract}
\ifpreprint\vspace{-6pt}\else\vspace{-4pt}\fi
\section{Introduction}
\label{sec:intro}
\ifpreprint\vspace{-4pt}\else\vspace{-3pt}\fi

A language model that scores candidate documents or responses is rarely the last component of the system it serves.
A threshold admits documents above a cutoff into a review queue or generation context, a reader answers from the highest-scoring few, and a preference model's scores determine a chosen/rejected response pair used to train another model \citep{cui2024}.
These three consumers act on the scores, so the output is a decision rather than a ranking,
and reordering the same candidates for the same query should not change it.
Ranking quality is the property such scorers are selected on, and two that match on it can still differ in every one of those three decisions.

Such a scorer must choose how many candidates share a prompt, and the standard designs each forgo either the per-candidate score or the shared context.
Pointwise cross-encoders return a per-candidate value that a threshold can consume, but score one candidate per forward pass (monoT5, \citealp{nogueira2020}; RankLLaMA, \citealp{ma2023}).
Generative listwise rerankers condition on several candidates jointly, but emit a ranked list and no per-candidate score (RankGPT, \citealp{sun2023}; RankZephyr, \citealp{pradeep2023}).
Batched pointwise scoring, which we study, retains both \citep{korikov2025}, and its shared prompt also lowers per-query latency, since the instruction, query and grading rubric are encoded once per call rather than once per candidate.

Because the candidates occupy a shared prompt, the score assigned to one is conditioned on the entire prompt rather than on that candidate alone. We term this \emph{presentation dependence}.
Language models (LMs) also attend unevenly across long inputs \citep{liu2024}, which induces sensitivity to candidate order wherever candidates are scored jointly \citep{zeng2025, hou2024}.
Off the shelf, the order decides whether a preference model's top choice is right on 65.4\% of queries across ten random permutations, so it also decides which response becomes a training label (\Cref{app:consumerk}).
However, order is not the only such property: with order held fixed, a candidate's score still varies with which other candidates share its prompt and with how the prompt is worded (\Cref{sec:phenomenon}).

\begin{figure}[t]
\ifpreprint\vspace*{-24pt}\fi
\centering
\begin{tikzpicture}[figbase,
  x=1cm, y=1cm,
  panelhead/.style={figtxt, figtitle, anchor=base west},
  quiet/.style={figtxt, figtick, anchor=base west},
  quietc/.style={figtxt, figtick, anchor=base},
  quietr/.style={figtxt, figtick, anchor=base east},
  rowlbl/.style={figtxt, figlbl, anchor=base west},
  value/.style={figtxt, figlbl, anchor=base},
  operator/.style={figtxt, figop, anchor=base},
  boxlbl/.style={figtxt, figmetric, anchor=base},
  serieslbl/.style={figtxt, figlbl, anchor=base west},
  boxline/.style={figrule, rounded corners=1.2pt},
  chipemph/.style={figemph, rounded corners=0.5pt},
  chipghost/.style={figghost, rounded corners=0.5pt},
  bandfill/.style={figband, line join=round},
  spanval/.style={figtxt, figlbl, anchor=base},
  collbl/.style={figtxt, figtick, anchor=base},
]
  \useasboundingbox (0.0000,0.6000) rectangle (13.9178,4.8400);
  \node[panelhead] at (0.0600,4.6000) {(a) One query, two permutations};
  \node[quiet] at (0.0600,4.2800) {off-the-shelf Qwen3-4B, Climate-FEVER};
  \node[quietc] at (4.2600,3.9000) {the same 100 candidates, one query};
  \draw[boxline] (1.5800,3.28) rectangle (3.5600,3.74);
  \node[boxlbl] at (2.5700,3.4200) {permutation A};
  \draw[figrule] (2.5700,3.22) -- (2.5700,3.0800);
  \fill[figrule] (2.4950,3.0900) -- (2.6450,3.0900) -- (2.5700,2.9800) -- cycle;
  \draw[boxline] (4.9600,3.28) rectangle (6.9400,3.74);
  \node[boxlbl] at (5.9500,3.4200) {permutation B};
  \draw[figrule] (5.9500,3.22) -- (5.9500,3.0800);
  \fill[figrule] (5.8750,3.0900) -- (6.0250,3.0900) -- (5.9500,2.9800) -- cycle;
  \draw[figrule] (0.0600,2.90) -- (7.3628,2.90);
  \draw[figrule] (0.0600,2.36) -- (7.3628,2.36);
  \draw[figrule] (0.0600,1.54) -- (7.3628,1.54);
  \node[rowlbl] at (0.0600,2.6000) {nDCG@10};
  \node[value] at (2.5700,2.6000) {0.6131};
  \node[operator] at (4.2600,2.6000) {$=$};
  \node[value] at (5.9500,2.6000) {0.6131};
  \node[rowlbl] at (0.0600,2.0300) {retained set};
  \fill[figband] (1.6697,1.9829) rectangle (1.9227,2.2571);
  \fill[figband] (2.0071,1.9829) rectangle (2.2601,2.2571);
  \fill[figband] (2.3445,1.9829) rectangle (2.5975,2.2571);
  \draw[chipemph] (2.3445,1.9829) rectangle (2.5975,2.2571);
  \fill[figband] (2.6819,1.9829) rectangle (2.9349,2.2571);
  \fill[figband] (3.0193,1.9829) rectangle (3.2723,2.2571);
  \draw[chipemph] (3.0193,1.9829) rectangle (3.2723,2.2571);
  \draw[chipghost] (3.3567,1.9829) rectangle (3.6097,2.2571);
  \draw[chipghost] (3.6941,1.9829) rectangle (3.9471,2.2571);
  \fill[figband] (5.0497,1.9829) rectangle (5.3027,2.2571);
  \fill[figband] (5.3871,1.9829) rectangle (5.6401,2.2571);
  \draw[chipghost] (5.7245,1.9829) rectangle (5.9775,2.2571);
  \fill[figband] (6.0619,1.9829) rectangle (6.3149,2.2571);
  \draw[chipghost] (6.3993,1.9829) rectangle (6.6523,2.2571);
  \fill[figband] (6.7367,1.9829) rectangle (6.9897,2.2571);
  \draw[chipemph] (6.7367,1.9829) rectangle (6.9897,2.2571);
  \fill[figband] (7.0741,1.9829) rectangle (7.3271,2.2571);
  \draw[chipemph] (7.0741,1.9829) rectangle (7.3271,2.2571);
  \node[quietc] at (2.8084,1.7200) {5 of 100};
  \node[operator] at (4.2600,2.0100) {$\neq$};
  \node[quietc] at (6.1884,1.7200) {5 of 100};
  \node[rowlbl] at (0.0600,1.2400) {reader};
  \node[value] at (2.5700,1.2400) {SUPPORTED};
  \node[operator] at (4.2600,1.2400) {$\neq$};
  \node[value] at (5.9500,1.2400) {NOT ENOUGH INFO};
  \node[panelhead] at (7.9046,4.6000) {(b) Equal quality, different retained sets};
  \node[quiet] at (7.9046,4.2800) {18 collections, $M{=}10$ permutations};
  \draw[figrule] (8.6169,1.28) -- (8.6169,3.90);
  \draw[figrule] (8.5569,1.4509) -- (8.6169,1.4509);
  \node[quietr] at (8.5572,1.3609) {0.45};
  \draw[figrule] (8.5569,2.0204) -- (8.6169,2.0204);
  \node[quietr] at (8.5572,1.9304) {0.55};
  \draw[figrule] (8.5569,2.5900) -- (8.6169,2.5900);
  \node[quietr] at (8.5572,2.5000) {0.65};
  \draw[figrule] (8.5569,3.1596) -- (8.6169,3.1596);
  \node[quietr] at (8.5572,3.0696) {0.75};
  \draw[figrule] (8.5569,3.7291) -- (8.6169,3.7291);
  \node[quietr] at (8.5572,3.6391) {0.85};
  \draw[bandfill] (9.0387,1.4737) -- (11.3888,3.1197) -- (11.3888,3.2165) -- (9.0387,1.4793) -- cycle;
  \draw[figbandcap] (11.3888,3.1197) -- (11.3888,3.2165);
  \draw[figseries] (9.0387,1.4452) -- (11.3888,2.6242);
  \draw[fighalo] (9.0387,1.5021) -- (11.3888,3.6437);
  \draw[figours] (9.0387,1.5021) -- (11.3888,3.6437);
  \draw[figmarkopen] (9.0387,1.4452) circle (1.20pt);
  \draw[figmarkopen] (11.3888,2.6242) circle (1.20pt);
  \draw[figmarkours] (9.0387,1.5021) circle (1.70pt);
  \draw[figmarkours] (11.3888,3.6437) circle (1.70pt);
  \draw[figmarkrule] (11.3138,3.1197) -- (11.4638,3.1197);
  \draw[figmarkrule] (11.3138,3.2108) -- (11.4638,3.2108);
  \draw[figmarkrule] (11.3138,3.2165) -- (11.4638,3.2165);
  \draw[figrule] (11.5388,2.6242) -- (11.5388,3.6437);
  \draw[figrule] (11.5388,2.6242) -- (11.4788,2.6242);
  \draw[figrule] (11.5388,3.6437) -- (11.4788,3.6437);
  \node[collbl] at (9.0387,1.0600) {nDCG@10};
  \node[collbl] at (11.3888,1.0600) {retained-set overlap};
  \node[spanval] at (9.0387,0.7226) {span 0.010};
  \node[spanval] at (11.3888,0.7226) {span 0.179};
  \node[serieslbl,figlblours] at (11.7086,3.5537) {OC-SFT};
  \node[serieslbl] at (11.7086,3.0781) {3 mitigations};
  \node[serieslbl] at (11.7086,2.5342) {Single-order};
\end{tikzpicture}%
\caption{\textbf{The same candidates reordered give the same nDCG@10 but a different decision.} 
(a) Each chip is one document, in the same position in both rows rather than by rank: solid where retained, dotted where not, outlined where they disagree.
(b) Five trained scorers; values in \Cref{tab:levels}.
} 
\label{fig:phenomenon}
\end{figure}
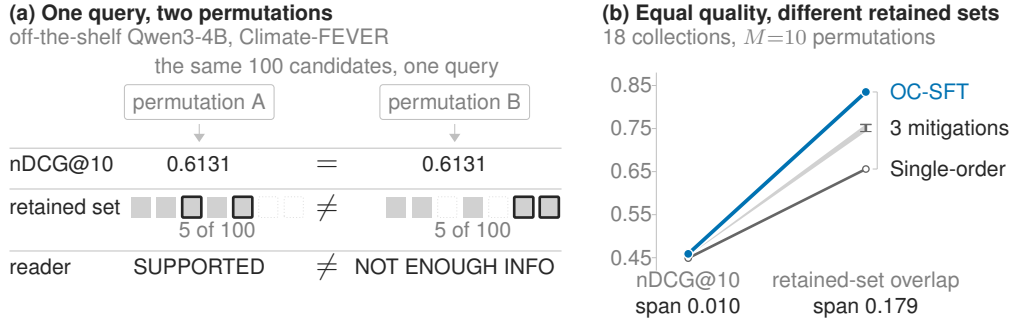

None of the prompt-construction changes we test removes this dependence on order, and the established remedy is to score the same pool under several permutations and average \citep{korikov2025}, which multiplies the serving cost by the number of permutations.
Nor does fixing the order remove the underlying dependence: it selects one presentation-dependent output, and first-stage order is not consistently best (\Cref{app:mechanism}).
Order stability is already reported beside ranking quality for rerankers \citep{bito2026} and beside accuracy for LLM judges \citep{norman2026}, but its effect on the decisions is not.
Reordering leaves both the query and the candidate set unchanged, yet each of the three consumers above can produce a different output, and we measure how often they do.
\Cref{fig:phenomenon} shows one query where they do, and that ranking quality does not register it: nDCG@10 reads only the order of the top ten, while a threshold reads the absolute scores.

The question is whether training can reduce what prompt construction cannot.
Because a threshold consumes a per-candidate score, we fix how scores are read and vary only how the scorer is trained.
Of the objectives we compare (\Cref{sec:method}), the one whose decisions change least is our order-consistency supervised fine-tuning (OC-SFT),
which penalizes the scorer's own disagreement across permutations.

We evaluate on three tasks that score a candidate set in one prompt.
Passage reranking is the primary setting across 18 collections and 12 bases in three families.
Multi-document question answering (QA) and response ranking test whether the dependence is a property of the shared prompt or of reranking alone (\Cref{sec:setup}).
What the shared context adds in quality differs by task and by collection: it helps on QA, not on passage relevance, and on some response-ranking collections (\Cref{sec:copresence}).

\ifpreprint\vspace{-0pt}\else\vspace{-6pt}\fi

We make the following contributions:

\begin{enumerate}
\item We measure what the consumers decide rather than how the ranking moves, and find that equal ranking quality does not imply equal decisions. Five trained scorers within 0.010 nDCG@10 span 0.66--0.84 in retained-set overlap under reshuffling. A comparison should therefore report decision stability, which a stability metric's permutations are enough to measure (\Cref{sec:buys}).
\item Fixes applied before training do not improve the decisions. Round-robin partitioning gains 0.052 nDCG@10, while logit calibration improves neither quality nor retained-set overlap measurably. Neither fix reaches any of the three consumers, so we train the score not to depend on order (\Cref{sec:threetask}).
\item Training improves decision stability without inference-time averaging. At similar ranking quality, OC-SFT is the most stable trained objective on all three tasks; its reader-answer flip rate is 0.125, versus 0.149--0.164 for three other order-aware objectives (\Cref{sec:instrument}).
\end{enumerate}   
\ifpreprint\vspace{-6pt}\else\vspace{-4pt}\fi
\section{Related work}
\label{sec:related}
\ifpreprint\vspace{-4pt}\else\vspace{-4pt}\fi

A candidate's score in a shared prompt depends on its position and on which other candidates accompany it.
Long-context models attend unevenly across their input, which produces the U-shaped ``lost in the middle'' effect \citep{liu2024}.
\citet{zeng2025} measured position sensitivity across retrieval collections and introduced the index we adapt to the rank level (\Cref{sec:phenomenon}).
The composition of the prompt exerts a comparable effect, since a candidate's estimated relevance shifts when the accompanying candidates change \citep{huang2026contextual}.
Both channels are documented in reward models and LLM judges that score responses side by side \citep{malik2025, frick2024, liu2025rmbench}.
Swapping the order alone can flip a two-candidate verdict \citep{wang2024fair}, and the flip rate tracks the quality gap between the two \citep{shi2025judging}.

Two further channels originate in the construction of the prompt rather than in the candidate set.
Partitioning a rank-sorted pool across several prompts renders scores from different prompts not directly comparable, which partitioned listwise reranking addresses by fusing or recalibrating across windows \citep{parry2024, ren2025}.
Where scores are extracted from a fixed answer skeleton, its wording is a further design choice, and no position-bias measurement we know of controls for it (\Cref{sec:threetask}).
Attempts to remove this dependence differ in which stage bears the cost.

\textbf{Inference-time fixes} require no training:
batched self-consistency averages per-document scores over permutations \citep{korikov2025}, extending permutation self-consistency \citep{tang2024, wang2022} and its pairwise form in judging \citep{wang2024fair}.
TourRank ensembles tournament groupings \citep{chen2025tourrank}, and \citet{huang2026contextual} sample contexts adaptively.
Stable-RAG decodes from the center of several retrieval orders downstream of the ranker \citep{stablerag2026}, while CapCal \citep{lv2026} and CalibraEval \citep{lihaitao2025} calibrate instead.
Resampling repeats its cost per query; calibration pays once, in the data that fits the correction.
\textbf{Trained fixes} incur it once:
DebiasFirst \citep{qiao2026} combines position-aware augmentation with inverse-propensity weighting, RankVicuna \citep{pradeep2023vicuna} distills from shuffled orders, \citet{brown2025} fine-tune on the order-stripped inputs of \citet{mcilroyyoung2024setbased}, and \citet{zheng2026} reward consistent decisions with reinforcement learning.
The cost is supervision and one training run, and the outcome is attenuation, not a guarantee \citep{bito2026}. %
Penalizing disagreement between views of the same input is a standard consistency-regularization device of semi-supervised learning \citep{tarvainen2017, liang2021}, and \citet{xiang2024} and \citet{zhang2026} apply it across permuted demonstrations and across prompt phrasings.
Both stabilize a discrete answer choice, which no threshold consumes, whereas we stabilize a continuous per-document score at the cutoff a threshold applies to it, and we measure what that threshold then retains (\Cref{sec:buys}).
\textbf{Architectural fixes} remove the dependence by construction:
SetRank \citep{pang2020} encodes the candidates as a set rather than prompting an LM, keeping the shared context, while LLM-side variants restrict cross-candidate attention to obtain order invariance in a single pass \citep{wang2025pine, mcilroyyoung2024setbased, bito2026}.
The cost falls on the mechanism itself, since that restriction reduces the cross-candidate information the shared prompt was adopted to supply.

Outside these families, rerankers are trained for quality rather than for presentation dependence, on large hard-negative-mined corpora \citep{pradeep2023, wang2025jina, li2026}, and reward models for response selection on large preference corpora \citep{liu2025skywork, jiang2023}.

We attenuate the dependence in training, which leaves the serving pass at one permutation and the shared prompt unrestricted.
Stability is reported beside accuracy for judges \citep{zheng2023judge, shi2025judging, norman2026} and beside ranking quality for reranking \citep{bito2026}, in each case on the scorer's own output.
We measure it instead on what the consumers of that output produce, and as a change across permutations rather than a value under a single permutation (\Cref{sec:buys}).

\ifpreprint\vspace{-8pt}\else\vspace{0pt}\fi
\section{Presentation dependence and how we measure it}
\ifpreprint\vspace{-4pt}\else\vspace{0pt}\fi
\label{sec:phenomenon}

A scorer that reads several candidates in one prompt assigns each of them a score that depends on the rest of the set.
We score a pool of candidates in windows of $B$ per prompt, partitioning the pool where it exceeds $B$, and sort it by the resulting scores.
In reranking, a first stage's top $100$ becomes five contiguous windows of $B=20$, the window size of listwise rerankers (RankGPT, \citealp{sun2023}; RankZephyr, \citealp{pradeep2023}).
A candidate's score depends on its window assignment, on which others share that window, and on its position among them (\Cref{app:mechanism}).
Reshuffling before partitioning moves all three, and with them the ranking\footnote{This is not specific to our readout setup. RankZephyr generates a ranked list and jina-reranker-v3 scores 100 candidates in a single context, and both are order-sensitive at their native geometries (\Cref{app:specialized}).}

\textbf{Presentation channels.}
Four properties of the prompt can vary while the query and the candidate pool are held fixed:
(i) the order in which candidates are presented, (ii) the rule partitioning that order into windows, (iii) the markers that label the slots, and (iv) the wording of the answer skeleton from which scores are extracted.
We call one setting of all four a \emph{presentation}, and a scorer's sensitivity to it \emph{presentation dependence}; pool composition is separate, since it changes the candidate set and lies outside the prompt's control.
Except where stated, we vary the order and hold the rest fixed.
The partition rule (ii) is set when the prompt is built, so a different rule removes its cost.
Where the pool fits one window, as for QA, nothing is partitioned.
The markers (iii) and the skeleton (iv) are set at the same point and held fixed by default, since no skeleton setting is neutral; we vary each separately (\Cref{sec:threetask}; \Cref{app:mechanism}).
None of these removes the order channel (i), where the decisions vary.

\textbf{Measuring order dependence.}
We measure the order channel with $\tau$-PSI, adapting the index of \citet{zeng2025} to the rank level:
for each instance we score its candidate set under $M=10$ random permutations and take the induced rankings $r_1, \ldots, r_M$.
We then average their pairwise Kendall correlation $\tau(r_i, r_j)$:
\ifpreprint\vspace{-2pt}\else\vspace{-5pt}\fi
\[
\tau\text{-PSI} = \frac{1 - \mathrm{mean}_{i < j} \tau(r_i, r_j)}{2},
\]
which is 0 when every order produces the same ranking, 0.5 when rankings are uncorrelated, and 1 when reversed.
A permutation-invariant scorer treats its input as a set and attains 0, so $\tau$-PSI is a distance from set-like behavior and lower is more stable.
We randomly shuffle candidate order to model ordinary deployment variation, not a worst case; index-driven pool changes are evaluated separately (\Cref{sec:instrument}).
$\tau$-PSI is conditional on the partition rule, markers and skeleton.
We hold all three constant, so a difference in $\tau$-PSI is a difference between scorers.

\textbf{Measuring what the consumers produce.}
For each consumer, we report output quality and how often the output changes, since a scorer could hold it fixed by making it worse.
Throughout, change is the rate at which two permutations disagree, averaged over pairs, and independent of how many are drawn.
$\tau$-PSI is not a substitute for those measures: it measures movement in the ranking rather than in what a consumer produces, so the two need not rank scorers the same way (\Cref{app:consumers}).

\textbf{Two separate quantities.}
\label{sec:twoterms}
We write the scorer output for a query and candidate set $x = (q, D)$, at width $B$ and under a candidate order $\pi$, aligned back to candidate identity, as
\ifpreprint\vspace{2pt}\else\vspace{-0pt}\fi
\[
f_B(x, \pi)
= \underbrace{\mu_B(x)}_{\text{order-marginal}}
+ \underbrace{\delta_B(x, \pi)}_{\text{order residual}},
\]
with $\mu_B(x) = \mathbb{E}_\pi[f_B(x, \pi)]$ and $\mathbb{E}_\pi[\delta_B] = 0$.
The \emph{order-marginal} is the score a permutation-invariant scorer at that width would emit. The \emph{residual} is what remains, and it varies when only the input order changes; $\tau$-PSI measures the residual through its effect on the ranking.
Batched self-consistency (BSC) estimates the order-marginal by averaging a scorer's output over permutations \citep{korikov2025}.
The objective of \Cref{sec:method} targets the same order-invariance at one permutation, by penalizing the residual's empirical variance while leaving $\mu_B$ unconstrained.
Training should therefore reduce order instability before any serving-time averaging (\Cref{sec:instrument}).

\textbf{Width and scope.}
We call $\mu_B - \mu_1$ the \emph{width effect}, the difference between scoring a candidate at width $B$ and scoring it in isolation.
A pointwise scorer sets $B=1$, where the residual and the width effect are both zero by construction, so it avoids the order channel and forgoes an interaction whose sign varies by collection (\Cref{app:tasks}).
The decomposition runs over permutations of a fixed candidate set, so replacing a candidate changes $x$ rather than $\pi$ and falls outside it.
Whether an objective targeting $\delta_B$ reaches that perturbation is measured in \Cref{sec:instrument}.%

\ifpreprint\vspace{-6pt}\else\vspace{-2pt}\fi
\section{Where to address order dependence}
\ifpreprint\vspace{-3pt}\else\vspace{-1pt}\fi

\label{sec:method}

\Cref{sec:phenomenon} locates the order dependence in the residual $\delta_B$.
Holding the readout and teacher fixed, we ask whether penalizing that residual changes what the consumers decide at a single serving permutation.

\textbf{Score readout.}
\label{sec:readout}
A generic instruction-tuned chat model does not emit the per-candidate score that batched pointwise scoring requires, so we extract the score without generating any text.
The prompt supplies a grading instruction, the query, and the $B$ candidates that share the prompt, each tagged \texttt{[i]}, and requests one integer grade in $\{0,1,2,3\}$ per candidate. 
Rather than sampling those grades, we append a fixed answer skeleton with a placeholder grade token per candidate, so every grade position is known in advance and no decoding temperature enters.
One forward pass over the prompt and skeleton yields the model's distribution over the four grade tokens at each position, renormalized over those four.
We take the expected grade $E[g] = \sum_g g\,P(g)$, normalized to $[0,1]$, as the score, chosen by comparing readouts on untrained bases.
One prompt thus yields all $B$ scores.
Unlike a generated ranking, the readout cannot omit or repeat a candidate.
The score is continuous and comparable across instances, so it admits a decision threshold, and the readout is unchanged at any width down to $B{=}1$.
We apply the same readout at $B{=}10$ for QA and $B{=}4$ for response ranking training (\Cref{app:readout}).\looseness=-1

\textbf{Supervision.}
A teacher scores the candidates of each training instance and produces one target per candidate.
Because the readout is the expected grade, the targets are continuous rather than binary relevance labels.
We hold the teacher and the training pool fixed, so the variants below differ only in their objectives.
By default the teacher is the student's own base model (self-distillation).

\textbf{Training objectives.}
\begin{figure}[t]
\ifpreprint\vspace*{-24pt}\fi
\centering
\begin{tikzpicture}[figbase,
  every node/.style={figtxt},
  stagelbl/.style={figlbl},
  raillbl/.style={figmetric},
  rowlbl/.style={figlbl},
  rowlblours/.style={figlblours},
  quiet/.style={figmetric},
  note/.style={figmetric},
  bandlbl/.style={figmetric},
  onmark/.style={figonmark},
  stagebox/.style={figbox},
  marker/.style={figboxmark},
  markerours/.style={figboxmarkours},
]
  \useasboundingbox (0,0) rectangle (396.00,120.50);
  \draw[figrule] (107.02,100.00) -- (180.61,100.00);
  \draw[figrule] (107.02,100.00) -- (107.02,97.00);
  \draw[figrule] (180.61,100.00) -- (180.61,97.00);
  \node[bandlbl,anchor=base] at (143.32,112.70) {once per pool,};
  \node[bandlbl,anchor=base] at (143.32,104.20) {offline};
  \draw[figrule] (184.11,100.00) -- (279.09,100.00);
  \draw[figrule] (184.11,100.00) -- (184.11,97.00);
  \draw[figrule] (279.09,100.00) -- (279.09,97.00);
  \node[bandlbl,anchor=base] at (231.60,112.70) {once per model};
  \node[bandlbl,anchor=base] at (231.60,104.20) {$\times$ steps};
  \draw[figrule] (282.59,100.00) -- (357.79,100.00);
  \draw[figrule] (282.59,100.00) -- (282.59,97.00);
  \draw[figrule] (357.79,100.00) -- (357.79,97.00);
  \node[bandlbl,anchor=base] at (320.69,112.70) {every query};
  \node[bandlbl,anchor=base] at (320.69,104.20) {forever};
  \node[stagelbl,anchor=base] at (143.32,90.00) {teacher};
  \node[stagelbl,anchor=base] at (231.60,90.00) {training};
  \node[stagelbl,anchor=base] at (320.69,90.00) {serving};
  \draw[figflow] (82.29,86.00) -- (82.29,5.50);
  \node[raillbl,rotate=90] at (76.60,45.75) {pool};
  \draw[figflow] (382.52,86.00) -- (382.52,5.50);
  \node[raillbl,rotate=90] at (388.26,45.75) {ranking};
  \node[rowlbl,anchor=base west] at (2.00,78.60) {Single-order};
  \node[rowlbl,anchor=base west] at (8.00,69.60) {distillation};
  \draw[stagebox] (128.25,71.00) rectangle (158.38,83.00);
  \node[quiet,anchor=base] at (143.32,74.15) {1};
  \draw[stagebox] (206.35,71.00) rectangle (256.86,83.00);
  \node[quiet,anchor=base] at (231.60,74.20) {1 view};
  \draw[stagebox] (304.82,71.00) rectangle (336.56,83.00);
  \node[quiet,anchor=base] at (320.69,74.15) {1};
  \draw[figflow] (82.29,77.00) -- (123.05,77.00);
  \fill[figflow] (126.25,77.00) -- (123.05,75.66) -- (123.05,78.34) -- cycle;
  \draw[figflow] (160.38,77.00) -- (201.15,77.00);
  \fill[figflow] (204.35,77.00) -- (201.15,75.66) -- (201.15,78.34) -- cycle;
  \draw[figflow] (258.86,77.00) -- (299.62,77.00);
  \fill[figflow] (302.82,77.00) -- (299.62,75.66) -- (299.62,78.34) -- cycle;
  \draw[figflow] (338.56,77.00) -- (379.32,77.00);
  \fill[figflow] (382.52,77.00) -- (379.32,75.66) -- (379.32,78.34) -- cycle;
  \node[rowlbl,anchor=base west] at (2.00,58.60) {Order-averaged};
  \node[rowlbl,anchor=base west] at (8.00,49.60) {distillation};
  \draw[marker] (128.25,51.00) rectangle (158.38,63.00);
  \node[onmark,anchor=base] at (143.32,54.27) {$T$};
  \draw[stagebox] (206.35,51.00) rectangle (256.86,63.00);
  \node[quiet,anchor=base] at (231.60,54.20) {1 view};
  \draw[stagebox] (304.82,51.00) rectangle (336.56,63.00);
  \node[quiet,anchor=base] at (320.69,54.15) {1};
  \draw[figflow] (82.29,57.00) -- (123.05,57.00);
  \fill[figflow] (126.25,57.00) -- (123.05,55.66) -- (123.05,58.34) -- cycle;
  \draw[figflow] (160.38,57.00) -- (201.15,57.00);
  \fill[figflow] (204.35,57.00) -- (201.15,55.66) -- (201.15,58.34) -- cycle;
  \draw[figflow] (258.86,57.00) -- (299.62,57.00);
  \fill[figflow] (302.82,57.00) -- (299.62,55.66) -- (299.62,58.34) -- cycle;
  \draw[figflow] (338.56,57.00) -- (379.32,57.00);
  \fill[figflow] (382.52,57.00) -- (379.32,55.66) -- (379.32,58.34) -- cycle;
  \node[rowlblours,anchor=base west] at (2.00,34.14) {OC-SFT};
  \draw[stagebox] (128.25,31.00) rectangle (158.38,43.00);
  \node[quiet,anchor=base] at (143.32,34.15) {1};
  \draw[markerours] (206.35,31.00) rectangle (256.86,43.00);
  \node[onmark,anchor=base] at (231.60,34.20) {2 views};
  \draw[stagebox] (304.82,31.00) rectangle (336.56,43.00);
  \node[quiet,anchor=base] at (320.69,34.15) {1};
  \draw[figflow] (82.29,37.00) -- (123.05,37.00);
  \fill[figflow] (126.25,37.00) -- (123.05,35.66) -- (123.05,38.34) -- cycle;
  \draw[figflow] (160.38,37.00) -- (201.15,37.00);
  \fill[figflow] (204.35,37.00) -- (201.15,35.66) -- (201.15,38.34) -- cycle;
  \draw[figflow] (258.86,37.00) -- (299.62,37.00);
  \fill[figflow] (302.82,37.00) -- (299.62,35.66) -- (299.62,38.34) -- cycle;
  \draw[figflow] (338.56,37.00) -- (379.32,37.00);
  \fill[figflow] (382.52,37.00) -- (379.32,35.66) -- (379.32,38.34) -- cycle;
  \node[rowlbl,anchor=base west] at (2.00,16.99) {Batched};
  \node[rowlbl,anchor=base west] at (8.00,7.99) {self-consistency};
  \draw[marker] (304.82,8.50) rectangle (336.56,20.50);
  \node[onmark,anchor=base] at (320.69,11.77) {$K$};
  \draw[figflow] (82.29,14.50) -- (299.62,14.50);
  \fill[figflow] (302.82,14.50) -- (299.62,13.16) -- (299.62,15.84) -- cycle;
  \draw[figflow] (338.56,14.50) -- (379.32,14.50);
  \fill[figflow] (382.52,14.50) -- (379.32,13.16) -- (379.32,15.84) -- cycle;
  \node[note,anchor=base] at (193.55,2.70) {no teacher, no training, no labels};
  \draw[fighair] (83.79,25.50) -- (381.02,25.50);
\end{tikzpicture}%
\caption{\textbf{Order dependence can be addressed at serving, in the labels, or once in training.} %
Each row is one approach, and the filled marker is where it encounters several permutations; single-order distillation, the main baseline, never sees more than one. %
$T$ counts permutations of one training instance and $K$ permutations of one query; the two are independent, and OC-SFT's two views are the $N{=}2$ of its objective.
We use $T{=}10$ and $K{=}10$; \Cref{fig:amortconsumers} measures how far averaging gets.}
\label{fig:remedies}
\ifpreprint\vspace{-6pt}\else\vspace{-3pt}\fi
\end{figure}
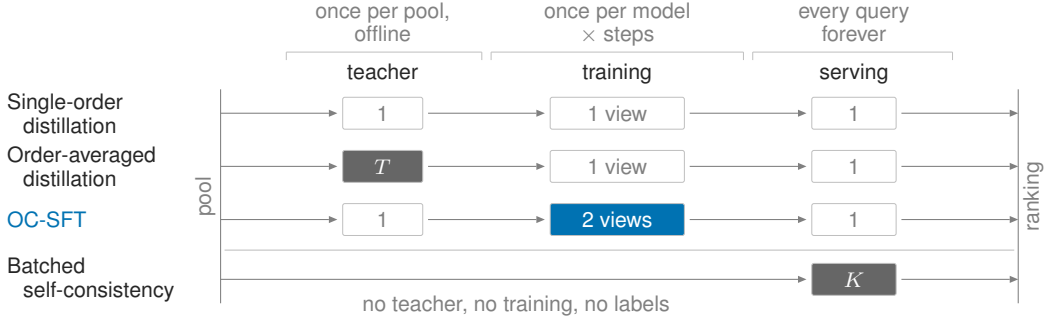
The approaches to $\mu_B$ differ in where they encounter several permutations (\Cref{fig:remedies}): self-consistency at serving time, order-averaged distillation offline in the labels, and OC-SFT in its $N$ views per step.
All trained variants distill from these targets, minimizing a squared error between the student's expected grade $s(d)$ and the teacher target $\tilde{y}(d)$, averaged over the candidates of a $B=20$ window.
Single-order distillation, or plain fine-tuning, uses the target from one teacher order and does not target $\mu_B$.
Order-averaged distillation uses the batched self-consistency target, the per-candidate average over $T=10$ shuffled teacher orders precomputed into the label.
The label absorbs the averaging, so order-averaged distillation targets $\mu_B$ through the teacher.

We introduce \textbf{order-consistency SFT} (OC-SFT), which instead constrains the student's own order residual $\delta_B$ (\Cref{sec:twoterms}), following the consistency-regularization principle of semi-supervised learning \citep{tarvainen2017, liang2021}.
A permutation-invariant scorer has no such residual, so penalizing it targets invariance in the student rather than through the label.
OC-SFT retains the single-order target and scores the same window under $N$ permutations $\pi_1, \ldots, \pi_N$, with per-candidate scores $s_i(d)$, each $f_B$ at $\pi_i$ restricted to $d$, and view-mean $\bar{s}(d) = \tfrac{1}{N}\sum_{i=1}^{N} s_i(d)$.
\footnote{These shuffles permute slots within a window, so a candidate keeps its companions, whereas a $\tau$-PSI shuffle can change them; the metric therefore perturbs at least as much as the penalty targets.}
A penalty with weight $\lambda$ then pulls each view toward that mean:
\ifpreprint\vspace{-1pt}\else\vspace{-3pt}\fi
\[
\mathcal{L}_{\text{OC-SFT}} = \underbrace{\mathrm{mean}_d\,(s_1(d) - \tilde{y}(d))^2}_{\text{relevance anchor}} \;+\; \lambda\,\underbrace{\frac{1}{N}\sum_{i=1}^{N} \mathrm{mean}_d\,(s_i(d) - \bar{s}(d))^2}_{\text{order variance: } \delta_B}.
\]

\ifpreprint\vspace{-1pt}\else\vspace{-2pt}\fi

Only the first view, itself one of the shuffles, carries the relevance anchor, so more views add no teacher supervision.
The second term is the empirical order-variance of the student's own scores, proportional in expectation to $\mathbb{E}_\pi \lVert \delta_B(x,\pi) \rVert^2$, so it directly penalizes the within-window residual.
The penalty acts on scores and bounds their variance rather than removing it, whereas $\tau$-PSI reads rankings; near a tie, the surviving variance can still move a rank.
The objective therefore yields attenuation rather than invariance, in the ranking and in the decisions read off it (\Cref{sec:instrument}).\looseness=-1\footnote{In group-relative terms, $\mu_B$ is the group mean and $\delta_B$ the deviation, as in GRPO \citep{shao2024}; the group varies the arrangement, not the sampled action, so the residual is differentiable and needs no rollouts.}
We use $N=2$.\footnote{One pair estimates that variance up to a constant absorbed into $\lambda$, and we detect no quality change to $N=8$ (\Cref{app:silver}). Two views roughly double the training step cost; all trained variants serve one permutation.}
The tenfold saving over order-averaged distillation is in offline teacher labels and not in training compute.

Both terms are required:
without the relevance anchor the model collapses to a consistent but relevance-free scorer, and at $\lambda=0$ the objective reduces to single-order distillation.
The readout is unchanged across widths, so the weights can serve a different $B$ without retraining (\Cref{app:width}).

\section{Experimental setup}
\label{sec:setup}

\textbf{Data.}
We train and evaluate on three tasks that score a candidate set in one prompt.
Each has its own training data and teacher labels; the readout and the objective are the same throughout.

\textbf{Passage reranking} is the primary setting.
We train on MS MARCO \citep{nguyen2016} at $B=20$, deriving labels from a teacher that grades the top-100 candidates of $\sim$30K queries under the readout of \Cref{sec:readout}.
Evaluation covers 18 collections:
five in-domain TREC Deep Learning collections built over MS MARCO (DL19 through DL23, \citealp{craswell2020}), 11 out-domain BEIR collections \citep{thakur2021}, and two internal legal collections from a commercial legal search provider, with the BEIR and legal groups both evaluated zero-shot.
The 16 public collections match the suite that recent listwise and distillation rerankers report on.
The only two that carry claim-verification labels, SciFact with 300 claims and Climate-FEVER with 1{,}381, also serve as the downstream collections for verdict flip.
\textbf{Multi-document QA} scores answer support rather than topical relevance:
the HotpotQA distractor setting, ten passages per question with exactly two gold and binary relevance \citep{yang2018}, trained on 30K questions and transferred zero-shot to 2WikiMultiHopQA \citep{ho2020} and MuSiQue \citep{trivedi2022}.
\textbf{Response ranking} scores candidate responses for preference or correctness, trained on $\sim$30.7K UltraFeedback prompts \citep{cui2024} with four candidates each.
Its five collections each cover a distinct regime, differing in the quality gap between candidates and in whether one can be judged in isolation: RewardBench-2 \citep{malik2025}, Nectar \citep{zhu2023}, PPE MMLU-Pro and PPE-MATH \citep{frick2024}, and RM-Bench \citep{liu2025rmbench}.
\Cref{app:metrics} lists every evaluation collection; scope and limitations are discussed in \Cref{app:limitations}.

\textbf{Models and teachers.}
Qwen3-4B \citep{yang2025} is our primary base model, one of 11 dense bases from 1.7B to 32B across three families: Qwen3, Gemma 4 \citep{gemmateam2026} and Granite 4.1 \citep{granite2026}.
Two bases serve as controls, the sparse Mixture-of-Experts Gemma-4 26B-A4B for architecture and Qwen3-Reranker-4B for reranking-specialized initialization.
The default teacher is the student's own base model (self-distillation), so the method needs no external model and the reported gains cannot come from a stronger teacher; we also evaluate larger open teachers within and across families (Qwen3, Gemma 4, Granite 4.1) and GPT-5.4. 
We train with LoRA \citep{hu2022} throughout, selecting checkpoints and hyperparameters per model and variant on a held-out split (\Cref{app:repro}); base and teacher models are varied in \Cref{app:crossbase}.

\textbf{Baselines.}
All reranking baselines use the same $100$ BM25 candidates as our trained models; we repeat the comparison on a dense (BGE-base-en-v1.5, \citealp{xiao2024}) and a learned-sparse (SPLADE++ ED, \citealp{formal2022}) first stage (\Cref{app:firststage}).
The inference-time comparison is batched self-consistency \citep{korikov2025}, the $K$-permutation ensemble, together with CapCal \citep{lv2026}, a content-agnostic logit calibration applied to the off-the-shelf base.
As a trained mitigation we adapt DebiasFirst \citep{qiao2026}, position-aware augmentation with inverse-propensity weighting, to batched score regression.
Alongside it we report permutation augmentation, as in RankVicuna \citep{pradeep2023vicuna}, which retains the $N$ shuffled views and anchors each to the teacher target.\looseness=-1

We also report the purpose-built jina-reranker-v3 \citep{wang2025jina} at our $B=20$ geometry and a frontier prompted scorer, GPT-5.4, on reranking, QA, and response ranking; nearly half its candidate pairs tie, so we omit its $\tau$-PSI.
\Cref{app:specialized} additionally evaluates RankZephyr \citep{pradeep2023} and mxbai-rerank-large-v2 \citep{li2026} at their native geometries.
On response ranking we compare against the pointwise reward model Skywork-Reward-V2-Qwen3-4B \citep{liu2025skywork} and the pairwise PairRM \citep{jiang2023}; neither sees a candidate order, so we report quality only (\Cref{app:respranking}).\looseness=-1

\textbf{Ranking metrics.}
We report nDCG@10 for quality and $\tau$-PSI for order instability (\Cref{sec:phenomenon}); truncation at 10 affects only reranking's 100 candidates, since QA ranks its ten candidates in full.
Response ranking instead reports nDCG@1, the normalized relevance gain of the top-ranked response.

\textbf{Decision metrics.}
For the threshold results, we freeze a per-collection F1-tuned cutoff, fitted on half of a collection's queries and measured on the other half.
Reshuffling therefore changes what the filter admits and not how it was tuned.
For that retained set we report its F1 and its mean pairwise Jaccard, the overlap between the sets retained under different permutations.
Downstream, two frozen readers consume the scorer's top 5 under the same $M{=}10$ permutations, Granite-4.1-8B throughout and Qwen3-4B as a second reader.
For them, we report verdict accuracy and \emph{verdict flip} on claim verification, and exact match, F1 and \emph{answer flip} on QA.
A preference model's scores determine a chosen/rejected response pair, for which we report \emph{pair flip}, how often that pair changes.
Each consumer is held fixed, so any change comes from the scorer alone.

\textbf{Aggregation.}
Task means equally weight 18 reranking, 3 QA and 5 response-ranking collections; $\tau$-PSI and flip metrics use $M{=}10$ random permutations.
Trained rows average 3 seeds, otherwise seed 42; intervals use paired 95\% bootstraps over collections then queries.

\begin{table}[t]
\centering
\small
\caption{\textbf{Quality saturates on all three tasks; what the consumers decide does not.}
Each task block gives quality, order instability, and its consumer's decision: retained-set overlap (Jacc.), answer flip and chosen/rejected-pair flip.
\textbf{Bold} marks a Holm-adjusted lead over the other trained variants on all three seeds (\Cref{app:inference}); no quality column qualifies (\Cref{sec:buys}).
Basis: Qwen3-4B, $M{=}10$ random permutations, trained rows self-distilled 3-seed means; %
BSC is batched self-consistency \citep{korikov2025} over the off-the-shelf scorer.
Collections and per-task widths are in \Cref{app:setup}.
}
\label{tab:levels}
\begin{threeparttable}
\setlength{\tabcolsep}{3pt}
\begin{tabular}{@{}l@{\hspace{8pt}}rrr@{\hspace{8pt}}rrr@{\hspace{8pt}}rrr@{}}
\toprule
& \multicolumn{3}{c}{Passage reranking (18)} & \multicolumn{3}{c}{Multi-doc QA (3)} & \multicolumn{3}{c}{Response ranking (5)} \\
\cmidrule(lr){2-4} \cmidrule(lr){5-7} \cmidrule(lr){8-10}
Variant & nDCG@10$\uparrow$ & $\tau$-PSI$\downarrow$ & Jacc.$\uparrow$ & nDCG@10$\uparrow$ & $\tau$-PSI$\downarrow$ & Ans.$\downarrow$ & nDCG@1$\uparrow$ & $\tau$-PSI$\downarrow$ & Pair$\downarrow$ \\
\midrule
\multicolumn{10}{@{}l}{\textbf{Untrained}} \\
\hspace*{3pt}Off the shelf & 0.370 & 0.298 & 0.439 & 0.911 & 0.224 & 0.221 & 0.655 & 0.345 & 0.877 \\ 
\hspace*{3pt}CapCal & 0.372 & 0.293 & 0.427 & 0.911 & 0.222 & 0.217 & 0.657 & 0.338 & 0.874 \\
\hspace*{3pt}Round-robin & 0.422 & 0.297 & 0.443 & 0.911 & 0.224 & 0.221 & 0.658 & 0.347 & 0.878 \\
\hspace*{3pt}BSC ($\times$10)\tnote{a} & 0.465 & 0.180 & 0.707 & 0.946 & 0.143 & 0.157 & 0.716 & 0.184 & 0.636 \\
\midrule
\multicolumn{10}{@{}l}{\textbf{External}} \\
\hspace*{3pt}\emph{jina-reranker-v3} & 0.447 & 0.177 & 0.667 & 0.949 & 0.163 & 0.172 & 0.479 & 0.226 & 0.685 \\
\hspace*{3pt}\emph{GPT-5.4} & 0.468 & ---\tnote{b} & 0.707 & 0.972 & ---\tnote{b} & 0.094 & 0.726 & ---\tnote{b} & 0.489 \\
\midrule
\multicolumn{10}{@{}l}{\textbf{Trained}} \\
\hspace*{3pt}Single-order & 0.449 & 0.209 & 0.656 & 0.951 & 0.159 & 0.177 & 0.684 & 0.333 & 0.869 \\
\hspace*{3pt}Order-averaged & 0.455 & 0.130 & 0.743 & 0.956 & 0.124 & 0.149 & 0.693 & 0.228 & 0.724 \\
\hspace*{3pt}DebiasFirst & 0.454 & 0.128 & 0.759 & 0.955 & 0.147 & 0.164 & 0.694 & 0.228 & 0.718 \\
\hspace*{3pt}Perm.\ aug.\ & 0.455 & 0.129 & 0.760 & 0.955 & 0.148 & 0.162 & 0.696 & 0.223 & 0.707 \\
\dashedmidrule
\hspace*{3pt}OC-SFT & 0.459 & \textbf{0.083} & \textbf{0.835} & 0.961 & \textbf{0.096} & \textbf{0.125} & 0.701 & \textbf{0.201} & \textbf{0.661} \\
\bottomrule
\end{tabular}
\begin{tablenotes}[flushleft]
\footnotesize
\item[a] Its order-instability and decision cells compare four independent ten-permutation ensembles, the unit such a system serves; every other row compares single permutations, at a tenth of the inference.
\item[b] Nearly half its candidate pairs tie. We use an order-independent tie-break; stable sorting raises answer/pair flip from 0.094/0.489 to 0.200/0.851. $\tau$-PSI credits fixed tie orders to the model, so we omit it.
\end{tablenotes}
\end{threeparttable}
\end{table}

\section{Results}
\label{sec:results}

\setcounter{topnumber}{1}
\setcounter{totalnumber}{1}

\subsection{Decisions at matched ranking quality}
\label{sec:buys}
\label{sec:downstream}

\begin{figure}[t]
    \centering
    \begin{minipage}[t]{0.5\linewidth}\centering
    \begin{tikzpicture}[figbase7,
  x=1cm, y=1cm,
  rowlbl/.style={figtxt, figlbl, anchor=base east},
  rowlblours/.style={figtxt, figlblours, anchor=base east},
  title/.style={figtxt, figtitle7, anchor=base west},
  header/.style={figtxt, figmetric, anchor=base east},
  quietc/.style={figtxt, figtick, anchor=base},
  axisttl/.style={figtxt, figaxislbl, anchor=base},
  dot/.style={figtxt, circle, minimum size=3.0pt},
]
  \useasboundingbox (0.020,-3.520) rectangle (6.658,0.690);
  \draw[figgrid] (2.346,-2.890) -- (2.346,0.120);
  \draw[figgrid] (3.623,-2.890) -- (3.623,0.120);
  \draw[figgrid] (4.900,-2.890) -- (4.900,0.120);
  \node[rowlbl] at (1.743,-0.085) {Off the shelf};
  \fill[figband] (2.114,-0.070) rectangle (5.505,0.070);
  \draw[figbandcap] (2.114,-0.105) -- (2.114,0.105);
  \draw[figbandcap] (5.505,-0.105) -- (5.505,0.105);
  \node[dot,figmark] at (3.842,0.000) {};
  \node[rowlbl] at (6.658,-0.085) {0.135};
  \draw[fighair] (0.020,-0.255) -- (6.658,-0.255);
  \node[rowlbl] at (1.743,-0.595) {Single-order};
  \fill[figband] (3.400,-0.580) rectangle (5.402,-0.440);
  \draw[figbandcap] (3.400,-0.615) -- (3.400,-0.405);
  \draw[figbandcap] (5.402,-0.615) -- (5.402,-0.405);
  \node[dot,figmark] at (4.411,-0.510) {};
  \node[rowlbl] at (6.658,-0.595) {0.079};
  \node[rowlbl] at (1.743,-1.105) {Order-averaged};
  \fill[figband] (3.648,-1.090) rectangle (5.279,-0.950);
  \draw[figbandcap] (3.648,-1.125) -- (3.648,-0.915);
  \draw[figbandcap] (5.279,-1.125) -- (5.279,-0.915);
  \node[dot,figmark] at (4.468,-1.020) {};
  \node[rowlbl] at (6.658,-1.105) {0.064};
  \node[rowlbl] at (1.743,-1.615) {DebiasFirst};
  \fill[figband] (3.685,-1.600) rectangle (5.261,-1.460);
  \draw[figbandcap] (3.685,-1.635) -- (3.685,-1.425);
  \draw[figbandcap] (5.261,-1.635) -- (5.261,-1.425);
  \node[dot,figmark] at (4.479,-1.530) {};
  \node[rowlbl] at (6.658,-1.615) {0.062};
  \node[rowlbl] at (1.743,-2.125) {Perm. aug.};
  \fill[figband] (3.707,-2.110) rectangle (5.257,-1.970);
  \draw[figbandcap] (3.707,-2.145) -- (3.707,-1.935);
  \draw[figbandcap] (5.257,-2.145) -- (5.257,-1.935);
  \node[dot,figmark] at (4.482,-2.040) {};
  \node[rowlbl] at (6.658,-2.125) {0.061};
  \node[rowlblours] at (1.743,-2.635) {OC-SFT};
  \fill[figbandours] (3.955,-2.620) rectangle (5.122,-2.480);
  \draw[figbandcapours] (3.955,-2.655) -- (3.955,-2.445);
  \draw[figbandcapours] (5.122,-2.655) -- (5.122,-2.445);
  \node[dot,figmarkours] at (4.540,-2.550) {};
  \node[rowlblours] at (6.658,-2.635) {0.046};
  \node[quietc] at (2.114,0.155) {worst};
  \node[quietc] at (3.842,0.155) {mean};
  \node[quietc] at (5.505,0.155) {best};
  \node[title] at (0.020,0.470) {(a) Per-query permutation spread};
  \node[header] at (6.658,0.470) {spread};
  \draw[figrule] (5.906,0.380) -- (6.658,0.380);
  \draw[figrule] (1.963,-2.890) -- (5.666,-2.890);
  \draw[figrule] (2.346,-2.890) -- (2.346,-2.975);
  \node[quietc] at (2.346,-3.090) {0.30};
  \draw[figrule] (3.623,-2.890) -- (3.623,-2.975);
  \node[quietc] at (3.623,-3.090) {0.40};
  \draw[figrule] (4.900,-2.890) -- (4.900,-2.975);
  \node[quietc] at (4.900,-3.090) {0.50};
  \node[axisttl] at (3.814,-3.430) {nDCG@10};
\end{tikzpicture}%
    \end{minipage}\hfill
    \begin{minipage}[t]{0.5\linewidth}\centering
    \input{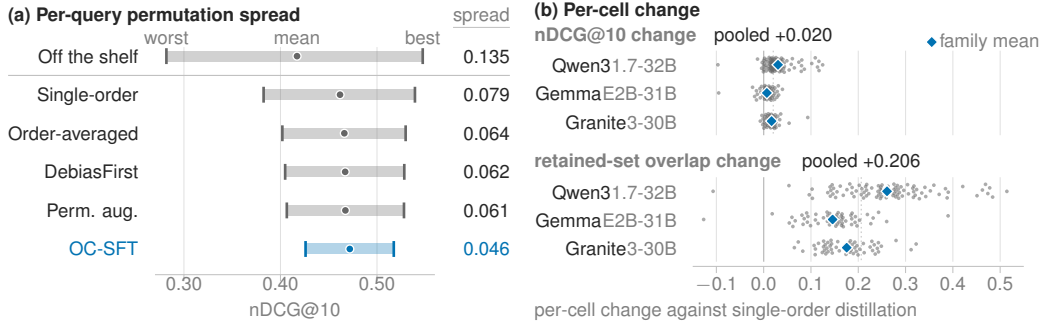}
    \end{minipage}
    \caption{%
    \textbf{Ranking quality moves across permutations and barely with the method; quality and retained-set overlap separate in every family.}
    (a) Qwen3-4B per-query nDCG@10 over $M{=}10$ random permutations on 18 collections; \emph{spread} is mean minus worst.
    (b) Dots are 198 base-collection cells over 11 dense bases; overlap at a per-collection F1-tuned cutoff, \emph{pooled} is their mean.}
    \label{fig:spread} 
    \end{figure}

\textbf{Ranking quality does not predict how far a scorer's decisions change under reshuffling.}
The five \emph{trained} variants in \Cref{tab:levels} differ by at most 0.010 nDCG@10 on passage reranking, less than a permutation moves a single query's score under any one of them (\Cref{fig:spread}a).
Across the $M{=}10$ random permutations of each pool, retained sets overlap pairwise by 0.656 under single-order distillation and 0.835 under OC-SFT, a difference more than an order of magnitude larger than that quality spread.
Among the seven single-permutation systems in \Cref{tab:levels} within 0.022 nDCG@10, different systems take the best quality and the best retained-set overlap on 11 of the 18 reranking collections (\Cref{app:rankmetrics}).
A threshold reads absolute scores; nDCG@10 reads only the discounted order of the top ten.
So a small score change near the cutoff admits or drops a document, while similarly graded candidates trading places barely moves nDCG@10.
A comparison of such scorers should report what a threshold retains or a reader answers, not the ranking metric alone.

\textbf{Two published systems lead on a quality measure while reproducing less of their own output than the scorer they outrank},
sharing neither our base model nor our supervision.
GPT-5.4 leads every variant we trained on nDCG@10 while reproducing less of its retained set than OC-SFT (\Cref{tab:levels}).
\texttt{jina-reranker-v3}, a purpose-built reranker rather than a frontier model, leads OC-SFT on retained-set F1 while trailing it on nDCG@10, and its retained sets overlap less, 0.667 against 0.835.\looseness=-1

\subsection{Channels fixed at prompt construction}
\label{sec:threetask}

\textbf{Fixes applied before training do not improve the decisions.}
The contiguous partition of a rank-sorted top 100 (\Cref{sec:phenomenon}) makes each window internally uniform on a four-grade scale, so equal grades in different windows denote different relevance levels and the merged ranking conflates them.
Round-robin assignment distributes consecutive first-stage ranks across distinct windows and recovers 0.052 nDCG@10 on reranking, without additional permutations (\Cref{tab:levels}, \emph{untrained} rows).
However, it leaves decision stability where it was: off the shelf, neither retained-set overlap, verdict flip on either collection, nor pair flip changes measurably.
A post-hoc logit calibration \citep{lv2026} improves neither quality nor retained-set overlap measurably, and leaves $\tau$-PSI near the off-the-shelf level.
Rescaling across windows \citep{ren2025} cannot correct the partition effect at all, since the windows do not differ by an offset.
We report under contiguous windows, the conventional partition.
Under round-robin the trained gain is smaller and still positive, and the partition rule moves no variant's $\tau$-PSI by more than 0.006.
Detailed results are in \Cref{app:mitigations}. %

Scores are read from a fixed grade slot, so the placeholder in every unscored slot is part of the presentation.
We adopt \texttt{Grade:} \texttt{0}, the in-format placeholder with the best average ranking quality across our three families, and hold it fixed; no other in-format value lowers $\tau$-PSI distinguishably (\Cref{app:readout}).
We are aware of no position-bias measurement, including the index we adapt, that reports its sensitivity to this choice; such numbers are therefore in part a property of the skeleton behind them.
None of the alternatives tested removes the order dependence that moves the decisions.

\FloatBarrier

\vspace{-2pt}

\subsection{What training changes for the consumers}
\label{sec:instrument}
\vspace{-1pt}

\textbf{The penalty moves what the consumers decide far more than it moves ranking quality.}
Order instability on passage reranking falls from 0.209 under single-order distillation to 0.083 under OC-SFT, and the retained set, the reader's output and the selected pair all become more reproducible (\Cref{tab:levels}).
On quality alone the two variants are close; on retained-set stability they are not.
As \Cref{fig:spread}a shows, OC-SFT also narrows per-query score movement below every other variant.

\textbf{The gain holds model by model and under unseen perturbations, and no control removes it.}
Quality and retained-set overlap separate the same way per model and collection across 11 dense bases in three families (\Cref{fig:spread}b), and OC-SFT stays below order-averaged distillation in $\tau$-PSI on all 12, including the sparse MoE.
The same ordering appears under dense and learned-sparse first stages.
Under pool replacement, a perturbation the objective was never trained on, instability under that perturbation is 0.031 against single-order distillation's 0.099.
Matching retention at the F1-tuned cutoff leaves the gain in retained-set overlap in place.
What the penalty removes is movement at the ranking head, not ties or collapsed scores, and the ordering holds under worst-permutation quality and dispersion.
The Climate-FEVER verdict changes less often than under single-order distillation, with no measurable accuracy change.
The lower verdict-flip rate persists among queries for which both variants select the same five documents. For detailed results, we refer to \Cref{app:mechanisms,app:mitigations,app:consumers}.\looseness=-1

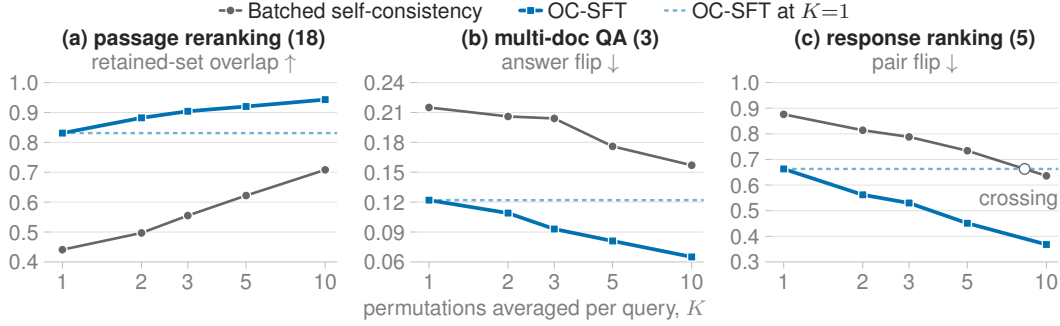
\begin{figure}[t]
\centering
\begin{tikzpicture}[figbase,
  every node/.style={figtxt},
  paneltitle/.style={figtitle},
  panelmetric/.style={figmetric},
  ticklbl/.style={figtick},
  axislbl/.style={figaxislbl},
  legendlbl/.style={figlbl},
  crossinglbl/.style={figmetric},
]
  \useasboundingbox (0,0) rectangle (396.00,120.70);
  \node[axislbl,anchor=base] at (198.00,2.80) {permutations averaged per query, $K$};
  \draw[figgrid] (15.62,33.85) -- (123.52,33.85);
  \draw[figgrid] (15.62,45.10) -- (123.52,45.10);
  \draw[figgrid] (15.62,56.35) -- (123.52,56.35);
  \draw[figgrid] (15.62,67.60) -- (123.52,67.60);
  \draw[figgrid] (15.62,78.85) -- (123.52,78.85);
  \draw[figgrid] (15.62,90.10) -- (123.52,90.10);
  \draw[figrule] (15.62,22.60) -- (123.52,22.60);
  \draw[figrule] (13.12,22.60) -- (15.62,22.60);
  \node[ticklbl,anchor=base east] at (11.12,19.75) {0.4};
  \draw[figrule] (13.12,33.85) -- (15.62,33.85);
  \node[ticklbl,anchor=base east] at (11.12,31.00) {0.5};
  \draw[figrule] (13.12,45.10) -- (15.62,45.10);
  \node[ticklbl,anchor=base east] at (11.12,42.25) {0.6};
  \draw[figrule] (13.12,56.35) -- (15.62,56.35);
  \node[ticklbl,anchor=base east] at (11.12,53.50) {0.7};
  \draw[figrule] (13.12,67.60) -- (15.62,67.60);
  \node[ticklbl,anchor=base east] at (11.12,64.75) {0.8};
  \draw[figrule] (13.12,78.85) -- (15.62,78.85);
  \node[ticklbl,anchor=base east] at (11.12,76.00) {0.9};
  \draw[figrule] (13.12,90.10) -- (15.62,90.10);
  \node[ticklbl,anchor=base east] at (11.12,87.25) {1.0};
  \draw[figrule] (20.12,22.60) -- (20.12,20.10);
  \node[ticklbl,anchor=base] at (20.12,12.40) {1};
  \draw[figrule] (49.89,22.60) -- (49.89,20.10);
  \node[ticklbl,anchor=base] at (49.89,12.40) {2};
  \draw[figrule] (67.31,22.60) -- (67.31,20.10);
  \node[ticklbl,anchor=base] at (67.31,12.40) {3};
  \draw[figrule] (89.25,22.60) -- (89.25,20.10);
  \node[ticklbl,anchor=base] at (89.25,12.40) {5};
  \draw[figrule] (119.02,22.60) -- (119.02,20.10);
  \node[ticklbl,anchor=base] at (119.02,12.40) {10};
  \draw[figlevel] (20.12,71.09) -- (123.52,71.09);  %
  \draw[figseries] (20.12,27.21) -- (49.89,33.51) -- (67.31,40.04) -- (89.25,47.58) -- (119.02,57.25);
  \draw[figours] (20.12,71.09) -- (49.89,76.83) -- (67.31,79.30) -- (89.25,81.10) -- (119.02,83.69);
  \draw[figmark] (20.12,27.21) circle (1.70pt);  %
  \draw[figmark] (49.89,33.51) circle (1.70pt);  %
  \draw[figmark] (67.31,40.04) circle (1.70pt);  %
  \draw[figmark] (89.25,47.58) circle (1.70pt);  %
  \draw[figmark] (119.02,57.25) circle (1.70pt);  %
  \draw[figmarkours] (18.57,69.54) rectangle (21.67,72.64);  %
  \draw[figmarkours] (48.34,75.28) rectangle (51.44,78.38);  %
  \draw[figmarkours] (65.76,77.75) rectangle (68.86,80.85);  %
  \draw[figmarkours] (87.70,79.55) rectangle (90.80,82.65);  %
  \draw[figmarkours] (117.47,82.14) rectangle (120.57,85.24);  %
  \node[paneltitle,anchor=base] at (69.57,104.30) {(a) passage reranking (18)};
  \node[panelmetric,anchor=base] at (69.57,95.10) {retained-set overlap~$\uparrow$};
  \draw[figgrid] (153.59,33.85) -- (261.48,33.85);
  \draw[figgrid] (153.59,45.10) -- (261.48,45.10);
  \draw[figgrid] (153.59,56.35) -- (261.48,56.35);
  \draw[figgrid] (153.59,67.60) -- (261.48,67.60);
  \draw[figgrid] (153.59,78.85) -- (261.48,78.85);
  \draw[figgrid] (153.59,90.10) -- (261.48,90.10);
  \draw[figrule] (153.59,22.60) -- (261.48,22.60);
  \draw[figrule] (151.09,22.60) -- (153.59,22.60);
  \node[ticklbl,anchor=base east] at (149.09,19.75) {0.06};
  \draw[figrule] (151.09,33.85) -- (153.59,33.85);
  \node[ticklbl,anchor=base east] at (149.09,31.00) {0.09};
  \draw[figrule] (151.09,45.10) -- (153.59,45.10);
  \node[ticklbl,anchor=base east] at (149.09,42.25) {0.12};
  \draw[figrule] (151.09,56.35) -- (153.59,56.35);
  \node[ticklbl,anchor=base east] at (149.09,53.50) {0.15};
  \draw[figrule] (151.09,67.60) -- (153.59,67.60);
  \node[ticklbl,anchor=base east] at (149.09,64.75) {0.18};
  \draw[figrule] (151.09,78.85) -- (153.59,78.85);
  \node[ticklbl,anchor=base east] at (149.09,76.00) {0.21};
  \draw[figrule] (151.09,90.10) -- (153.59,90.10);
  \node[ticklbl,anchor=base east] at (149.09,87.25) {0.24};
  \draw[figrule] (158.09,22.60) -- (158.09,20.10);
  \node[ticklbl,anchor=base] at (158.09,12.40) {1};
  \draw[figrule] (187.86,22.60) -- (187.86,20.10);
  \node[ticklbl,anchor=base] at (187.86,12.40) {2};
  \draw[figrule] (205.27,22.60) -- (205.27,20.10);
  \node[ticklbl,anchor=base] at (205.27,12.40) {3};
  \draw[figrule] (227.21,22.60) -- (227.21,20.10);
  \node[ticklbl,anchor=base] at (227.21,12.40) {5};
  \draw[figrule] (256.98,22.60) -- (256.98,20.10);
  \node[ticklbl,anchor=base] at (256.98,12.40) {10};
  \draw[figlevel] (158.09,45.85) -- (261.48,45.85);  %
  \draw[figseries] (158.09,80.73) -- (187.86,77.35) -- (205.27,76.60) -- (227.21,66.10) -- (256.98,58.98);
  \draw[figours] (158.09,45.85) -- (187.86,40.98) -- (205.27,34.98) -- (227.21,30.48) -- (256.98,24.48);
  \draw[figmark] (158.09,80.73) circle (1.70pt);  %
  \draw[figmark] (187.86,77.35) circle (1.70pt);  %
  \draw[figmark] (205.27,76.60) circle (1.70pt);  %
  \draw[figmark] (227.21,66.10) circle (1.70pt);  %
  \draw[figmark] (256.98,58.98) circle (1.70pt);  %
  \draw[figmarkours] (156.54,44.30) rectangle (159.64,47.40);  %
  \draw[figmarkours] (186.31,39.43) rectangle (189.41,42.53);  %
  \draw[figmarkours] (203.72,33.43) rectangle (206.82,36.53);  %
  \draw[figmarkours] (225.66,28.93) rectangle (228.76,32.03);  %
  \draw[figmarkours] (255.43,22.93) rectangle (258.53,26.03);  %
  \node[paneltitle,anchor=base] at (207.53,104.30) {(b) multi-doc QA (3)};
  \node[panelmetric,anchor=base] at (207.53,95.10) {answer flip~$\downarrow$};
  \draw[figgrid] (287.10,32.24) -- (395.00,32.24);
  \draw[figgrid] (287.10,41.89) -- (395.00,41.89);
  \draw[figgrid] (287.10,51.53) -- (395.00,51.53);
  \draw[figgrid] (287.10,61.17) -- (395.00,61.17);
  \draw[figgrid] (287.10,70.81) -- (395.00,70.81);
  \draw[figgrid] (287.10,80.46) -- (395.00,80.46);
  \draw[figgrid] (287.10,90.10) -- (395.00,90.10);
  \draw[figrule] (287.10,22.60) -- (395.00,22.60);
  \draw[figrule] (284.60,22.60) -- (287.10,22.60);
  \node[ticklbl,anchor=base east] at (282.60,19.75) {0.3};
  \draw[figrule] (284.60,32.24) -- (287.10,32.24);
  \node[ticklbl,anchor=base east] at (282.60,29.39) {0.4};
  \draw[figrule] (284.60,41.89) -- (287.10,41.89);
  \node[ticklbl,anchor=base east] at (282.60,39.04) {0.5};
  \draw[figrule] (284.60,51.53) -- (287.10,51.53);
  \node[ticklbl,anchor=base east] at (282.60,48.68) {0.6};
  \draw[figrule] (284.60,61.17) -- (287.10,61.17);
  \node[ticklbl,anchor=base east] at (282.60,58.32) {0.7};
  \draw[figrule] (284.60,70.81) -- (287.10,70.81);
  \node[ticklbl,anchor=base east] at (282.60,67.96) {0.8};
  \draw[figrule] (284.60,80.46) -- (287.10,80.46);
  \node[ticklbl,anchor=base east] at (282.60,77.61) {0.9};
  \draw[figrule] (284.60,90.10) -- (287.10,90.10);
  \node[ticklbl,anchor=base east] at (282.60,87.25) {1.0};
  \draw[figrule] (291.60,22.60) -- (291.60,20.10);
  \node[ticklbl,anchor=base] at (291.60,12.40) {1};
  \draw[figrule] (321.37,22.60) -- (321.37,20.10);
  \node[ticklbl,anchor=base] at (321.37,12.40) {2};
  \draw[figrule] (338.79,22.60) -- (338.79,20.10);
  \node[ticklbl,anchor=base] at (338.79,12.40) {3};
  \draw[figrule] (360.73,22.60) -- (360.73,20.10);
  \node[ticklbl,anchor=base] at (360.73,12.40) {5};
  \draw[figrule] (390.50,22.60) -- (390.50,20.10);
  \node[ticklbl,anchor=base] at (390.50,12.40) {10};
  \draw[figlevel] (291.60,57.60) -- (395.00,57.60);  %
  \draw[figseries] (291.60,78.14) -- (321.37,72.16) -- (338.79,69.66) -- (360.73,64.45) -- (390.50,55.00);
  \draw[figours] (291.60,57.60) -- (321.37,47.86) -- (338.79,44.78) -- (360.73,37.16) -- (390.50,29.16);
  \draw[figmark] (291.60,78.14) circle (1.70pt);  %
  \draw[figmark] (321.37,72.16) circle (1.70pt);  %
  \draw[figmark] (338.79,69.66) circle (1.70pt);  %
  \draw[figmark] (360.73,64.45) circle (1.70pt);  %
  \draw[figmark] (390.50,55.00) circle (1.70pt);  %
  \draw[figmarkours] (290.05,56.05) rectangle (293.15,59.15);  %
  \draw[figmarkours] (319.82,46.31) rectangle (322.92,49.41);  %
  \draw[figmarkours] (337.24,43.23) rectangle (340.34,46.33);  %
  \draw[figmarkours] (359.18,35.61) rectangle (362.28,38.71);  %
  \draw[figmarkours] (388.95,27.61) rectangle (392.05,30.71);  %
  \draw[figmarkopen] (382.29,57.60) circle (2.00pt);  %
  \node[crossinglbl,anchor=base east] at (395.00,43.86) {crossing};
  \node[paneltitle,anchor=base] at (341.05,104.30) {(c) response ranking (5)};
  \node[panelmetric,anchor=base] at (341.05,95.10) {pair flip~$\downarrow$};
  \draw[figseries] (79.38,117.24) -- (87.38,117.24);
  \draw[figmark] (83.38,117.24) circle (1.70pt);
  \node[legendlbl,anchor=base west] at (89.88,114.30) {Batched self-consistency};
  \draw[figours] (192.36,117.24) -- (200.36,117.24);
  \draw[figmarkours] (194.81,115.69) rectangle (197.91,118.79);
  \node[legendlbl,anchor=base west] at (202.86,114.30) {OC-SFT};
  \draw[figlevel] (246.63,117.24) -- (254.63,117.24);
  \node[legendlbl,anchor=base west] at (257.13,114.30) {OC-SFT at $K{=}1$};
\end{tikzpicture}%
\caption{\textbf{OC-SFT at $K{=}1$ beats BSC at $K{=}10$ permutations on two of the three consumers.}
Batched self-consistency (BSC, grey; \citealp{korikov2025}) reaches OC-SFT's own $K{=}1$ (dashed) only on (c), near $K{=}8$; averaging keeps improving OC-SFT itself.
BSC's $K{=}10$ is its \Cref{tab:levels} row.
}
\label{fig:amortconsumers}
\end{figure}

\vspace{-4pt}

Permutation augmentation, which trains on the same shuffled views as OC-SFT but omits its explicit consistency penalty, closes about two thirds of the $\tau$-PSI gap, as does DebiasFirst.
A generic consistency mechanism closes as much of that gap and still fails the consumer: an EMA mean-teacher closes 71\% of it while costing the reader 0.023 exact match (\Cref{app:mitigations,app:consumers} for full results).
The penalty closes the remaining third; shuffled-view training alone does not match OC-SFT's stability.

\textbf{The designs that do reach the decisions pay for it at every query or every label.}
A self-consistency ensemble repeats the serving pass on every query \citep{korikov2025}, and at $K{=}10$ permutations it does not reach OC-SFT at $K{=}1$ on two of the three consumers (\Cref{fig:amortconsumers}).
Averaging the trained scorer improves its reproducibility too, so training and averaging compose rather than substitute.
Order-averaged distillation pays in the labels instead, needing $T{=}10$ teacher passes against OC-SFT's one, and is less stable on 30K labeled queries than OC-SFT on 1K.
A stronger teacher buys quality but not stability, so the gain comes from the penalty rather than the supervision (\Cref{app:mitigations,app:fullresults}).

\FloatBarrier

\vspace{-2pt} 
\subsection{Serving width and the choice of design}
\label{sec:copresence}
\vspace{-1pt}

\textbf{The penalty suppresses order dependence without sacrificing shared context where it helps.}
Serving one checkpoint at its training width rather than at $B{=}1$ gains 0.043 nDCG@10 on multi-document QA.
On passage relevance at $B{=}20$ the width effect is indistinguishable from zero, and on response ranking the gain is largest on the PPE collections while RM-Bench loses (\Cref{app:tasks,app:width}).

\textbf{Which design to serve follows from the serving budget and the task; the shared prompt earns its place in the labels either way.}
Served one candidate at a time, where both designs are order-free, a pointwise student trained for that width matches the batched checkpoint on passage relevance, 0.466 against 0.459 on an interval covering zero.
One batched checkpoint holds quality across 1--30 candidates; the pointwise student degrades at 20.
The labels allow no such choice: batched-teacher labels beat pointwise ones at both training widths, by $0.029$ nDCG@10 at $B_{\mathrm{train}}{=}1$ and $0.010$ at $B_{\mathrm{train}}{=}20$, both evaluated at $B{=}1$.
At $B{=}20$, batching holds that quality at 45\% lower per-query latency.
Detailed comparisons are in \Cref{app:mitigations,app:width}.

\FloatBarrier

\setcounter{topnumber}{2}
\setcounter{totalnumber}{3}

\vspace{-2pt}
\section{Conclusion}
\label{sec:conclusion}
\vspace{-1pt}

Scorers a ranking metric cannot separate can still differ in what their consumers decide, so we compared remedies by where their cost falls: at serving, in teacher-label construction, or once in training.
Fixes applied before training reached no decision: round-robin partitioning improved ranking quality without making any of the three decisions measurably more reproducible, while logit calibration improved neither quality nor retained-set overlap. \looseness=-1
The order channel resisted both; OC-SFT, a consistency penalty over shuffled views, attenuated it in the weights from one teacher pass and one serving permutation, while costing no ranking quality and making all three decisions more reproducible.
It also lowered order instability on all of the 12 bases we trained.
Telling quality-matched scorers apart therefore takes another measurement, and a stability metric's permutations are enough to run it on.

\label{body:end} %

\ifpreprint\vspace{-6pt}\else\vspace{-4pt}\fi
\section*{Ethics Statement}
\ifpreprint\vspace{-6pt}\else\vspace{-4pt}\fi

This work studies the reliability of LLM scorers rather than a new capability, since the same candidate set producing different rankings under reordering is a reproducibility concern in high-stakes retrieval such as legal search.
We train on MS MARCO, a public web-search dataset, and evaluate on public TREC-DL and BEIR collections with two proprietary legal collections from a commercial legal search provider, reported as Legal-A and Legal-B.
Both contain professionally authored content and attorney relevance judgments.
We collected no new annotations and release no raw queries, documents, or annotator identifiers from them.
They were evaluated under institutional data-governance controls.
TREC-NEWS and Robust04 are distributed and reported under NIST terms.
One closed model, GPT-5.4, appears as a comparison scorer and teacher, accessed through its provider's API under the terms of service.
Its weights and training data are not inspectable and the provider can retire the served version, so the GPT-5.4 comparisons are less reproducible than the rest.
The method requires no closed model, because the default teacher is the student's own base model and every base we train is open-weight.
We evaluate closed teachers only as a comparison, and an open teacher recovers most of their advantage without GPT-5.4 labels (\Cref{app:crossbase}).

\ifpreprint\vspace{-6pt}\else\vspace{-4pt}\fi
\section*{Reproducibility Statement}
\ifpreprint\vspace{-6pt}\else\vspace{-4pt}\fi

\Cref{app:readout} gives the scoring readout and reranking prompt template, including the scored answer skeleton.
\Cref{app:silver} covers silver-label construction and the consistency-weight selection protocol.
\Cref{sec:setup} defines $\tau$-PSI and states the permutation budget and the resampling unit for intervals, with the evaluation collections in \Cref{app:metrics}.
\Cref{app:repro} lists the shared training configuration and principal task-specific settings: base model, LoRA settings, optimizer and schedule, training width, consistency weight, selection rule, and hardware.
Per-collection reranking results are in \Cref{app:fullresults}, with the per-task, fixed-width and perturbation-specific breakdowns in \Cref{app:tasks,app:width,app:mechanism}.

We release the $\tau$-PSI implementation and the evaluation harness, the scoring prompt and scored skeleton of \Cref{app:readout}, and the training and evaluation code, including the inference script.
We also release the silver labels behind the primary result, the LoRA adapters for the 11 dense-base students of \Cref{sec:setup}, and the full evaluation configurations and their commands.
The repository is at \codeurlfull.
The full public evaluation suite runs in 60--90 hours on a single L40S-48GB GPU, and one collection such as DL19 in about an hour.
Training the student behind the primary result takes about 90 GPU-hours on one 8-GPU A100-40GB node from the released silver labels.
The full study, including unreported runs, totals roughly 100{,}000 GPU-hours.

Legal-A and Legal-B are proprietary, so the 18-collection reranking means in \Cref{tab:levels} and \Cref{sec:results} include two collections others cannot score; \Cref{app:fullresults} reports the sixteen-collection public mean alongside every per-collection cell.
We accessed GPT-5.4 through an Azure OpenAI \texttt{gpt-5.4} deployment, served by model version \texttt{gpt-5.4-2026-03-05}.
Evaluation calls ran between June and August 2026, and its teacher labels for \Cref{app:crossbase} between May and June.
Temperature and top-$p$ are not configurable on this interface, so provider defaults applied.
Rescoring byte-identical prefixes in separate inference batches does not return the same number, so a reader recomputing a score-level figure from the released outputs should not expect an exact match (\Cref{app:width}).

\ifpreprint\vspace{-6pt}\else\vspace{-4pt}\fi
\section*{AI use statement} 
\ifpreprint\vspace{-6pt}\else\vspace{-4pt}\fi

Large language models are an object of study and a component of the method in this work.
This covers the scorers we train and evaluate, the teachers that generate the silver relevance labels, and the prompted closed-model baselines, all described in \Cref{sec:method} and \Cref{sec:setup}.
They are studied here rather than used to produce the work, and the silver-label generation of \Cref{app:silver} is part of the method rather than a synthetic-data tool use.

We used generative AI tools for implementing software and preparing data, for literature search, brainstorming and interpreting results, for drafting the manuscript and editing it for length and readability, and for formatting references.
We verified references against primary sources rather than relying on model output, and we checked numeric claims in the text against the floats they cite.
We did not use generative AI tools to propose or refine the hypotheses or to formulate the decomposition of \Cref{sec:phenomenon}.
The authors reviewed all AI-assisted work, made all scientific decisions, and take responsibility for the final content, including text, claims and artifacts produced with generative AI.

\ifpreprint
\vspace*{-10pt}
\subsubsection*{Acknowledgments}
\vspace{-4pt}
We are grateful to Joel Stremmel, Mahtab Farrokh, Ashkan Alinejad and Dezhao Song for their feedback on the project and on earlier drafts of this manuscript.
\vspace{-8pt}
\fi

\ifpreprint\addtolength{\bibsep}{-1.2pt}\fi
\bibliography{references}
\bibliographystyle{iclr2027_conference}

\appendix
\crefalias{section}{appendix}    %
\crefalias{subsection}{appendix} %
\crefalias{subsubsection}{appendix} %

The appendix is organized as follows, each answering a different question:

\begin{description}[leftmargin=0pt,labelsep=0.5em,itemsep=1pt,parsep=0pt,topsep=3pt]
\item[\Cref{app:limitations}] the scope and limitations of the study.
\item[\Cref{app:setup}] how a score is read from the prompt, and what reproduces a run.
\item[\Cref{app:mechanisms}] the origin of the order effect, and controls that exclude other explanations.
\item[\Cref{app:mitigations}] whether cheaper alternatives suffice, what they cost, and first-stage robustness.
\item[\Cref{app:tasks}] whether the effect is specific to reranking, on QA and on response ranking.
\item[\Cref{app:consumers}] four consumers of a ranking, and what training changes for each.
\item[\Cref{app:width}] how many candidates to serve, and what width costs in quality and latency.
\item[\Cref{app:fullresults}] per-collection results, alternative first stages, and external rerankers.
\end{description}

Unless a caption says otherwise, every table and figure below uses Qwen3-4B at $B=20$ on the 18 reranking collections at seed 42, over $M=10$ random permutations.
A basis line states where a table or figure departs from that.

\section{Scope and limitations}
\label{app:limitations}

The serving pass takes its order from the first stage, so the permutations we measure under are not the ones it runs on.
Controls on the first-stage order leave both $\tau$-PSI and the output's correlation with BM25 nearly unchanged, so the measurement is representative of the served condition (\Cref{tab:mechanism-controls}).
They are also uniform draws, so the spreads we report describe typical rather than worst-case presentations.
The threshold, the reader and the preference model are frozen rather than refitted to each scorer's output.
A co-adapted consumer could absorb some of the instability we measure, so the reductions we report may be smaller in a fully retrained pipeline.
We measure the chosen/rejected pairs supplied to preference training, not the effect of fitting a downstream model to those labels; whether the instability survives downstream optimization remains untested.

Round-robin partitioning and logit calibration are two prompt-time constructions, not all possible ones, and both run under the single answer skeleton of \Cref{app:placeholderchoice}.
InvariRank \citep{bito2026} restricts cross-candidate attention to obtain one-pass order invariance; \Cref{app:defensive} ports the same mechanism from \citet{wang2025pine} and finds it costs both quality and stability at this readout, so built-in invariance is unavailable here rather than untried.
PA-GRPO \citep{zheng2026} rewards consistent decisions with reinforcement learning, and its reward is defined on the decision rather than on a score.
Redefining it over the continuous score our objective targets would change the method rather than merely porting it. %

The objective targets stability and the part of quality that order-averaging recovers, not raw quality: a pointwise $B=1$ scorer trained in our own pipeline matches our students on nDCG@10.
Heavily supervised commercial rerankers exceed them, on far larger relevance supervision than ours.
The case for scoring candidates together ($B>1$) does not cover pools of two: no collection we evaluate has fewer than four candidates (\Cref{tab:datasets}).
The one recurring collection-level cost is ArguAna, the suite's only counterargument-retrieval collection (\Cref{tab:grids-ndcg}).
Rescoring one permutation does not return identical grades for any scorer here: ours reproduce 0.985 of the retained set at fixed order, and GPT-5.4's floor is unmeasured (\Cref{tab:width-controls}).
Its overlap is therefore lowered by noise as well as by order, which flatters our comparison.
Nearly half its candidate pairs tie, so its reported figures turn on the tie-break: it reaches 0.468 nDCG@10 under the order-independent rule we keep against 0.511 under input-order ties.
That rule also gives it a lower pair-flip rate than OC-SFT, which reverses under the common stable sort.
The $\tau$-PSI we withhold would favour it too: at 0.118 it would rank second in the reranking column (\Cref{tab:levels}), and about sixth once tied pairs are discounted.
The convention we report is the one that favours the baseline.

Finally, the scorer uses no chain-of-thought \citep{lu2025, jedidi2025, weller2025}, so reasoning-intensive retrieval (BRIGHT, \citealp{su2025}) and reasoning rerankers in their own regime are untested.
All collections are English, and long-context corpora that overflow the $B=20$ prompt fall outside the geometry studied here (\Cref{app:readout}).

\FloatBarrier
\section{Setup: readout, evaluation, and configuration}
\label{app:setup}

We specify how a score is read out of a single prompt, including the placeholder and answer-skeleton choices it depends on (\Cref{app:readout}), which collections carry each measurement (\Cref{app:metrics}), the consolidated training and evaluation configuration (\Cref{app:repro}), and the interval and bolding protocol for \Cref{tab:levels} (\Cref{app:inference}).

\subsection{Score readout and prompt}
\label{app:readout}

The reranker scores a query against $B$ candidate documents in one prompt.
After a short grading instruction, the prompt places the query and the $B=20$ documents, each tagged \texttt{[i]}, then appends a fixed \texttt{Grades:} skeleton carrying one placeholder grade per slot.
Documents are truncated to 1200 characters on TREC Deep Learning and non-reranking corpora, and 500 on the BEIR and legal collections.
We score the fixed \texttt{Grades:} skeleton in one forward pass, reading four grade-token probabilities per candidate rather than generating grades (\Cref{sec:readout}).
No token is sampled, so a score is determined by the weights and the prompt, and no decoding temperature or generation seed enters.
Candidate order is the variation we report; with order held fixed, engine-call batching still moves retained-set overlap, but far less than reordering does (\Cref{tab:width-controls}).
GPT-5.4 is the one exception, and its decoding parameters are not configurable.
The scoring client's nominal temperature (1.0) and top-$p$ (10$^{-8}$) do not reach the provider.
The interface omits temperature, the client does not forward top-$p$, and no generation seed is exposed; the run seeds shuffle documents rather than generation.
FIRST \citep{reddy2024} also uses logits instead of decoding, but takes a listwise ranking from the first generated identifier, whereas we take one score per candidate.
The expectation also keeps scores near-continuous, so ties rarely decide the ranking.
At $B=20$, 99.6\% of scores within a query and permutation are distinct, and re-ranking the saved scores under an order-independent tie-break on document id moves every $\tau$-PSI value we report by less than 0.0005. The full prompt is shown below.

\begin{lstlisting}
System: You are a search relevance grader. For each numbered document, output one fixed
        integer relevance grade in {0, 1, 2, 3}. Do not rank the documents and do not explain.

User:   Instruction: Given a web search query, retrieve relevant passages that answer the query
        Query: <q>

        Documents:
        [1] <doc 1 text>            # whitespace-normalized + truncated to 1200 chars
         ...
        [20] <doc 20 text>

        Evaluate every document independently for relevance to the query.
        Assign exactly one integer relevance grade to every document:
        - 3 = directly and completely answers the query
        - 2 = strongly relevant but incomplete
        - 1 = weakly relevant or topical background
        - 0 = not relevant
        Output one line per document in the form: [<id>] Grade: <0|1|2|3>

Assistant:  Grades:                 # not sampled; a fixed skeleton is scored
        [1] Grade: 0                #   read P(0),P(1),P(2),P(3) at this slot
         ...                        #   score = E[g]/3
        [20] Grade: 0
\end{lstlisting}

The readout and the placeholder are both free choices, so we selected each once on the off-the-shelf Qwen3-4B scorer and held them fixed thereafter.
\Cref{tab:readout} reports both comparisons, comparing the argmax grade, $P(g \geq 2)$ and the expected grade, all three read off the same saved grade distribution.
The argmax grade is lower than either probability readout on the aggregate grid.
The two probability readouts are nearly level; $P(g \geq 2)$ leads by less than 0.001 nDCG@10, so we retain the expected grade as the graded readout and hold it fixed.

\begin{table}[htbp]
\centering
\small
\caption{\textbf{Score-readout and answer-skeleton placeholder choices.}
Basis: off-the-shelf Qwen3-4B over the 18 reranking collections.
Quality is one-permutation nDCG@10; instability is $\tau$-PSI at $B=20$ over $M=10$ random permutations.
Higher quality and lower instability are better; the final two columns count collections with lower / higher instability than \texttt{Grade: 0}.
Bold marks the selected protocol choice rather than the per-column optimum.}
\label{tab:readout}
\setlength{\tabcolsep}{4pt}
\begin{tabular}{lccrrr}
\toprule
& \multicolumn{2}{c}{Quality, one permutation} & \multicolumn{3}{c}{Order instability ($M=10$)} \\
\cmidrule(lr){2-3} \cmidrule(lr){4-6}
Choice & nDCG@10 $\uparrow$ & Wins & $\tau$-PSI $\downarrow$ & Lower & Higher \\
\midrule
\multicolumn{6}{l}{\emph{Score readout}, quality only} \\
argmax grade & 0.3365 & 1 / 18 & --- & --- & --- \\
$P(g \geq 2)$ & 0.3701 & 9 / 18 & --- & --- & --- \\
expected grade & \textbf{0.3694} & \textbf{8 / 18} & --- & --- & --- \\
\midrule
\multicolumn{6}{l}{\emph{Answer skeleton placeholder}} \\
\texttt{Grade: 0} & \textbf{0.3694} & \textbf{17 / 18} & \textbf{0.2983} & ref. & ref. \\
\texttt{Grade: 1} & 0.3541 & 0 / 18 & 0.2928 & 11 & 7 \\
\texttt{Grade: 2} & 0.3413 & 1 / 18 & 0.3348 & 6 & 12 \\
\texttt{Grade: 3} & 0.2013 & 0 / 18 & 0.3857 & 0 & 18 \\
\bottomrule
\end{tabular}
\end{table}

\subsubsection{Answer-skeleton placeholder}
\label{app:placeholderchoice}

Varying only the in-format placeholder from \texttt{Grade: 0} through \texttt{Grade: 3}, quality decreases monotonically from the selected \texttt{Grade: 0} (\Cref{tab:readout}).
\texttt{Grade: 1} has the lowest instability point estimate but lower quality, while \texttt{Grade: 2} and \texttt{Grade: 3} worsen both axes.
No in-format placeholder improves both quality and instability over the selected \texttt{Grade: 0}, so placeholder choice is not a substitute for  training.

\subsection{Evaluation collections}
\label{app:metrics}

The eleven public out-domain collections are the BEIR subset introduced for listwise LLM reranking by \citet{sun2023} and carried forward by the distilled listwise rerankers that followed \citep{pradeep2023, pradeep2023vicuna}.
That subset includes the three collections later work often drops: Signal-1M, TREC-NEWS and Robust04.
Those papers report DL19 and DL20 on MS MARCO v1; we add the three TREC DL v2 tracks, DL21 through DL23, so the sixteen extend that suite rather than merely reuse it.
Legal-A and Legal-B are proprietary collections from a commercial legal search provider, holding professionally authored legal content with relevance judgments made by attorneys.
Their candidate lists come with the internal fixtures rather than being retrieved by BM25.

\begin{table}[htbp]
\centering
\caption{\textbf{Evaluation collections across the three tasks.}
Counts are evaluated queries or prompts after filtering; protocols are in \Cref{sec:setup}.
}
\label{tab:datasets}
\scriptsize
\setlength{\tabcolsep}{3pt}
\begin{tabular*}{\textwidth}{@{}p{2.5cm}p{2.2cm}p{2.8cm}>{\raggedleft\arraybackslash}p{2.3cm}@{\extracolsep{\fill}}>{\raggedright\arraybackslash}p{3.2cm}@{}}
\toprule
Task / group & Collection & Domain / role & Queries / prompts & Candidate pool / protocol \\
\midrule
Reranking: in-domain & DL19 & web passage; TREC DL & 43 & BM25 top 100; graded judgments \\
 & DL20 & web passage; TREC DL & 54 & BM25 top 100; graded judgments \\
 & DL21 & web passage; TREC DL v2 & 53 & BM25 top 100; graded judgments \\
 & DL22 & web passage; TREC DL v2 & 76 & BM25 top 100; graded judgments \\
 & DL23 & web passage; TREC DL v2 & 82 & BM25 top 100; graded judgments \\
\midrule
Reranking: public OOD & NFCorpus & biomedical & 308 & BM25 top 100 \\
 & FiQA & financial QA & 648 & BM25 top 100 \\
 & Touch\'e-2020 & argument retrieval & 49 & BM25 top 100 \\
 & ArguAna & counterargument retrieval & 1{,}406 & BM25 top 100 \\
 & Climate-FEVER & climate fact-checking & 1{,}535 (1{,}381 with verdict labels) & BM25 top 100 \\
 & TREC-COVID & biomedical & 50 & BM25 top 100 \\
 & DBPedia & entity retrieval & 400 & BM25 top 100 \\
 & SciFact & scientific fact-checking & 300 & BM25 top 100 \\
 & Signal-1M & tweet / news & 97 & BM25 top 100 \\
 & TREC-NEWS & news & 57 & BM25 top 100 \\
 & Robust04 & news & 249 & BM25 top 100 \\
\midrule
Reranking: legal & Legal-A & legal & 97 & precomputed top 100 \\
 & Legal-B & legal & 175 & precomputed top 100 \\
\midrule
Multi-doc QA & HotpotQA & multi-hop QA; in-domain & 200 & 10 passages; 2 supporting \\
 & 2WikiMultiHopQA & multi-hop QA; zero-shot & 200 & 10 passages; 2 supporting \\
 & MuSiQue & multi-hop QA; zero-shot & 200 & 10 passages; 2 supporting \\
\midrule
Response ranking & RewardBench-2 & response quality & 1{,}763 & 4 responses \\
 & Nectar & response preference & 434 & 7 responses; 64 overlaps removed \\
 & PPE-MATH & math correctness & 512 & 8 responses \\
 & PPE MMLU-Pro & knowledge correctness & 512 & 8 responses \\
 & RM-Bench & adversarial preference & 1{,}327 & 6 responses \\
\bottomrule
\end{tabular*}
\end{table}

\FloatBarrier
\subsection{Consolidated configuration}
\label{app:repro}

Every task selects its consistency weight $\lambda$ and its checkpoint on that training task's own held-out split, using held-out ranking quality as the only criterion.
The grid is $\lambda \in \{0.5, 1, 2, 3, 4, 5\}$.
The grid's upper end is selected for passage reranking; the same rule selects interior values elsewhere, $\lambda{=}3$ for multi-document question answering and $\lambda{=}1$ for response ranking (\Cref{tab:taskconfig}), so the selection is not saturating by construction. We did not train above $\lambda{=}5$.
OC-SFT is the only variant with a consistency weight to select.
The alternative regularizers of \Cref{app:defensive}, permutation augmentation and DebiasFirst use fixed recipes and are reported at the checkpoint that maximizes the same held-out ranking-quality criterion.

\Cref{tab:taskconfig} gives what differs between the three tasks. The settings they share:

\begin{itemize}
\item LoRA \citep{hu2022}: rank 16, alpha 32, dropout 0.05, on the seven attention and MLP projections (\texttt{q}, \texttt{k}, \texttt{v}, \texttt{o}, \texttt{gate}, \texttt{up}, \texttt{down}) in every transformer layer.
\item Optimizer: AdamW \citep{loshchilov2019} ($\beta_1=0.9$, $\beta_2=0.95$, weight decay 0.01), learning rate 2e-4 with a cosine schedule and a 10\% linear warmup, 1 epoch; per-device batch 1 with gradient accumulation 8 across the 8-GPU node, an effective batch of 64; bf16 with FlashAttention-2.
\item OC-SFT only: the two views of a window are permutations drawn from seeds derived from the query and window identifiers. At one epoch, each window has a fixed pair of orders, and the anchored view is the first of the two rather than the first stage's own order. The loss uses the raw $0$--$3$ grade scale; the $[0,1]$ normalization is applied at scoring time, so $\lambda$ is on the grade scale.
\item OC-SFT only: $\lambda$ ramps linearly from 0 to its target over the first 500 optimizer steps, separately from the learning-rate warmup.
\item Training hardware: 8-GPU DDP nodes, with an A10G for the 1.7B student, A100-40GB for the 4B to 8B Qwen3 students and the small Gemma and Granite students, and A100-80GB for the 14B, 32B and 30B-class students.
\item Evaluation hardware: a L40S-48GB GPU for evaluation, scoring and for the $\tau$-PSI computation.\looseness=-1

\end{itemize}

\begin{table}[htbp]
\centering
\small
\caption{\textbf{What differs between the three tasks.} Every other setting is shared.}
\label{tab:taskconfig}
\setlength{\tabcolsep}{4pt}
\begin{tabular}{@{}l>{\raggedright\arraybackslash}p{0.25\linewidth}>{\raggedright\arraybackslash}p{0.23\linewidth}>{\raggedright\arraybackslash}p{0.23\linewidth}@{}}
\toprule
& Passage reranking & Multi-document QA & Response ranking \\
\midrule
Training data & MS MARCO, $\sim$30K queries and $\sim$3M document labels & HotpotQA distractor, 30K questions & UltraFeedback, $\sim$30.7K prompts \\
Training width & $B=20$ & $B=10$ & $B=4$ \\
Rubric & graded $0$--$3$ topical relevance & answer support, binary & response quality \\
Consistency weight & $\lambda=5$ & $\lambda=3$ & $\lambda=1$ \\
Selection split & held-out MS MARCO & disjoint 500 questions & held-out UltraFeedback \\
Quality metric & nDCG@10 & nDCG@10 & nDCG@1 \\
\bottomrule
\end{tabular}
\end{table}

\vspace{-6pt}
\subsection{Inference for \Cref{tab:levels}}
\label{app:inference}
\vspace{-4pt}
Intervals follow \Cref{sec:setup}'s protocol at $R{=}20{,}000$ resamples; seeds are averaged rather than resampled, since three cannot support a sampling model over training randomness.
Holm adjustment covers every contrast between OC-SFT and the other trained variants  and each has adjusted $p \le 0.0013$ with all seeds favorable.
\vspace{-8pt}

\FloatBarrier

\FloatBarrier
\section{Mechanism and stability controls}
\label{app:mechanisms}

We decompose order dependence into slot and window-index effects plus a residual consistent with companion coupling, test transfer to pool-membership changes and marker renaming (\Cref{app:mechanism}), rule out measurement artifacts (\Cref{app:validity}), and test robustness across metrics (\Cref{app:boundary}).

\subsection{Mechanisms and perturbation transfer}
\label{app:mechanism}

\begin{figure}[htbp]
\centering
\begin{subfigure}[t]{0.49\linewidth}
\includegraphics[width=0.84\linewidth]{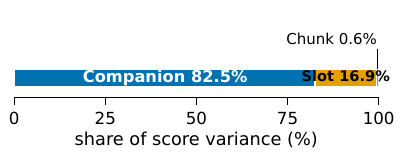}
\caption{Off-the-shelf within-document score-variance decomposition over 11 reranking collections.}
\end{subfigure}
\hfill
\begin{subfigure}[t]{0.49\linewidth}
\includegraphics[width=0.84\linewidth]{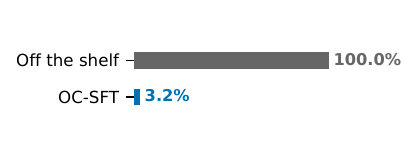}
\caption{Residual variance after slot and window-index controls, relative to off the shelf over four matched collections.}
\end{subfigure}

\par\vspace{-10pt}
\begin{subfigure}[t]{0.49\linewidth}
\includegraphics[width=0.84\linewidth]{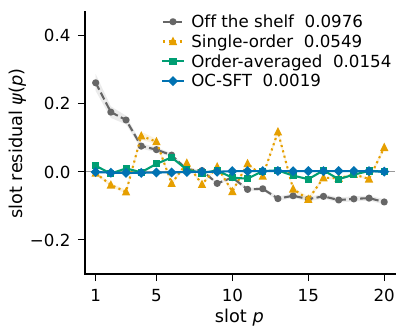}
\caption{Mean slot residual $\psi(p)$ over all documents; RMS is in the legend.}
\end{subfigure}
\hfill
\begin{subfigure}[t]{0.49\linewidth}
\includegraphics[width=0.84\linewidth]{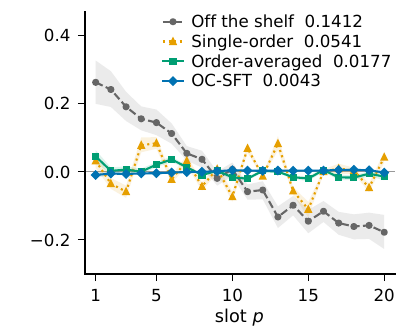}
\caption{Mean slot residual over gold-positive documents on the same scale.}
\end{subfigure}
\caption{\textbf{Residual variation consistent with companion coupling dominates off-the-shelf score variance; OC-SFT removes nearly all of it and flattens the slot residual.}
Basis: Qwen3-4B; (a--b) use eleven and four collections, while (c--d) use a separate held-out MS MARCO sample of 200 queries at 500 permutations.}
\label{fig:psiprofile}
\end{figure}
\vspace{-12pt}

In passage reranking, the first stage supplies the top 100 candidates, which are scored in five $B=20$ windows, so a document's grade can vary with its slot, window index and window companions (\Cref{fig:psiprofile}).
For each document, we subtract its mean score across presentations, removing its fixed score level.
The decomposition assigns separate shares of the remaining within-document variance to slot and window-index fixed effects; the residual, which is consistent with companion coupling but may include unmodelled interactions, accounts for more than four fifths of off-the-shelf per-document score variance.
OC-SFT removes 96.8\% of that residual variance.

Averaged over queries, each batch position $p$ has a mean order residual $\psi(p)$; under the cumulative \texttt{Grade:\,0} skeleton of \Cref{app:readout}, the off-the-shelf profile is monotone and early-over-scored.
This differs from the U-shaped ``lost in the middle'' effect in long-context QA and resembles the primacy pattern attributed to causal attention \citep{liu2024,tang2024,parry2024}.
OC-SFT flattens both the pooled and gold-only profiles.
An alternate in-format placeholder from the set compared in \Cref{app:readout} attenuates the off-the-shelf profile, so part of the slot effect belongs to the readout.
A forced-position nDCG control shows neither monotone primacy nor a U-shape on any collection, so the score-level slot residual does not become a ranking-level positional preference (\Cref{tab:mechanism-controls}).\looseness=-1

\textbf{Order robustness transfers to unseen pool-membership changes (\Cref{tab:pool-controls}).}
During training, every shuffled view contains the same candidate documents.
The loss never observes a candidate being replaced or removed.
To test transfer beyond that perturbation, we sample one document from the top 100 uniformly with a stable per-query seed and either replace it with the next first-stage candidate, appended at the end of the pool, or drop it, hold the serving order fixed and compare only the shared documents.
Pool-PSI applies the $\tau$-PSI transform $(1-\tau)/2$ to the original and perturbed rankings over those shared documents.
On all 15 collections, order-averaged distillation is between single-order distillation and OC-SFT on Pool-PSI, and the same aggregate ordering holds for top-10 churn; mean nDCG@10 remains level.
Averaging the teacher targets therefore provides part of the pool robustness, and the consistency penalty provides the remaining reduction.
The drop-only control changes Pool-PSI by less than 0.001 for any variant, so the reduction reflects changed companions rather than the introduction of rank 101.

\begin{table}[t]
\centering
\small
\caption{\textbf{Pool perturbations change decisions without changing mean quality.}
Lower is better except for signed mean $\Delta$nDCG@10.
Basis: Qwen3-4B, seed 42, 15 reranking collections, up to 100 judged queries each; rankings are restricted to the 99 documents shared by the two pools; the sampled rank, dropped document and replacement document are identical across the four columns for every query.
DL21--23 are excluded because their first-stage files stop at rank 100 and do not contain the replacement candidate at rank 101.}
\label{tab:pool-controls}
\begin{tabular}{lrrrr}
\toprule
Quantity & Off the shelf & Single-order & Order-avg. & OC-SFT \\
\midrule
Pool-PSI & 0.102 & 0.099 & 0.063 & \textbf{0.031} \\
Top-10 changed & 74.6\% & 65.0\% & 51.5\% & \textbf{36.3\%} \\
Mean $\Delta$nDCG@10 & $-0.001$ & $+0.001$ & $+0.001$ & $+0.001$ \\
Mean absolute $\Delta$nDCG@10 & 0.0356 & 0.0240 & 0.0153 & \textbf{0.0113} \\
\bottomrule
\end{tabular}
\end{table}

\enlargethispage{\baselineskip}
\textbf{Transfer depends on the perturbation: order-consistency training generalizes to changed pool membership but not to marker renaming (\Cref{tab:pool-controls,tab:mechanism-controls}).}
Candidate markers are arbitrary identifiers attached to the documents, so renaming them cannot change relevance but does move off-the-shelf scores.
Neither order-consistency training nor exposure to varied markers during SFT removes this sensitivity.
Adding a marker-consistency penalty, the same term with marker renamings substituted for order shuffles, lowers held-out marker $\tau$-PSI on all six held-out cells.

\textbf{The first-stage deficit follows window assignment rather than slot direction or copying (\Cref{tab:levels,tab:mechanism-controls}).}
Round-robin assignment improves the untrained scorer by distributing adjacent first-stage ranks across windows (\Cref{tab:levels}).
Reversing the candidate order does not repair the deficit: quality stays below both the forward and random-permutation rankings, and $\tau$-PSI is effectively unchanged.
Because BM25 supplies the first stage, the forward condition presents candidates in BM25 order; its output ranking is only marginally more correlated with BM25 than the output from a random permutation.
Together, these controls attribute the deficit to which candidates share a window.

\begin{table}[htbp]
\centering
\small
\caption{\textbf{Three controls separate order mechanisms and perturbation transfer.}
Block (a) tests whether the slot residual reaches ranking quality; block (b) tests transfer to marker renaming; block (c) tests the first-stage-order deficit.
Basis: blocks (a) and (c) use Qwen3-4B on the 18 reranking collections; block (b) uses Qwen3-1.7B and Qwen3-4B on DL19, NFCorpus, and Legal-A , with Roman-numeral and random three-character marker schemes held out from training.}
\label{tab:mechanism-controls}
\setlength{\tabcolsep}{3pt}
\begin{tabular*}{\textwidth}{l@{\extracolsep{\fill}}rr}
\toprule
\multicolumn{3}{l}{\emph{(a) Ranking-level slot preference}} \\
Diagnostic & Collections with primacy & Collections with U-shape \\
\midrule
Forced-position nDCG@10 & 0/18 & 0/18 \\
\bottomrule
\end{tabular*}
\vspace{1pt}

\begin{tabular*}{\textwidth}{l@{\extracolsep{\fill}}rrr}
\toprule
\multicolumn{4}{l}{\emph{(b) Marker-renaming transfer}} \\
Measure & Minimum & Maximum & Coverage / wins \\
\midrule
Held-out marker $\tau$-PSI & 0.035 & 0.059 & 6 cells \\
Penalty improvement over marker-varied SFT & 0.013 & 0.037 & 6/6 \\
\bottomrule
\end{tabular*}
\vspace{1pt}

\begin{tabular*}{\textwidth}{l@{\extracolsep{\fill}}rrr}
\toprule
\multicolumn{4}{l}{\emph{(c) First-stage-order controls}} \\
Measure & Reverse & Forward & Random permutation \\
\midrule
nDCG@10 & 0.347 & 0.370 & 0.415 \\
$\tau$-PSI & 0.2992 & 0.2983 & --- \\
Kendall correlation with BM25 & --- & 0.166 & 0.158 \\
\bottomrule
\end{tabular*}
\end{table}
\FloatBarrier

\subsection{Validity controls for stability}
\label{app:validity}
\vspace{-10pt}
\enlargethispage{\baselineskip}

\begin{table}[htbp]
\centering
\small
\caption{\textbf{Validity controls for the stability result.}
Block (a) separates stability from quality gain; block (b) tests score collapse and tied rankings; block (c) tests movement at the ranking head.
Blocks (b--c) use the 18 reranking collections.
Basis: Qwen3-4B, seed 42, $M=10$ random permutations for blocks (b--c), and the document-ID tie-break.}
\label{tab:stability-controls}
\setlength{\tabcolsep}{3pt}
\begin{tabular*}{\textwidth}{l@{\extracolsep{\fill}}lrrr}
\toprule
\multicolumn{5}{l}{\emph{(a) Quality controls}} \\
Diagnostic & Collection & Single-order & Order-avg. & OC-SFT \\
\midrule
$|\mathrm{corr}(\Delta\mathrm{nDCG@10},\text{ off-the-shelf }\tau\text{-PSI})|$ & DL19 & 0.257 & 0.194 & 0.196 \\
& DL20 & 0.070 & 0.037 & 0.106 \\
Queries improving over off the shelf & DL19 & 93.0\% & 97.7\% & 90.7\% \\
& DL20 & 90.7\% & 90.7\% & 88.9\% \\
Regressions relative to single-order & DL19 & reference & 37.2\% & 27.9\% \\
& DL20 & reference & 29.6\% & 24.1\% \\
\midrule
\multicolumn{5}{l}{\emph{(b) Score-geometry controls}} \\
Diagnostic & & Single-order & Order-avg. & OC-SFT \\
\midrule
Residual normalized by score dispersion & & 0.328 & 0.237 & 0.154 \\
Exact tied pairs & & 0.017\% & 0.050\% & 0.351\% \\
\bottomrule
\end{tabular*}
\vspace{1pt}

\begin{tabular*}{\textwidth}{l@{\extracolsep{\fill}}r}
\toprule
\multicolumn{2}{l}{\emph{(c1) Ranking-head levels}} \\
Variant & Top-10 instability \\
\midrule
Single-order & 0.343 \\
Order-avg. & 0.275 \\
OC-SFT & 0.193 \\
\bottomrule
\end{tabular*}
\vspace{1pt}

\begin{tabular*}{\textwidth}{l@{\extracolsep{\fill}}rrr}
\toprule
\multicolumn{4}{l}{\emph{(c2) Ranking-head paired contrasts}} \\
Contrast & Estimate & 95\% CI & Collection wins \\
\midrule
Single-order $-$ OC & $+0.150$ & [$+0.140$, $+0.159$] & OC 18/18 \\
Order-avg. $-$ OC & $+0.081$ & [$+0.074$, $+0.087$] & OC 18/18 \\
Single-order $-$ Order-avg. & $+0.069$ & [$+0.061$, $+0.076$] & Order-avg. 18/18 \\
\bottomrule
\end{tabular*}
\end{table}
\FloatBarrier

\textbf{Lower $\tau$-PSI reflects smaller permutation-induced movement rather than improved relevance, tied scores or tail-only changes (\Cref{tab:stability-controls}).}
Per-query quality gain against the off-the-shelf scorer is weakly related to starting instability and is non-negative for 89--98\% of queries across the three trained variants on DL19 and DL20.
Quality gains are aggregate: both order-averaged distillation and OC-SFT regress on some queries relative to single-order distillation.
Scale-normalized residuals fall monotonically across the three variants while ties remain rare, so score collapse does not account for lower instability.
Top-10 instability is $1-\mathrm{RBO}@10$ at persistence $p{=}0.9$ \citep{webber2010}, and it decreases in the same order on every collection.
The reduction includes the highest-ranked documents, including the top five passages passed to the downstream reader (\Cref{sec:setup}).

We do not report $\tau$-PSI for GPT-5.4, whose grades do not reproduce at a fixed order (\Cref{app:readout}).
Rescoring the same permutation, on one query from each of DL19, FiQA and Climate-FEVER, agrees on 0.963 of grades and leaves a same-order $\tau$-PSI floor of 0.023.
Sampling no token removes that source for the open-weight scorers, but not the effect: at fixed candidate order they still reproduce 0.985 of the retained set (\Cref{tab:width-controls}).

\subsection{Robustness across metrics}
\label{app:boundary}

\textbf{OC-SFT remains best across alternative quality and stability metrics (\Cref{tab:metrics}).}
The four columns test average quality, worst-permutation quality, dispersion across permutations and rank instability, so the ordering does not depend on $\tau$-PSI alone.
From the off-the-shelf scorer to OC-SFT, worst-permutation nDCG improves more than mean-permutation nDCG; because order-spread is mean minus worst, its reduction equals the difference between those gains.

\begin{table}[htbp]
\centering
\small
\caption{\textbf{The variant ordering holds across metrics.}
Columns report mean and worst nDCG, order-spread and $\tau$-PSI for four variants; higher nDCG and lower instability are better.
Basis: eleven dense bases by the 18 reranking collections and $M=10$ random permutations; nDCG uses the order-independent document-ID tie-break defined in \Cref{app:readout}; order-spread is mean minus worst before rounding.
}
\label{tab:metrics}
\begin{tabular*}{\textwidth}{l@{\extracolsep{\fill}}rrrr}
\toprule
Variant & mean-perm nDCG & worst-perm nDCG & order-spread & $\tau$-PSI \\
\midrule
Off the shelf & 0.4208 & 0.2940 & 0.1268 & 0.2874 \\
Single-order & 0.4606 & 0.3736 & 0.0870 & 0.2245 \\
Order-avg. & 0.4726 & 0.4076 & 0.0650 & 0.1388 \\
OC-SFT & \textbf{0.4771} & \textbf{0.4296} & \textbf{0.0475} & \textbf{0.0858} \\
\bottomrule
\end{tabular*}
\end{table}

\FloatBarrier
\section{Mitigation, supervision cost, and robustness}
\label{app:mitigations}

We re-measure the trained gain with the partition held fixed (\Cref{app:matchedgain}) and compare OC-SFT with mitigation baselines and architectural controls (\Cref{app:defensive}).
We then price its supervision and training cost (\Cref{app:silver}), weigh inference-time averaging against training for consistency (\Cref{app:kcurves}), measure sample efficiency (\Cref{app:sampleeff}), repeat the averaging comparison on consumers (\Cref{app:consumerk}), and check that the gain survives a different first stage (\Cref{app:firststage}).

\subsection{Partition controls and matched training gain}
\label{app:matchedgain}

\textbf{Round-robin partitioning does not explain the OC-SFT gain (\Cref{tab:partitioncontrols}).}
Round-robin raises nDCG@10 for both scorers but raises the off-the-shelf scorer more, reducing the training gain without eliminating it.
Across the six listed variants, the partition change moves $\tau$-PSI by at most 0.0053.
For the off-the-shelf scorer, neither retained-set overlap, verdict flip on either collection, nor pair flip changes measurably, with all four intervals covering zero.
\begin{table}[htbp]
\centering
\small
\caption{\textbf{Round-robin partitioning does not explain the trained gain.}
Block (a) gives first-stage-order single-pass nDCG@10 under each partition rule; training gain is OC-SFT minus off the shelf.
Block (b) gives round-robin minus contiguous $\tau$-PSI over $M=10$ random permutations.
Block (c) gives round-robin minus contiguous consumer output for the off-the-shelf scorer; every interval covers zero.
Basis: Qwen3-4B and 18 reranking collections; blocks (a--b) use seed 42 and equal-weight collection means, while block (c) uses $M=10$ permutations under each rule and the two claim-verification collections for verdict flip.
Block (a) intervals, and block (c)'s retained-set-overlap and selected-pair rows, are paired collection bootstraps with 20{,}000 resamples; the two verdict-flip rows are paired query bootstraps within SciFact and Climate-FEVER.}
\label{tab:partitioncontrols}
\label{tab:matchedgain}
\label{tab:partitiondecisions}
\begin{tabular*}{\textwidth}{l@{\extracolsep{\fill}}rrr}
\toprule
\multicolumn{4}{l}{\emph{(a) Training gain under fixed partitioning}} \\
Partition rule & Off the shelf & OC-SFT & Training gain [95\% CI] \\
\midrule
Contiguous & 0.370 & 0.459 & $+0.089$ [$+0.069$, $+0.110$] \\
Round-robin & 0.422 & 0.474 & $+0.052$ [$+0.034$, $+0.073$] \\
Contiguous minus round-robin & $-0.052$ & $-0.016$ & $+0.036$ [$+0.022$, $+0.050$] \\
\bottomrule
\end{tabular*}
\vspace{4pt}

\begin{tabular*}{\textwidth}{l@{\extracolsep{\fill}}r}
\toprule
\multicolumn{2}{l}{\emph{(b) Partition effect on instability}} \\
Variant & Round-robin minus contiguous $\tau$-PSI \\
\midrule
Off the shelf & $-0.0010$ \\
Single-order & $+0.0053$ \\
Order-avg. & $+0.0029$ \\
Perm.\ aug. & $+0.0005$ \\
DebiasFirst & $+0.0008$ \\
OC-SFT & $-0.0004$ \\
\bottomrule
\end{tabular*}
\vspace{4pt}

\begin{tabular*}{\textwidth}{l@{\extracolsep{\fill}}rr}
\toprule
\multicolumn{3}{l}{\emph{(c) Off-the-shelf consumer effect}} \\
Consumer & Change & 95\% CI \\
\midrule
Retained-set overlap & $+0.004$ & [$-0.007$, $+0.016$] \\
Verdict flip, SciFact & $+0.0147$ & [$-0.0102$, $+0.0404$] \\
Verdict flip, Climate-FEVER & $-0.0002$ & [$-0.0062$, $+0.0058$] \\
Selected pair & $+0.0004$ & [$-0.0033$, $+0.0034$] \\
\bottomrule
\end{tabular*}
\end{table}
\FloatBarrier

\subsection{Mitigation baselines and architectural controls}
\label{app:defensive}

\textbf{No mitigation baseline reaches OC-SFT's mean $\tau$-PSI over the 18 reranking collections (\Cref{tab:variants}).}
EMA closes 71\% of the stability difference between single-order distillation and OC-SFT; order-averaged distillation at $T{=}10$, permutation augmentation, and DebiasFirst close 63--64\%.
Averaging fewer teacher orders does not reach that: the $\tau$-PSI floor over $T{=}2$, $3$ and $5$ is 0.153, and the curve is not saturated at five, so $T{=}10$ is not an over-provisioned comparator.
LoRA dropout leaves $\tau$-PSI level with single-order distillation, KL regularization worsens it, and CapCal remains at the off-the-shelf level (\Cref{tab:levels}).
DebiasFirst and permutation augmentation differ only in propensity weighting and yield 0.128 and 0.129 $\tau$-PSI, so the weighting does not measurably help in our score-regression port.

\begin{table}[htbp]
\centering
\small
\caption{\textbf{Mitigation baselines and regularizers on the Qwen3-4B student.}
Higher nDCG@10 and lower $\tau$-PSI are better.
Each variant is the mean $\pm$ sample standard deviation.
Bold is a descriptive seed-spread heuristic, not the Holm-adjusted marking of \Cref{tab:levels}: a lead larger than the seed spread of the rows it leads.
The $\tau$-PSI column clears that heuristic, whereas the quality column does not, so quality is unmarked.
\% closed is the fraction of the stability difference between single-order distillation and OC-SFT that the method removes.
$T$ is the number of shuffled teacher orders averaged into each label; the $T{=}2$, $3$ and $5$ draws are the first $T$ orders of the $T{=}10$ pool rather than independent draws.
Basis: Qwen3-4B, 18 reranking collection means, 3 training seeds.}
\label{tab:variants}
\begin{tabular}{lrrr}
\toprule
Variant & Mean nDCG@10 & Mean $\tau$-PSI & \% closed \\
\midrule
Single-order ($T{=}1$) & $0.449 \pm 0.003$ & $0.209 \pm 0.008$ & 0\% \\
\quad + LoRA dropout 0.20 & $0.448 \pm 0.004$ & $0.208 \pm 0.001$ & $+0.8\%$ \\
\quad + KL-to-base $\beta=0.05$ & $0.447 \pm 0.002$ & $0.231 \pm 0.012$ & $-17.5\%$ \\
\quad + EMA mean-teacher $\alpha=0.5$ & $0.453 \pm 0.003$ & $0.119 \pm 0.003$ & 71.4\% \\
Order-avg.\ $T{=}2$ & $0.451 \pm 0.002$ & $0.176 \pm 0.008$ & 26.2\% \\
Order-avg.\ $T{=}3$ & $0.451 \pm 0.000$ & $0.153 \pm 0.002$ & 44.4\% \\
Order-avg.\ $T{=}5$ & $0.450 \pm 0.001$ & $0.161 \pm 0.009$ & 38.1\% \\
Order-avg.\ $T{=}10$ & $0.455 \pm 0.002$ & $0.130 \pm 0.007$ & 62.7\% \\
Perm.\ aug. & $0.455 \pm 0.002$ & $0.129 \pm 0.010$ & 63.5\% \\
DebiasFirst & $0.454 \pm 0.001$ & $0.128 \pm 0.005$ & 64.3\% \\
OC-SFT ($\lambda^\ast=5$) & $0.459 \pm 0.002$ & $\mathbf{0.083 \pm 0.002}$ & 100\% \\
\bottomrule
\end{tabular}
\end{table}

\textbf{PINE's patched attention breaks our multi-candidate slot-grade readout \citep{wang2025pine}.}
On DL19 at $B=20$, the ported scorer loses 0.122 nDCG@10 and adds 0.114 $\tau$-PSI relative to the base Qwen3-4B scorer.
A segmented prompt recovers most of the nDCG loss, and scoring one document at a time matches the base, so the failure arises from attention across documents rather than the readout alone.

\FloatBarrier
\subsection{Supervision and training cost}
\label{app:silver}

To separate information in the labels from exposure during student training, we cross teacher scoring width ($B=1$ or $20$) with student training width ($B=1$ or $20$).
All four variants use Qwen3-4B as teacher and student; \Cref{app:crossbase} varies teacher identity separately.

\textbf{Batched-teacher labels improve student nDCG@10 at both training widths, although the gain is smaller when the student also trains at $B=20$ (\Cref{tab:teacher-design}).}
The pointwise teacher ranks known MS MARCO positives higher under direct teacher metrics, yet its labels train a worse student.
Remapping the pointwise labels to the batched teacher's score distribution while preserving document order does not close the student gap.
Neither teacher-side ranking quality nor marginal score distribution therefore explains the benefit of batched-teacher labels.

\begin{table}[htbp]
\centering
\small
\caption{\textbf{Batched-teacher labels improve student nDCG@10 at both training widths.}
The upper block reports the difference between students trained on batched- and pointwise-teacher labels; positive values favor batched-teacher labels.
All students are evaluated at serving width $B=1$ in first-stage order.
Rank-quantile matching assigns the batched teacher's sorted score values to documents in the pointwise teacher's rank order.
The residual is the batched-label student's nDCG@10 minus that of the recalibrated pointwise-label student; minimum and maximum are over three evaluated checkpoints.
Basis: mean over the 18 reranking collections; Qwen3-4B; training seed 42; 100{,}000 paired collection-bootstrap samples.}
\label{tab:teacher-design}
\begin{tabular*}{\textwidth}{l@{\extracolsep{\fill}}rrr}
\toprule
Student training width & $\Delta$nDCG@10 & 95\% CI & Wins (of 18) \\
\midrule
$B_{\mathrm{train}}=1$ & $+0.0291$ & [$+0.0195$, $+0.0385$] & 17 \\
$B_{\mathrm{train}}=20$ & $+0.0097$ & [$+0.0023$, $+0.0169$] & 13 \\
\bottomrule
\end{tabular*}

\vspace{4pt}
\begin{tabular*}{\textwidth}{l@{\extracolsep{\fill}}rr}
\toprule
Control & Minimum residual & Maximum residual \\
\midrule
Rank-quantile matching & $+0.0307$ & $+0.0333$ \\
\bottomrule
\end{tabular*}
\end{table}

\textbf{Increasing student views beyond $N=2$ does not help, while additional teacher passes mainly improve stability in a plain-SFT control (\Cref{tab:view-controls}).}
From $N{=}2$--$8$, nDCG@10 and $\tau$-PSI remain within the ranges in \Cref{tab:view-controls} while forward-pass training cost rises fourfold.
On a plain-SFT student with the teacher held fixed, increasing teacher permutations from $T{=}5$ to $T{=}40$ changes nDCG@10 by $+0.0042$ and $\tau$-PSI by $-0.0398$ at 8$\times$ the label-generation cost.

\begin{table}[htbp]
\centering
\small
\caption{\textbf{Extra student views raise training cost; extra teacher passes mainly lower $\tau$-PSI.}
$N$ is the number of shuffled student views per training step; $T$ is the number of teacher permutations averaged into each stored label.
Student-view rows report the range over $N\in\{2,3,4,8\}$; teacher-label rows report the change from $T{=}5$ to $T{=}40$.
Cost is relative to $N{=}2$ training for student-view rows and $T{=}5$ labeling for teacher-label rows.
Basis: Qwen3-4B at the selected operating point; student-view rows are a five-collection mean over DL19, DL20, Touch\'e-2020, NFCorpus and Legal-A at a single seed, not the 18-collection default, so they are not comparable to the means in \Cref{tab:levels}.
}\label{tab:view-controls}
\setlength{\tabcolsep}{4pt}
\begin{tabular}{lllrr}
\toprule
Stage varied & Sweep & Metric & Observed range / change & Forward-pass cost \\
\midrule
Student training & $N\in\{2,3,4,8\}$ & nDCG@10 & range 0.4786--0.4795 & 100--400\% training \\
Student training & $N\in\{2,3,4,8\}$ & $\tau$-PSI & range 0.067--0.077 & 100--400\% training \\
Averaged silver, plain SFT & $T=5\rightarrow40$ & nDCG@10 & change $+0.0042$ & 800\% labeling \\
Averaged silver, plain SFT & $T=5\rightarrow40$ & $\tau$-PSI & change $-0.0398$ & 800\% labeling \\
\bottomrule
\end{tabular}
\end{table}

\textbf{Re-distillation helps in the first round and then largely saturates (\Cref{tab:redistill}).}
Qwen3-4B gains mostly in stability on the first round and changes by only 0.0011 on each metric in the second, with $\tau$-PSI increasing; Gemma-E4B retains a smaller second-round gain.
A second round therefore buys a change of that size for another full labeling and training cycle, so iterating further is not an efficient way to add stability.

\begin{table}[htbp]
\centering
\small
\caption{\textbf{Iterative self-distillation gains diminish after the first round.}
Positive $\Delta$nDCG@10 and negative $\Delta\tau$-PSI are improvements.
Round 1 is measured against the initial trained student and round 2 against round 1.
Basis: 18 reranking collections, one iterative training lineage per base.}
\label{tab:redistill}
\begin{tabular}{lrrr}
\toprule
Base & Round & $\Delta$nDCG@10 & $\Delta\tau$-PSI \\
\midrule
Qwen3-4B & 1 & $+0.0069$ & $-0.0183$ \\
Qwen3-4B & 2 & $+0.0011$ & $+0.0011$ \\
Gemma-E4B & 1 & $+0.0141$ & $-0.0311$ \\
Gemma-E4B & 2 & $+0.0036$ & $-0.0077$ \\
\bottomrule
\end{tabular}
\end{table}

\vspace{-1em}
\subsection{Inference-time amortization}
\label{app:kcurves}
\vspace{-1em}

\begin{figure}[htbp]
\centering
\includegraphics[width=\linewidth]{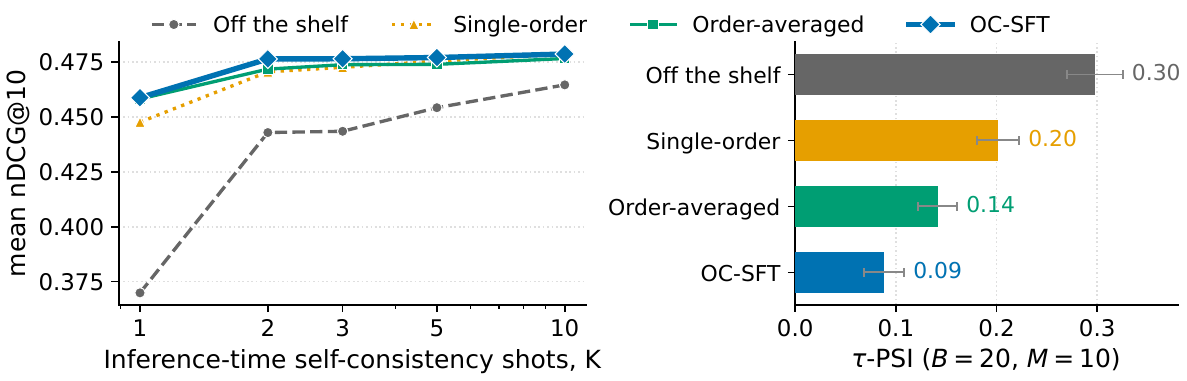}
\caption{\textbf{OC-SFT at $K=1$ is within 0.006 nDCG@10 of the off-the-shelf $K=10$ ensemble.}
Left: nDCG@10 against self-consistency permutations $K$, higher is better.
Right: single-pass $\tau$-PSI at $B=20$ for off the shelf, single-order distillation, order-averaged distillation and OC-SFT, evaluated over $M=10$ random permutations; lower is better.
}
\label{fig:amort}
\end{figure}

\textbf{Averaging raises nDCG@10, while training supplies most of the $\tau$-PSI reduction (\Cref{fig:amort}).}
At $K=5$, the off-the-shelf scorer still trails its $K=10$ nDCG@10, so we retain $K=10$ as the quality comparator.
OC-SFT has the lowest single-pass $\tau$-PSI among the four variants.
From $K=1$ to $K=10$, averaging raises off-the-shelf nDCG@10 by 0.095 but each trained variant by at most 0.035, so training has already captured most of the nDCG@10 gain that permutation averaging provides.

\FloatBarrier
\subsection{Sample efficiency}
\label{app:sampleeff}

\begin{figure}[htbp]
\centering
\begin{subfigure}{\linewidth}
\centering
\includegraphics[width=0.90\linewidth]{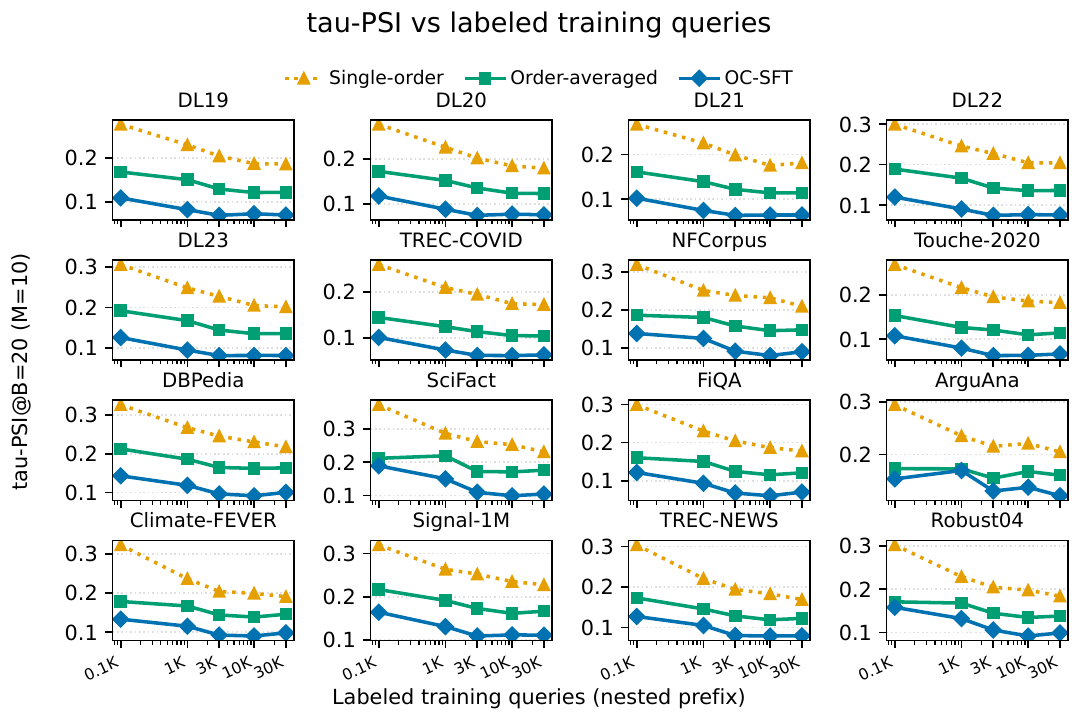}
\caption{$\tau$-PSI vs.\ labeled queries}
\end{subfigure}\\[0pt]
\begin{subfigure}{\linewidth}
\centering
\includegraphics[width=0.90\linewidth]{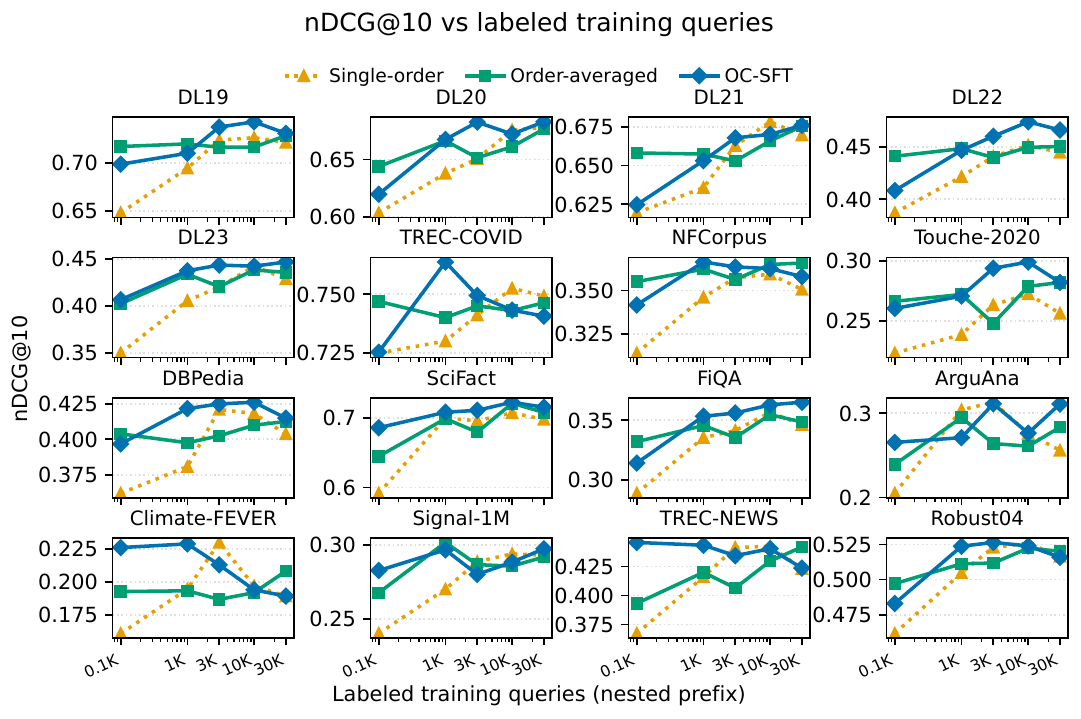}
\caption{nDCG@10 vs.\ labeled queries}
\end{subfigure}
\caption{\textbf{Stability saturates with fewer labeled queries than quality.}
(a) Student $\tau$-PSI, lower is better; (b) nDCG@10, higher is better, both against the number of unique labeled training queries over nested pools.
Basis: Qwen3-4B; 16 public reranking collections; Legal-A and Legal-B follow the same pattern.}
\label{fig:sampleeff}
\end{figure}

\textbf{$\tau$-PSI saturates with fewer labeled queries than nDCG@10 (\Cref{fig:sampleeff}).}
With 1K labeled queries, OC-SFT is more stable than order-averaged distillation trained on 30K queries on every collection except ArguAna.
For quality, models trained from one teacher order need about 3K queries to approach their 30K nDCG@10, whereas models trained from ten-order averages saturate earlier.

\vspace{-4pt}

\FloatBarrier
\subsection{Serving-time averaging after training}
\label{app:consumerk}

\Cref{app:kcurves} sweeps the serving budget for nDCG@10 and $\tau$-PSI.
We repeat that sweep on the three consumers, so that whether training removes what averaging buys is answered on the measures the decisions are taken from and not on ranking quality alone.

\textbf{Serving-time averaging improves all three consumer measures at least as much for OC-SFT as for the off-the-shelf scorer (\Cref{fig:amortconsumers}).}
From $K=1$ to $K=10$, the pair-flip reduction is larger after OC-SFT, and the answer-flip reduction is equal.
Under matched retention, OC-SFT also closes a larger share of the available overlap range.
OC-SFT leads at every measured budget on all three consumers and on every collection from $K=2$ onward, with a single tie on TREC-COVID at $K=1$.
OC-SFT reduces but does not eliminate presentation sensitivity.
Averaging therefore further improves reproducibility after training, but produces a smaller gain in label accuracy (\Cref{tab:labelcrossing}).

\textbf{On reranking, OC-SFT at $K{=10}$ exceeds its $K{=1}$ result, while BSC at $K{=10}$ still trails OC-SFT at $K{=1}$.}
At $K{=1}$, OC-SFT reaches 0.832 retained-set overlap against 0.663 for single-order distillation and 0.737 for order-averaged distillation; at $K{=10}$ the three reach 0.950, 0.869 and 0.904; all three-seed means.
OC-SFT therefore leads both baselines at matched $K$, although each baseline at $K{=10}$ exceeds OC-SFT at $K{=1}$.
\Cref{fig:amortconsumers} draws the same two variants on the queries they share, and at each $K$ it sets OC-SFT's cutoff to the fraction the off-the-shelf scorer's F1-tuned cutoff retains, so both are thresholded to equal set size.
We match them, as retained-set overlap rises with set size and the F1-tuned fraction itself moves with $K$.
The same two OC-SFT points are 0.831 and 0.943 there.
The tenfold comparison in \Cref{fig:amortconsumers} holds against the untrained ensemble of \citet{korikov2025} and does not extend to a trained one.

Pair flip counts how often the selected pair moves, not whether the movement crosses the gold label.
We therefore classify each query by whether its top-scored response agrees with gold under all $M=10$ random permutations, under none of them, or under some.

\begin{table}[htbp]
    \centering
    \small
    \caption{\textbf{Gold crossing and label accuracy on response ranking.}
    \emph{Gold crossing} is the share of queries whose top-scored response agrees with gold under some permutations and not others; \emph{1 perm.} and \emph{10 perm.} are label accuracy from a single permutation and from the ten-permutation average.
    Lower crossing and higher accuracy are better.
    The binary-gold mean excludes Nectar, whose gold is graded and whose off-the-shelf accuracy is near chance.
    Basis: $M=10$ random permutations; off the shelf is a fixed evaluation, single-order distillation uses seed 42, and OC-SFT is the mean over seeds 42--44.
    }
    \label{tab:labelcrossing}
    \scriptsize
    \setlength{\tabcolsep}{2pt}
    \begin{tabular*}{\textwidth}{l@{\extracolsep{\fill}}rrrrrrrrr}
    \toprule
    & \multicolumn{3}{c}{Off the shelf} & \multicolumn{3}{c}{Single-order} & \multicolumn{3}{c}{OC-SFT} \\
    \cmidrule(lr){2-4} \cmidrule(lr){5-7} \cmidrule(lr){8-10}
    Collection & Gold cross.$\downarrow$ & 1 perm. & 10 perm. & Gold cross.$\downarrow$ & 1 perm. & 10 perm. & Gold cross.$\downarrow$ & 1 perm. & 10 perm. \\
    \midrule
    RewardBench-2 & 0.661 & 0.594 & 0.711 & 0.656 & 0.594 & 0.684 & \textbf{0.295} & 0.683 & 0.705 \\
    Nectar & 0.714 & 0.225 & 0.295 & 0.698 & 0.261 & 0.300 & \textbf{0.597} & 0.276 & 0.318 \\
    PPE-MATH & 0.654 & 0.754 & 0.834 & 0.523 & 0.806 & 0.846 & \textbf{0.426} & 0.812 & 0.843 \\
    PPE MMLU-Pro & 0.727 & 0.662 & 0.691 & 0.770 & 0.663 & 0.668 & \textbf{0.593} & 0.667 & 0.677 \\
    RM-Bench & 0.576 & 0.709 & 0.711 & 0.564 & 0.707 & 0.725 & \textbf{0.496} & 0.712 & 0.722 \\
    \midrule
    Binary-gold mean & 0.654 & 0.680 & 0.737 & 0.628 & 0.693 & 0.731 & \textbf{0.453} & 0.719 & 0.737 \\
    \bottomrule
    \end{tabular*}
    \end{table}

\textbf{Most pair flips occur on queries that cross the gold label (\Cref{tab:labelcrossing}).}
Off the shelf, 0.654 of queries agree with gold under some permutations and not others, and 0.682 of all pair flips occur on those queries.
The $K=10$ average raises binary-gold mean label accuracy (\Cref{tab:labelcrossing}).

The gain is concentrated in RewardBench-2 and PPE-MATH, while averaging adds almost nothing on RM-Bench or PPE MMLU-Pro.
The mean therefore overstates how reliably averaging repairs a label.
OC-SFT lowers crossing on all four binary-gold collections, and single-order distillation lowers it on three and raises it on PPE MMLU-Pro.
The two trained variants are not matched on seeds, so we read the ordering rather than the size of the difference.
At one presentation, OC-SFT approaches the accuracy of the ten-presentation off-the-shelf scorer, but further averaging improves label accuracy less for OC-SFT than for the off-the-shelf scorer.
Averaging therefore adds little to label correctness once the scorer is trained, while on reproducibility it keeps composing (\Cref{fig:amortconsumers}).
Nectar stays out of the mean, since its gold is the maximum of graded scores rather than a binary preference and the off-the-shelf scorer agrees under all $M=10$ random permutations on 0.002 of its queries.
Its single-permutation accuracy of 0.225 is near what a random choice among its responses would reach.

\FloatBarrier
\subsection{Robustness to the first stage}
\label{app:firststage}

Because all main reranking results use BM25 candidates, we test first-stage dependence by evaluating the same BM25-trained Qwen3-4B checkpoints on BGE-base-en-v1.5 \citep{xiao2024} and SPLADE++ ED \citep{formal2022} candidates without retraining.

\textbf{OC-SFT retains the lowest $\tau$-PSI while remaining near order-averaged distillation in quality under all three first stages (\Cref{tab:firststage}).}
On the 13-collection mean, OC-SFT ties order-averaged distillation under BM25, trails it by 0.004 under BGE, and leads it by 0.002 under SPLADE.
OC-SFT also has the lowest per-collection $\tau$-PSI on all 13 collections (\Cref{tab:firststage-perdataset}).

\begin{table}[htbp]
\centering
\caption{\textbf{Four variants under three first stages.}
Each variant has separate nDCG@10 and $\tau$-PSI columns, averaged over the same 13 collections.
All trained variants use BM25 candidates during training, so the BGE and SPLADE rows test eval-only transfer.
DL21--23 and Legal-A/B are excluded because no BGE or SPLADE candidate sets were available for them.
Basis: Qwen3-4B, seed 42; BM25-trained checkpoints; per-collection results are in \Cref{tab:firststage-perdataset}.}
\label{tab:firststage}
\setlength{\tabcolsep}{3pt}
\begin{tabular*}{\textwidth}{l@{\extracolsep{\fill}}rrrrrrrr}
\toprule
& \multicolumn{2}{c}{Off the shelf} & \multicolumn{2}{c}{Single-order} & \multicolumn{2}{c}{Order-avg.} & \multicolumn{2}{c}{OC-SFT} \\
\cmidrule(lr){2-3}\cmidrule(lr){4-5}\cmidrule(lr){6-7}\cmidrule(lr){8-9}
First stage & nDCG & $\tau$-PSI & nDCG & $\tau$-PSI & nDCG & $\tau$-PSI & nDCG & $\tau$-PSI \\
\midrule
BM25 & 0.374 & 0.287 & 0.452 & 0.198 & 0.463 & 0.140 & 0.463 & 0.089 \\
BGE (dense) & 0.373 & 0.315 & 0.462 & 0.213 & 0.478 & 0.147 & 0.474 & 0.089 \\
SPLADE (learned-sparse) & 0.358 & 0.307 & 0.455 & 0.210 & 0.463 & 0.146 & 0.465 & 0.090 \\
\bottomrule
\end{tabular*}
\end{table}

\FloatBarrier
\section{Transfer to multi-document QA and response ranking}
\label{app:tasks}

We repeat the measurements outside passage reranking, on multi-document question answering (\Cref{app:qa}) and on response ranking (\Cref{app:respranking}), to test whether presentation dependence and the effect of training on it are specific to the reranking setting.
\Cref{tab:width} gives the serving-width effect for all three.

\begin{table}[htbp]
\centering
\small
\caption{\textbf{Serving-width effects differ by task and collection.}
Rows compare OC-SFT at $B=1$ and at the training width in parentheses.
Width effect is training-width quality minus $B=1$ quality.
One-permutation effects include the order residual, whereas order-averaged effects average the training-width scores over $K=10$ serving permutations.
Basis: Qwen3-4B, seeds 42--44.}
\label{tab:width}
\setlength{\tabcolsep}{1.5pt}
\begin{tabular}{llrrrr}
\toprule
& & \multicolumn{2}{c}{Level} & \multicolumn{2}{c}{Width effect} \\
\cmidrule(lr){3-4} \cmidrule(lr){5-6} 
Task & Metric & $B=1$ & Training width & One permutation & Order-averaged \\
\midrule
Reranking: 18 collections ($B=20$) & nDCG@10 & 0.459 & 0.459 & $+0.000$ & $+0.020$ \\
Multi-doc QA ($B=10$), 3 collections & nDCG@10 & 0.918 & 0.961 & $+0.043$ & $+0.047$ \\
\midrule
\addlinespace[2pt]
\multicolumn{6}{l}{\emph{Response ranking ($B=4$), 5 collections}} \\
\addlinespace[1pt]
\quad RewardBench-2 (4 candidates) & nDCG@1 & 0.670 & 0.690 & $+0.020$ & $+0.036$ \\
\quad Nectar (7 candidates) & nDCG@1 & 0.610 & 0.617 & $+0.006$ & $+0.026$ \\
\quad PPE-MATH (8 candidates) & nDCG@1 & 0.711 & 0.807 & $+0.096$ & $+0.132$ \\
\quad PPE MMLU-Pro (8 candidates) & nDCG@1 & 0.602 & 0.675 & $+0.074$ & $+0.076$ \\
\quad RM-Bench (6 candidates) & nDCG@1 & 0.755 & 0.715 & $-0.040$ & $-0.033$ \\
\bottomrule
\end{tabular}
\end{table}

\subsection{Multi-document question answering}
\label{app:qa}

\begin{table}[htbp]
\centering
\small
\caption{\textbf{Support ranking transfers zero-shot; reader accuracy depends on scorer context.}
Support columns report the OC-SFT scorer at $B=10$.
Reader columns report the same frozen Granite-4.1-8B reader over the top five passages from OC-SFT, a pointwise scorer, or a $B=1$ scorer trained on labels averaged over ten multi-context teacher orders.
Higher nDCG@10, EM and F1 are better; lower $\tau$-PSI is better.
All metrics are averaged over the same permutations.
Basis: Qwen3-4B, 200 questions per collection, seed 42, $M=10$ random permutations.}
\label{tab:qa-transfer-reader}
\setlength{\tabcolsep}{4pt}
\begin{tabular}{lrrrrrrrr}
\toprule
& \multicolumn{2}{c}{OC-SFT support} & \multicolumn{2}{c}{Reader: OC-SFT} & \multicolumn{2}{c}{Reader: pointwise} & \multicolumn{2}{c}{Reader: ten-order labels} \\
\cmidrule(lr){2-3} \cmidrule(lr){4-5} \cmidrule(lr){6-7} \cmidrule(lr){8-9}
Collection & nDCG@10 & $\tau$-PSI & EM & F1 & EM & F1 & EM & F1 \\
\midrule
HotpotQA & 0.964 & 0.082 & 0.621 & 0.743 & 0.575 & 0.684 & 0.623 & 0.734 \\
2WikiMultiHopQA & 0.975 & 0.102 & 0.431 & 0.546 & 0.427 & 0.532 & 0.398 & 0.524 \\
MuSiQue & 0.944 & 0.078 & 0.417 & 0.531 & 0.340 & 0.459 & 0.381 & 0.487 \\
\bottomrule
\end{tabular}
\end{table}

\textbf{Ten-order labels transfer part of the benefit of shared-context support selection, but not consistently across QA collections (\Cref{tab:qa-transfer-reader}).}
With the same frozen reader, a $B=1$ student trained on those labels matches the $B=10$ scorer on HotpotQA EM and recovers most of its F1 advantage, recovers only part of either gap on MuSiQue, and falls below the pointwise baseline on 2WikiMultiHopQA.
Teacher averaging can therefore distill some context benefit into a pointwise student, but it does not replace presenting the companion passages at serving time.

\begin{table}[htbp]
\centering
\small
\caption{\textbf{QA width effects are positive on all three collections and come from real companion passages.}
Block (a) gives the mean one-permutation and order-averaged width effects by collection.
Block (b) decomposes the one-permutation three-collection mean into informative-alternative, prompt-length/position and answer-skeleton terms.
Width effect is $B=10$ quality minus $B=1$ quality.
Basis: Qwen3-4B OC-SFT, three independently trained checkpoints; each served at both widths.}
\label{tab:qa-width}
\begin{tabular*}{\textwidth}{l@{\extracolsep{\fill}}rr}
\toprule
\multicolumn{3}{l}{\emph{(a) Mean width effect by collection}} \\
Collection & One permutation & Order-averaged \\
\midrule
HotpotQA & $+0.0345$ & $+0.0377$ \\
2WikiMultiHopQA & $+0.0441$ & $+0.0496$ \\
MuSiQue & $+0.0518$ & $+0.0545$ \\
\bottomrule
\end{tabular*}
\vspace{4pt}

\begin{tabular*}{\textwidth}{l@{\extracolsep{\fill}}r}
\toprule
\multicolumn{2}{l}{\emph{(b) Three-collection one-permutation width decomposition}} \\
Three-collection component & Mean \\
\midrule
Observed width effect & $+0.0435$ \\
Informative alternatives & $+0.0482$ \\
Prompt length and target position & $+0.0034$ \\
Answer skeleton & $-0.0081$ \\
\bottomrule
\end{tabular*}
\end{table}

\textbf{Real companion passages, not the longer answer template, produce the QA width gain (\Cref{tab:qa-width}).}
To separate companion content from the larger prompt and readout, we score the same target passage in four conditions: alone with one grade slot; with its nine real candidate passages and ten slots; with nine unrelated, length-matched filler passages and ten slots; and alone under the same ten-slot answer template.
The total width effect compares real companions with the one-slot condition.
Within that total, real companions minus fillers measures informative companion content, fillers minus the target-alone ten-slot condition measures prompt length and target position, and the two target-alone conditions isolate the answer template.
Only the real-companion term is large and positive; the length/position term is small and the answer-template term is negative.
A separate check replaces one $B=1$ engine call with ten independently batched $B=1$ calls; the three-collection mean changes only from $+0.0470$ to $+0.0459$.
The gain therefore comes from information in the other candidate passages rather than from the larger template or request batching (\Cref{sec:copresence}).

\FloatBarrier
\subsection{Response ranking}
\label{app:respranking}

The five collections contain four to eight candidates, with four requiring more than one $B=4$ prompt.
The pointwise control is trained at $B=1$ on pointwise-teacher labels under the same Qwen3-4B evaluation protocol used for the batched student.

\textbf{Shared-context serving raises mean nDCG@1 on both PPE collections, changes little on Nectar, and lowers it on RM-Bench; the PPE-MATH and RM-Bench signs transfer to Gemma-E4B (\Cref{tab:rr-controls}).}
Across the two PPE collections and RM-Bench, the ten-order-label and pointwise-label $B=1$ variants differ by at most 0.006 nDCG@1; shared-context serving then improves both PPE collections but degrades RM-Bench.
The quality change therefore emerges when candidates share the serving prompt rather than from averaging shared-context information into the teacher labels.
Gemma-E4B reproduces the positive PPE-MATH and negative RM-Bench effects, while the PPE MMLU-Pro interval covers zero.

\begin{table}[htbp]
\centering
\small
\caption{\textbf{Response-ranking width and design effects vary by collection and model family.}
Block (a) reports mean fixed-weight width effects.
Block (b) compares nDCG@1 levels for three training/serving designs under one Qwen3-4B evaluation-file readout; it omits Nectar, whose pointwise level is in \Cref{tab:pairrm}.
Block (c) repeats the batched-minus-pointwise comparison on Gemma-E4B with paired prompt-bootstrap intervals.
Blocks (a) and (b) estimate different comparisons.
Comparing prompts against UltraFeedback found no exact overlap for RewardBench-2 or either PPE collection.
Basis: blocks (a--b) use Qwen3-4B and block (c) uses Gemma-E4B.}
\label{tab:rr-controls}
\setlength{\tabcolsep}{3pt}
\begin{tabular*}{\textwidth}{l@{\extracolsep{\fill}}rrr}
\toprule
\multicolumn{4}{l}{\emph{(a) Mean fixed-weight width effects}} \\
Collection & One-permutation effect & Order-averaged effect & Pool size \\
\midrule
RewardBench-2 & $+0.020$ & $+0.036$ & 4 \\
Nectar & $+0.006$ & $+0.026$ & 7 \\
PPE-MATH & $+0.096$ & $+0.132$ & 8 \\
PPE MMLU-Pro & $+0.074$ & $+0.076$ & 8 \\
RM-Bench & $-0.040$ & $-0.033$ & 6 \\
\bottomrule
\end{tabular*}
\vspace{4pt}

\begin{tabular*}{\textwidth}{l@{\extracolsep{\fill}}rrr}
\toprule
\multicolumn{4}{l}{\emph{(b) Label construction and serving geometry, Qwen3-4B}} \\
Collection & Ten-order labels ($B=1$) & Pointwise labels ($B=1$) & OC-SFT ($B=4$) \\
\midrule
RewardBench-2 & 0.657 & 0.605 & 0.690 \\
PPE-MATH & 0.723 & 0.719 & 0.805 \\
PPE MMLU-Pro & 0.594 & 0.592 & 0.666 \\
RM-Bench & 0.748 & 0.754 & 0.721 \\
\bottomrule
\end{tabular*}
\vspace{4pt}

\begin{tabular*}{\textwidth}{l@{\extracolsep{\fill}}rr}
\toprule
\multicolumn{3}{l}{\emph{(c) Family replication, Gemma-E4B}} \\
Collection & OC-SFT ($B=4$) minus pointwise ($B=1$) & 95\% CI \\
\midrule
PPE-MATH & $+0.1094$ & [$+0.0664$, $+0.1523$] \\
PPE MMLU-Pro & $+0.0098$ & [$-0.0391$, $+0.0566$] \\
RM-Bench & $-0.0648$ & [$-0.0897$, $-0.0407$] \\
\bottomrule
\end{tabular*}
\end{table}

Skywork-Reward-V2-Qwen3-4B \citep{liu2025skywork} is a 4B pointwise reward model that scores each response independently; PairRM \citep{jiang2023} is a 0.4B pairwise reward model that compares two responses at a time (\Cref{tab:pairrm}).
Skywork has higher nDCG@1 on RewardBench-2, Nectar and RM-Bench, whereas OC-SFT leads on PPE-MATH and PPE MMLU-Pro.
PairRM falls below the pointwise control on all five collections; the paired interval excludes zero on four. %
PairRM was trained on six large preference datasets, including UltraFeedback, and Skywork also uses much heavier preference supervision than our roughly 30K prompts. Architecture, scale and training data all differ at once, so the comparison cannot attribute the gap to any one of them.

\begin{table}[htbp]
\centering
\small
\caption{\textbf{PairRM and Skywork against matched Qwen3-4B controls.}
Each row gives matched nDCG@1 levels; PairRM contrasts use paired query bootstraps with 20{,}000 resamples.
Basis: seed 42; matched query IDs per collection; UltraFeedback is excluded because it is train-on-test for PairRM.}
\label{tab:pairrm}
\scriptsize
\setlength{\tabcolsep}{2pt}
\begin{tabular*}{\textwidth}{l@{\extracolsep{\fill}}rrrrrrrrr}
\toprule
\multicolumn{2}{c}{} & \multicolumn{4}{c}{nDCG@1} & \multicolumn{2}{c}{PairRM $-$ pointwise} & \multicolumn{2}{c}{OC-SFT $-$ PairRM} \\
\cmidrule(lr){3-6}\cmidrule(lr){7-8}\cmidrule(lr){9-10}
Collection & $n$ & Pointwise & PairRM & Skywork & OC-SFT & Estimate & 95\% CI & Estimate & 95\% CI \\
\midrule
Nectar & 434 & 0.5975 & 0.5914 & 0.6467 & 0.6152 & $-0.0061$ & [$-0.0430$, $+0.0315$] & $+0.0238$ & [$-0.0115$, $+0.0591$] \\
RewardBench-2 & 1{,}763 & 0.6052 & 0.4974 & 0.8174 & 0.6903 & $-0.1078$ & [$-0.1361$, $-0.0788$] & $+0.1929$ & [$+0.1651$, $+0.2201$] \\
PPE-MATH & 512 & 0.7188 & 0.4707 & 0.7656 & 0.8047 & $-0.2480$ & [$-0.3047$, $-0.1914$] & $+0.3340$ & [$+0.2793$, $+0.3867$] \\
PPE MMLU-Pro & 512 & 0.5918 & 0.4961 & 0.6445 & 0.6660 & $-0.0957$ & [$-0.1523$, $-0.0371$] & $+0.1699$ & [$+0.1133$, $+0.2246$] \\
RM-Bench & 1{,}327 & 0.7543 & 0.6353 & 0.8237 & 0.7212 & $-0.1191$ & [$-0.1492$, $-0.0889$] & $+0.0859$ & [$+0.0543$, $+0.1176$] \\
\bottomrule
\end{tabular*}
\end{table}

\FloatBarrier
\section{Consumer outputs and reported scores}
\label{app:consumers}

We apply a frozen threshold (\Cref{app:retained}), a frozen reader (\Cref{app:reader}) and a fixed chosen/rejected-pair selector (\Cref{app:prefpair}) to the same $M=10$ random permutations, relate the retained set to standard ranking metrics (\Cref{app:rankmetrics}), and recompute a reported benchmark score (\Cref{app:reportedscore}), to test whether presentation dependence propagates from rankings to downstream decisions and reported quality.
Holding each consumer fixed isolates variation caused by the scorer because the consumer itself does not change.

\subsection{A frozen threshold: the retained set}
\label{app:retained}

For each pair of scorer permutations, let $A$ and $B$ be the documents admitted by the same frozen threshold.
\emph{Retained-set overlap} averages the Jaccard index $|A\cap B|/|A\cup B|$ over all permutation pairs; it equals one when every permutation retains the same set.

\textbf{The retained-set gain transfers across model families (\Cref{tab:jaccard-perbase}).}
Every dense base improves on the 18-collection mean, with positive collection-level gains in 196 of 198 cells; the mixture-of-experts base improves on all 18 collections.
What the threshold retains is therefore made more reproducible by the objective rather than by the primary base.

\begin{table}[htbp]
\centering
\small
\caption{\textbf{Retained-set overlap gain of OC-SFT over single-order distillation by base model.}
Each gain is an equal-weight collection mean.
Single-order distillation is the baseline because the off-the-shelf fixed threshold degenerates to retain-all on TREC-COVID.
The 95\% CI column gives paired collection-bootstrap intervals; the last column counts positive collection-level gains.
Basis: 18 reranking collections at a frozen per-collection F1 threshold, seed 42, independently selected checkpoints on each base.}
\label{tab:jaccard-perbase}
\begin{tabular*}{\textwidth}{l@{\extracolsep{\fill}}rrr}
\toprule
Base & Gain & 95\% CI & Improving \\
\midrule
Qwen3-1.7B        & $+0.369$ & [$+0.298$, $+0.423$] & 17/18 \\
Qwen3-4B          & $+0.169$ & [$+0.152$, $+0.186$] & 18/18 \\
Qwen3-8B          & $+0.254$ & [$+0.233$, $+0.274$] & 18/18 \\
Qwen3-14B         & $+0.248$ & [$+0.210$, $+0.282$] & 18/18 \\
Qwen3-32B         & $+0.264$ & [$+0.237$, $+0.292$] & 18/18 \\
\midrule
Gemma-E2B         & $+0.186$ & [$+0.173$, $+0.201$] & 18/18 \\
Gemma-E4B         & $+0.116$ & [$+0.100$, $+0.131$] & 18/18 \\
Gemma-31B         & $+0.135$ & [$+0.088$, $+0.182$] & 17/18 \\
\midrule
Granite-3B        & $+0.174$ & [$+0.149$, $+0.202$] & 18/18 \\
Granite-8B        & $+0.137$ & [$+0.123$, $+0.152$] & 18/18 \\
Granite-30B       & $+0.216$ & [$+0.197$, $+0.235$] & 18/18 \\
\midrule
Gemma-26B-A4B MoE & $+0.106$ & [$+0.085$, $+0.135$] & 18/18 \\
\bottomrule
\end{tabular*}
\end{table}

\subsection{Ranking metrics and the retained set}
\label{app:rankmetrics}

\textbf{Retained-set overlap measures a real loss, and no ranking metric substitutes for it (\Cref{tab:jaccard-percollection}).}
For both single-order distillation and OC-SFT, per-query retained-set F1 varies about as much as nDCG@10 across presentations (\Cref{fig:spread}).
The loss is concentrated near the cutoff, where churned documents have relevance density 0.215, versus 0.438 among documents retained under both permutations.
The ordering $\tau$-PSI induces agrees with the ordering overlap induces more often than nDCG@10's does, but the agreement is incomplete.
Collection averaging reduces the apparent spread, so neither a ranking metric nor a collection mean shows how a threshold's output changes under reshuffling.

\begin{table}[htbp]
\centering
\small
\caption{\textbf{Retained-set overlap by collection for seven single-permutation systems within 0.022 nDCG@10.}
The quality and overlap winners differ on 11 of 18 collections.
Excluding GPT-5.4, whose $\tau$-PSI is not reported, median pairwise disagreement with overlap is 0.067 for $\tau$-PSI and 0.267 for nDCG@10.
The six-system discordance calculation uses seed-42 nDCG@10 and $\tau$-PSI, with jina at matched $B=20$.
\emph{F1 spread} is a query's mean retained-set F1 minus its worst permutation, averaged over queries, so the set a threshold admits is about as permutation-sensitive as the ranking it is read from.
Basis: three-seed means for trained rows and single evaluations for external rows, at the frozen per-collection F1 thresholds of \Cref{tab:levels}; per-collection quality and instability are in \Cref{tab:grids-ndcg,tab:grids-tau,tab:public-full}.}
\label{tab:jaccard-percollection}
\scriptsize
\begin{tabular*}{\textwidth}{l@{\extracolsep{\fill}}rrrrrrr}
\toprule
Collection & \emph{GPT-5.4} & OC-SFT & Order-avg. & Perm.\ aug. & DebiasFirst & Single-order & \emph{jina-v3} \\
\midrule
DL19 & 0.779 & 0.882 & 0.827 & 0.832 & 0.835 & 0.732 & 0.769 \\
DL20 & 0.782 & 0.881 & 0.812 & 0.827 & 0.826 & 0.743 & 0.721 \\
DL21 & 0.846 & 0.899 & 0.814 & 0.832 & 0.832 & 0.692 & 0.756 \\
DL22 & 0.685 & 0.825 & 0.785 & 0.774 & 0.774 & 0.638 & 0.685 \\
DL23 & 0.617 & 0.814 & 0.701 & 0.747 & 0.741 & 0.634 & 0.621 \\
Touch\'e-2020 & 0.832 & 0.832 & 0.757 & 0.795 & 0.773 & 0.622 & 0.670 \\
FiQA & 0.640 & 0.850 & 0.723 & 0.750 & 0.748 & 0.661 & 0.662 \\
NFCorpus & 0.738 & 0.862 & 0.770 & 0.796 & 0.791 & 0.679 & 0.666 \\
ArguAna & 0.607 & 0.615 & 0.539 & 0.544 & 0.540 & 0.484 & 0.618 \\
Climate-FEVER & 0.624 & 0.798 & 0.707 & 0.704 & 0.711 & 0.621 & 0.589 \\
TREC-COVID & 0.831 & 0.933 & 0.887 & 0.901 & 0.902 & 0.778 & 0.791 \\
DBPedia & 0.696 & 0.836 & 0.730 & 0.756 & 0.753 & 0.653 & 0.631 \\
SciFact & 0.806 & 0.886 & 0.785 & 0.787 & 0.796 & 0.744 & 0.688 \\
Signal-1M & 0.597 & 0.813 & 0.731 & 0.721 & 0.726 & 0.634 & 0.625 \\
TREC-NEWS & 0.723 & 0.841 & 0.728 & 0.741 & 0.760 & 0.644 & 0.655 \\
Robust04 & 0.739 & 0.859 & 0.770 & 0.784 & 0.782 & 0.694 & 0.665 \\
Legal-A & 0.579 & 0.779 & 0.604 & 0.671 & 0.651 & 0.544 & 0.581 \\
Legal-B & 0.606 & 0.828 & 0.701 & 0.720 & 0.715 & 0.610 & 0.605 \\
\midrule
Mean overlap $\uparrow$ & 0.707 & \textbf{0.835} & 0.743 & 0.760 & 0.759 & 0.656 & 0.667 \\
\midrule
\multicolumn{8}{l}{\emph{Summary measures, 18-collection means}} \\
nDCG@10 $\uparrow$ & 0.468 & 0.459 & 0.455 & 0.455 & 0.454 & 0.449 & 0.447 \\
Retained-set F1 $\uparrow$ & 0.415 & 0.407 & 0.410 & 0.406 & 0.409 & 0.405 & 0.420 \\
Retained-set F1 spread $\downarrow$ & 0.081 & \textbf{0.040} & 0.057 & 0.056 & 0.055 & 0.071 & 0.087 \\
$\tau$-PSI $\downarrow$ & --- & \textbf{0.083} & 0.130 & 0.129 & 0.128 & 0.209 & 0.177 \\
\bottomrule
\end{tabular*}
\end{table}

\subsection{A frozen reader: the verdict and the answer}
\label{app:reader}

\textbf{At 4B, OC-SFT lowers verdict and answer flip while reader accuracy remains level (\Cref{tab:downstream}).}
The default reader is a frozen Granite-4.1-8B decoding greedily into three constrained claim-verification labels.
A frozen Qwen3-4B reader repeats the comparison on both claim-verification and all three question-answering collections.
\emph{Verdict flip} and \emph{answer flip} are mean pairwise disagreement rates over the $M=10$ random permutations.
The reductions are positive under both readers on claim verification and QA, while all reported reader-accuracy intervals cover zero.
The same-evidence controls are positive for claim verification and QA, and their magnitudes fall within the range of the all-query effects.
The lower flip rates therefore do not require a change in which evidence reaches the reader.
EMA closes more of the rank-level gap than augmentation but loses 0.023 QA exact match; together, these controls show that OC-SFT's rank-level gain reaches the reader output without a measurable accuracy trade-off (\Cref{tab:variants,tab:downstream}).

\begin{table}[htbp]
\centering
\small
\caption{\textbf{Verdict- and answer-flip reductions and controls at 4B.}
Positive estimates favor the variant after the arrow: lower flip for flip rows and higher accuracy for accuracy rows.
Block (a) reports SD where repeated fits are available; block (b) reports query-paired 95\% intervals at one fixed fit.
In block (c), percentages are the share of queries on which both variants select the same top five.
Block (d) uses a query-paired interval for Climate-FEVER and hierarchical query-within-collection intervals for the QA aggregates.
Basis: Granite-4.1-8B reader unless Qwen3-4B is named; QA rows use 200 questions per collection.}
\label{tab:downstream}
\scriptsize
\setlength{\tabcolsep}{3pt}
\begin{tabular*}{\textwidth}{l@{\extracolsep{\fill}}lrr}
\toprule
\multicolumn{4}{l}{\emph{(a) Repeated-fit flip and accuracy estimates}} \\
Contrast & Collection / output & Estimate & SD \\
\midrule
Single-order $\to$ OC-SFT & Climate-FEVER & $+0.0263$ & 0.0031 \\
Single-order $\to$ OC-SFT & SciFact & $+0.0251$ & 0.0014 \\
Single-order $\to$ OC-SFT, second reader & Climate-FEVER & $+0.0304$ & 0.0025 \\
Perm.\ aug. $\to$ EMA & QA exact match & $-0.0228$ & 0.0029 \\
\bottomrule
\end{tabular*}
\vspace{6pt}

\begin{tabular*}{\textwidth}{l@{\extracolsep{\fill}}lrrr}
\toprule
\multicolumn{5}{l}{\emph{(b) Flip reductions with query-paired intervals}} \\
Contrast & Collection / output & $n$ & Estimate & 95\% CI \\
\midrule
Single-order $\to$ OC-SFT, second reader & SciFact & 300 & $+0.0368$ & [$+0.0207$, $+0.0532$] \\
Single-order $\to$ OC-SFT & HotpotQA answer & 200 & $+0.0487$ & [$+0.0283$, $+0.0698$] \\
Single-order $\to$ OC-SFT & 2Wiki answer & 200 & $+0.0459$ & [$+0.0221$, $+0.0707$] \\
Single-order $\to$ OC-SFT & MuSiQue answer & 200 & $+0.0700$ & [$+0.0427$, $+0.0979$] \\
Single-order $\to$ OC-SFT, second reader & HotpotQA answer & 200 & $+0.0274$ & [$+0.0054$, $+0.0494$] \\
Single-order $\to$ OC-SFT, second reader & 2Wiki answer & 200 & $+0.0891$ & [$+0.0643$, $+0.1143$] \\
Single-order $\to$ OC-SFT, second reader & MuSiQue answer & 200 & $+0.0823$ & [$+0.0538$, $+0.1122$] \\
\bottomrule
\end{tabular*}
\vspace{6pt}

\begin{tabular*}{\textwidth}{l@{\extracolsep{\fill}}llr}
\toprule
\multicolumn{4}{l}{\emph{(c) Same-evidence point controls}} \\
Contrast & Collection / output & Matched-query share & Estimate \\
\midrule
Single-order $\to$ OC-SFT, same top 5 & Climate-FEVER verdict & 17.5\% & $+0.030$ \\
Single-order $\to$ OC-SFT, same top 5 & QA answer & 45.2\% & $+0.061$ \\
\bottomrule
\end{tabular*}
\vspace{6pt}

\begin{tabular*}{\textwidth}{l@{\extracolsep{\fill}}lrrr}
\toprule
\multicolumn{5}{l}{\emph{(d) Reader accuracy}} \\
Contrast & Collection / output & $n$ & Estimate & 95\% CI \\
\midrule
Single-order $\to$ OC-SFT & Climate-FEVER verdict & 1{,}381 & $-0.0036$ & [$-0.0111$, $+0.0038$] \\
Single-order $\to$ OC-SFT & QA exact match & 200$\times$3 & $-0.0058$ & [$-0.0153$, $+0.0035$] \\
Single-order $\to$ OC-SFT & QA F1 & 200$\times$3 & $-0.0016$ & [$-0.0104$, $+0.0071$] \\
\bottomrule
\end{tabular*}
\end{table}

\FloatBarrier
\subsection{A preference model: the selected pair}
\label{app:prefpair}

A preference pipeline consumes both ends of a ranking, so \emph{pair flip} is the mean pairwise rate at which permutations select different chosen and rejected responses.

\textbf{OC-SFT is the most reproducible trained variant on every response-ranking collection (\Cref{tab:rr-pairflip}).}
On each collection, it changes the chosen/rejected pair less often than every other trained row, so the rank-level stability gain reaches the preference decision rather than only the complete ranking.
Gold crossing and serving-time $K$ sweeps for this consumer are reported in \Cref{app:consumerk}.

\begin{table}[htbp]
\centering
\small
\caption{\textbf{Preference-pair flip on response ranking.}
Lower is better.
Basis: Qwen3-4B seed 42; Nectar uses the deduplicated 434-prompt subset.}
\label{tab:rr-pairflip}
\begin{tabular*}{\textwidth}{l@{\extracolsep{\fill}}rrrrrr}
\toprule
Collection & Off the shelf & Single-order & Order-avg. & Perm.\ aug. & DebiasFirst & OC-SFT \\
\midrule
RewardBench-2 & 0.763 & 0.757 & 0.496 & 0.447 & 0.441 & \textbf{0.356} \\
Nectar & 0.905 & 0.872 & 0.718 & 0.704 & 0.700 & \textbf{0.650} \\
PPE-MATH & 0.920 & 0.914 & 0.796 & 0.794 & 0.792 & \textbf{0.744} \\
PPE MMLU-Pro & 0.930 & 0.940 & 0.845 & 0.835 & 0.837 & \textbf{0.818} \\
RM-Bench & 0.869 & 0.861 & 0.769 & 0.776 & 0.776 & \textbf{0.739} \\
\bottomrule
\end{tabular*}
\end{table}

\subsection{A reported benchmark score}
\label{app:reportedscore}

A reported benchmark score may use one permutation, whereas $\tau$-PSI and retained-set overlap aggregate all ten.
We therefore test whether a single permutation supports the same cross-system quality conclusion as the order-marginal score.
Skywork \citep{liu2025skywork} is pointwise and has one fixed nDCG@1 per collection, whereas OC-SFT has one score per permutation; the system with the higher collection-level nDCG@1 is the leader.

\textbf{Permutation range exceeds the mean OC-SFT--Skywork gap on two collections without reversing the observed leader (\Cref{tab:reportedscore}).}
The range is wider than the absolute gap on Nectar and PPE MMLU-Pro, but no one of the $M=10$ random permutations reverses the leader.
Permutation choice therefore changes the apparent margin rather than the observed ordering on these runs.
Only PPE MMLU-Pro's paired query-bootstrap interval covers zero, so query sampling leaves that comparison unresolved as well.
We report the order-marginal OC-SFT--Skywork gap rather than selecting one arbitrary permutation; the paired bootstrap quantifies the remaining query-sampling uncertainty.

\begin{table}[htbp]
\centering
\small
\caption{\textbf{Permutation range exceeds the mean system gap on two collections.}
Each OC-SFT permutation is aggregated over the prompts scored by pointwise Skywork, so both sides share a basis.
Range is max minus min of OC-SFT's nDCG@1 across $M{=}10$ permutations.
Bold marks ranges larger than the absolute system gap.
Perm.\ SD is the standard deviation across permutations.
Gap is OC-SFT minus Skywork; its intervals are paired prompt bootstraps and carry query sampling rather than permutation range.
Basis: Qwen3-4B seed 42; Nectar uses the deduplicated 434-prompt subset.
}
\label{tab:reportedscore}
\begin{tabular}{lrrrrr}
\toprule
Collection & $n$ & Perm.\ SD & Range & Gap & 95\% CI \\
\midrule
RewardBench-2 & 1{,}763 & 0.0065 & 0.0199 & $-0.136$ & [$-0.154$, $-0.118$] \\
Nectar & 434 & 0.0162 & \textbf{0.0476} & $-0.031$ & [$-0.060$, $-0.002$] \\
PPE-MATH & 512 & 0.0102 & 0.0332 & $+0.047$ & [$+0.014$, $+0.080$] \\
PPE MMLU-Pro & 512 & 0.0129 & \textbf{0.0410} & $+0.019$ & [$-0.017$, $+0.055$] \\
RM-Bench & 1{,}327 & 0.0109 & 0.0317 & $-0.110$ & [$-0.128$, $-0.092$] \\
\bottomrule
\end{tabular}
\end{table}

\FloatBarrier
\section{Serving width and system trade-offs}
\label{app:width}

We measure how serving width affects quality and stability (\Cref{app:widthrerank}), decompose the width effect into its sources (\Cref{app:widthdecomp}), test tolerance to a mismatch between training and serving width (\Cref{app:trainserve}), and report the range that is safe to serve (\Cref{app:widthrange}) and the latency of batched serving (\Cref{app:deployment}).

\subsection{Passage-reranking width and estimator controls}
\label{app:widthrerank}

\textbf{A single permutation does not measurably improve with serving width on passage reranking, while order averaging adds 0.020 nDCG@10 (\Cref{tab:width-controls}).}
The order-averaged comparison averages ten $B=20$ permutations and compares them with one $B=1$ pass, whereas the one-permutation column compares one pass at $B=20$ with one at $B=1$.
Matching the $B=1$ estimator with ten independently rebatched passes leaves the order-averaged result nearly unchanged (\Cref{tab:width-controls}).
Each $B=1$ prompt still contains one candidate, so this intervention changes engine-call batching, meaning how many prompts the inference engine groups into one call, rather than what shares a prompt.
The $+0.020$ effect is therefore not an artifact of averaging only the $B=20$ scores.

\begin{table}[htbp]
\centering
\small
\caption{\textbf{The $+0.020$ passage-reranking gain survives estimator matching, while engine-call batching has little effect on retained-set overlap.}
Block (a) reports the one-permutation and order-averaged width effects, including a matched $B=1$ estimator built from ten independently rebatched passes.
Block (b) holds candidate order fixed and varies engine-call batching, then contrasts it with the candidate-order perturbation.
Basis: block (a) uses the 18 reranking collections at seed 42; block (b) uses the 16 public ones at seed 42 and the frozen-threshold protocol of \Cref{app:consumers}.}
\label{tab:width-controls}
\begin{tabular*}{\textwidth}{l@{\extracolsep{\fill}}lrr}
\toprule
\multicolumn{4}{l}{\emph{(a) nDCG@10 width effect}} \\
Comparison & Collection basis & Effect & 95\% CI \\
\midrule
One permutation & 18 reranking & $+0.0001$ & [$-0.0133$, $+0.0177$] \\
Order-averaged, original & 18 reranking & $+0.0201$ & [$+0.0063$, $+0.0429$] \\
Order-averaged, matched $B=1$ & 18 reranking & $+0.0191$ & [$+0.0050$, $+0.0424$] \\
Order-averaged, original & Excluding ArguAna & $+0.0098$ & [$+0.0051$, $+0.0147$] \\
Order-averaged, matched $B=1$ & Excluding ArguAna & $+0.0088$ & [$+0.0035$, $+0.0144$] \\
\bottomrule
\end{tabular*}
\vspace{6pt}

\begin{tabular*}{\textwidth}{l@{\extracolsep{\fill}}llr}
\toprule
\multicolumn{3}{l}{\emph{(b) Retained-set overlap under matched perturbations}} \\
Serving width & Perturbation & Overlap \\
\midrule
$B=1$ & Engine-call batching & 0.9797 \\
$B=20$ & Engine-call batching, fixed candidate order & 0.9848 \\
$B=20$ & Candidate order & 0.8353 \\
\bottomrule
\end{tabular*}
\end{table}

\textbf{Engine-call batching has a much smaller effect than candidate order (\Cref{tab:width-controls}).}
The two batching-only readings are close despite different serving widths, while changing candidate order at $B=20$ produces lower retained-set overlap; these losses are directional controls rather than additive components because changing order also changes engine-call composition.

\subsection{Decomposition of the width effect}
\label{app:widthdecomp}

\textbf{Outside ArguAna, the order-averaged width effect is almost entirely companion content (\Cref{tab:decomp}).}
ArguAna instead derives most of its effect from length, slot and skeleton terms. %

\begin{table}[htbp]
\centering
\small
\caption{\textbf{Decomposition of the width effect.}
At fixed Qwen3-4B OC-SFT weights, $q_{\mathrm{alone}}$ scores a target alone under one answer slot, $q_{\mathrm{comp}}$ adds its 19 real companions, $q_{\mathrm{fill}}$ substitutes 19 unrelated length-matched fillers, and $q_{\mathrm{skel}}$ returns the target alone under the full 20-slot template.
Width is $q_{\mathrm{comp}}-q_{\mathrm{alone}}$, content is $q_{\mathrm{comp}}-q_{\mathrm{fill}}$, length plus slot is $q_{\mathrm{fill}}-q_{\mathrm{skel}}$, and skeleton is $q_{\mathrm{skel}}-q_{\mathrm{alone}}$.
The upper block uses the first-stage permutation; the lower block averages the $B=20$ scores over $K=10$ serving permutations.
Positive entries mean the wider serving width ranks better.
Content, length plus slot, and skeleton sum to width up to rounding.
Starred cells have query-paired 95\% bootstrap intervals clear of zero.
Basis: Qwen3-4B OC-SFT, seed 42, 18 reranking collections.}
\label{tab:decomp}
\begin{tabular}{lrrrr}
\toprule
Collection basis & Width & Content & Length + slot & Skeleton \\
\midrule
\multicolumn{5}{l}{\emph{First-stage order}} \\
18 reranking & $-0.0002$ & $-0.0072$ & $+0.0039$ & $+0.0031$ \\
Excluding ArguAna & $-0.0069$ & $-0.0047$ & $-0.0016$ & $-0.0007$ \\
ArguAna & $+0.1144^\ast$ & $-0.0496^\ast$ & $+0.0965^\ast$ & $+0.0675^\ast$ \\
\midrule
\multicolumn{5}{l}{\emph{Averaged over $K=10$ serving permutations}} \\
18 reranking & $+0.0194$ & $+0.0085$ & $+0.0026$ & $+0.0083$ \\
Excluding ArguAna & $+0.0089$ & $+0.0086$ & $+0.0007$ & $-0.0004$ \\
ArguAna & $+0.1979^\ast$ & $+0.0071^\ast$ & $+0.0341^\ast$ & $+0.1567^\ast$ \\
\bottomrule
\end{tabular}
\end{table}

\subsection{Training width and train-serve mismatch}
\label{app:trainserve}

\textbf{Training at $B=20$ makes the weights width-tolerant, whereas a $B=1$-trained student loses quality when served at $B=20$ (\Cref{tab:traindeploy}).}
Only the off-the-shelf scorer and the $B=1$-trained student have width-effect intervals excluding zero.
In the train-serve comparison, the $B=20$-trained student transfers to $B=1$, while the $B=1$-trained student loses 0.050 nDCG@10 on a wide pass and recovers it only by averaging ten evaluations.
The same train-serve question for the teacher's labelling width is reported in \Cref{tab:teacher-design}.

\begin{table}[htbp]
\centering
\small
\caption{\textbf{Training width determines tolerance to serving-width changes.}
Block (a) measures $\tau$-PSI at $B{=20}$ and the single-pass nDCG@10 change from serving the same seed-42 weights at $B{=1}$ and $B{=20}$; intervals are paired 95\% bootstraps.
Block (b) crosses training and serving width using three-seed means; its final column averages ten $B=20$ permutations.
$B{=1}$ $\tau$-PSI entries are zero by construction.
Basis: Qwen3-4B, 18 reranking collections, equal-weight collection mean.}
\label{tab:traindeploy}
\begin{tabular*}{\textwidth}{l@{\extracolsep{\fill}}rrr}
\toprule
\multicolumn{4}{l}{\emph{(a) Fixed-weight width sensitivity, seed 42}} \\
Variant & $\tau$-PSI & Width effect & 95\% CI \\
\midrule
\multicolumn{4}{l}{\emph{Not trained at the serving width}} \\
Off the shelf & 0.298 & $-0.062$ & [$-0.082$, $-0.043$] \\
Student trained at $B=1$ & 0.222 & $-0.048$ & [$-0.062$, $-0.036$] \\
\midrule
\multicolumn{4}{l}{\emph{Trained at $B=20$}} \\
Order-avg. & 0.139 & $-0.004$ & [$-0.012$, $+0.004$] \\
DebiasFirst & 0.129 & $-0.005$ & [$-0.015$, $+0.005$] \\
Perm.\ aug. & 0.122 & $-0.006$ & [$-0.016$, $+0.003$] \\
\textbf{OC-SFT} & \textbf{0.085} & $+0.000$ & [$-0.013$, $+0.018$] \\
\bottomrule
\end{tabular*}
\vspace{6pt}

\begin{tabular*}{\textwidth}{l@{\extracolsep{\fill}}rrrrr}
\toprule
\multicolumn{6}{l}{\emph{(b) Train-serve width comparison, three-seed means}} \\
& \multicolumn{2}{c}{Serve $B=1$} & \multicolumn{2}{c}{Serve $B=20$, one pass} & Serve $B=20$, $K=10$ \\
\cmidrule(lr){2-3} \cmidrule(lr){4-5} \cmidrule(lr){6-6}
Train & nDCG@10 & $\tau$-PSI & nDCG@10 & $\tau$-PSI & nDCG@10 \\
\midrule
$B=20$ (OC-SFT) & 0.459 & 0 & 0.459 & 0.083 & 0.479 \\
$B=1$-trained student & 0.466 & 0 & 0.415 & 0.227 & 0.470 \\
\bottomrule
\end{tabular*}
\end{table}

\textbf{Training attenuates the score-level width effect without removing it, and the consistency penalty adds no distinguishable further reduction (\Cref{tab:copresence-level}).}
Single-order distillation and permutation augmentation reduce $|\mu_B-\mu_1|$, but augmentation and OC-SFT differ by only 0.013 score standard deviations with an interval covering zero.
Although training attenuates its magnitude, OC-SFT remains well above zero and therefore has not become pointwise.

\begin{table}[htbp]
\centering
\small
\caption{\textbf{Magnitude of the score-level width effect across training variants.}
Entries are collection-first mean $|\mu_B-\mu_1|$ normalized by pooled across-document score SD; they include companion content and prompt geometry.
The final row isolates the consistency penalty.
Order-averaged distillation is omitted because it changes the teacher targets.
Basis: 18 reranking collections, Qwen3-4B, seed 42; $B=20$ averages $K=10$ serving permutations, $B=1$ scores each document alone, and intervals bootstrap collections.}
\label{tab:copresence-level}
\begin{tabular}{lrr}
\toprule
Variant / contrast & Level or difference & 95\% CI \\
\midrule
Off the shelf & 0.521 & [0.449, 0.593] \\
Single-order & 0.300 & [0.246, 0.383] \\
Perm.\ aug. & 0.245 & [0.198, 0.323] \\
OC-SFT & 0.232 & [0.175, 0.330] \\
\midrule
Augmentation minus OC-SFT & $+0.013$ & [$-0.012$, $+0.034$] \\
\bottomrule
\end{tabular}
\end{table}

\subsection{Serving-width operating range}
\label{app:widthrange}

\textbf{Quality remains stable from $B=1$ through $B=30$ (\Cref{tab:deploywidth}).}
Across that range, mean nDCG@10 over the 18 reranking collections changes little, and $\tau$-PSI remains near its training-width value.
At $B=34$ both measures degrade, and at $B=40$ and $B=50$ they deteriorate further.

\begin{table}[htbp]
\centering
\small
\caption{\textbf{OC-SFT quality is stable through $B=1$ to $B=30$.}
The selected $B=20$-trained OC-SFT checkpoint is served at nine widths; higher nDCG@10 and lower $\tau$-PSI are better.
The training width is bold in the header.
Basis: Qwen3-4B, seed 42, $M=10$ random permutations for $B>1$; at $B=1$, $\tau$-PSI is zero by construction.}
\label{tab:deploywidth}
\setlength{\tabcolsep}{3pt}
\begin{tabular*}{\textwidth}{l@{\extracolsep{\fill}}rrrrrrrrr}
\toprule
Serve $B$ & 1 & 5 & 10 & \textbf{20} & 25 & 30 & 34 & 40 & 50 \\
\midrule
18-collection nDCG@10 $\uparrow$ & 0.459 & 0.453 & 0.454 & \textbf{0.459} & 0.461 & 0.460 & 0.446 & 0.345 & 0.263 \\
18-collection $\tau$-PSI $\downarrow$ & 0 & 0.094 & 0.091 & \textbf{0.085} & 0.083 & 0.094 & 0.135 & 0.233 & 0.275 \\
\bottomrule
\end{tabular*}
\end{table}

\subsection{Deployment and system trade-offs}
\label{app:deployment}

\begin{figure}[htbp]
\centering
\includegraphics[width=\linewidth]{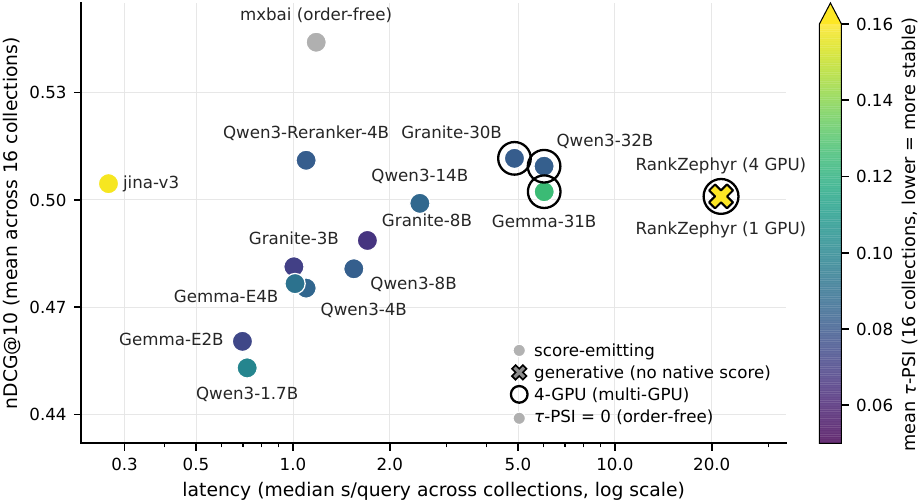}
\caption{\textbf{The system frontier across 16 public reranking collections.}
The horizontal axis is median per-query latency, the vertical axis mean nDCG@10, color mean $\tau$-PSI, and marker shape whether the system emits a per-document score.
The model-family points are trained OC-SFT checkpoints on the named bases.
jina, mxbai and RankZephyr are external rerankers shown at their stated serving configurations.
Gray denotes $\tau$-PSI zero by construction for the order-free point.
nDCG@10 and $\tau$-PSI are equal-weight means and latency is the median of per-collection medians over 16 public reranking collections.
Ringed points use four L40S GPUs rather than one.}
\label{fig:frontier}
\end{figure}

\textbf{Shared-context students match RankZephyr quality at lower latency while retaining per-document scores (\Cref{fig:frontier}).}
All model-family points in the blue cluster are OC-SFT students, not off-the-shelf checkpoints.
The Qwen3-Reranker-4B point starts from a purpose-built reranking checkpoint; our OC-SFT adaptation raises its public-16 nDCG@10 by 0.05 and lowers $\tau$-PSI by 72\% (\Cref{tab:specialized}).
The 30B-class students similarly match RankZephyr's mean quality at lower latency while emitting per-document scores.
jina and mxbai are external, purpose-built rerankers trained on substantially broader corpora and with specialized procedures, including cross-system hard-negative mining for jina and reinforcement, contrastive and preference learning for mxbai \citep{wang2025jina,li2026}.
They are therefore reference systems rather than supervision-matched baselines: mxbai leads on native-geometry quality and jina on latency, while RankZephyr remains more order-sensitive.

\textbf{Candidate batching roughly halves per-query latency at concurrency 1 (\Cref{tab:batching-latency}).}
Across the 16 public collections, the median reduction relative to $B=1$ is 51.3\% at $B=10$ and 45.4\% at $B=20$.
We retain $B=20$ to match the training geometry, while $B=10$ offers lower median latency except on Signal-1M and DL21--23.
The batching reduction compounds with serving one permutation rather than ten (\Cref{fig:amortconsumers}).

\begin{table}[htbp]
\centering
\small
\caption{\textbf{Candidate batching roughly halves per-query latency.}
Collection rows report seconds per query and the reduction relative to $B=1$.
Summary rows are equal-collection medians; brackets give 95\% bootstrap intervals for the reduction.
Basis: Qwen3-4B OC-SFT, seed 42, 18 collections, one L40S, warm vLLM and client concurrency 1.
Each collection is timed three times on the same GPU, with the width order varied across repetitions.
These within-model timings use query batch size 1 and are separate from the system-level latency comparison in \Cref{fig:frontier}.}
\label{tab:batching-latency}
\begin{tabular*}{\textwidth}{l@{\extracolsep{\fill}}rrrrr}
\toprule
Collection & $B=1$ & $B=10$ & $B=20$ & $B=10$ $\downarrow$ & $B=20$ $\downarrow$ \\
\midrule
DL19 & 2.223 & 1.020 & 1.086 & 54.1\% & 51.2\% \\
DL20 & 2.237 & 1.032 & 1.124 & 53.9\% & 49.7\% \\
TREC-COVID & 2.214 & 1.111 & 1.255 & 49.8\% & 43.3\% \\
NFCorpus & 2.249 & 1.092 & 1.226 & 51.5\% & 45.5\% \\
Touch\'e-2020 & 2.404 & 1.176 & 1.315 & 51.1\% & 45.3\% \\
DBPedia-Entity & 2.235 & 1.036 & 1.100 & 53.6\% & 50.8\% \\
SciFact & 2.272 & 1.145 & 1.304 & 49.6\% & 42.6\% \\
Signal-1M & 2.098 & 0.590 & 0.557 & 71.9\% & 73.4\% \\
TREC-News & 2.241 & 1.181 & 1.354 & 47.3\% & 39.6\% \\
Robust04 & 2.285 & 1.230 & 1.418 & 46.2\% & 37.9\% \\
FiQA & 2.296 & 1.193 & 1.373 & 48.1\% & 40.2\% \\
ArguAna & 2.345 & 1.228 & 1.372 & 47.6\% & 41.5\% \\
Climate-FEVER & 2.273 & 1.144 & 1.317 & 49.7\% & 42.0\% \\
DL21 & 2.188 & 0.970 & 0.913 & 55.7\% & 58.3\% \\
DL22 & 2.206 & 0.974 & 0.914 & 55.8\% & 58.6\% \\
DL23 & 2.172 & 0.944 & 0.914 & 56.5\% & 57.9\% \\
Legal-A & 2.278 & 1.134 & 1.274 & 50.2\% & 44.1\% \\
Legal-B & 2.264 & 1.111 & 1.243 & 50.9\% & 45.1\% \\
\midrule
Public-16 median & 2.239 & 1.101 & 1.240 & 51.3\% [48.9, 54.1] & 45.4\% [42.0, 51.7] \\
Primary-18 median & 2.245 & 1.111 & 1.249 & 51.0\% [49.7, 54.0] & 45.2\% [42.3, 51.0] \\
\bottomrule
\end{tabular*}
\end{table}

\FloatBarrier

\FloatBarrier
\section{Per-collection and cross-base reference results}
\label{app:fullresults}

Finally, we report per-collection results for the Qwen3-4B baselines (\Cref{app:percollection}), under alternative first stages (\Cref{app:altfirststage}), against specialized and external rerankers (\Cref{app:specialized}), and across twelve base models (\Cref{app:crossbase}), so that each aggregate finding can be checked collection by collection and its exceptions identified.

\subsection{Qwen3-4B per-collection reference}
\label{app:percollection}

\begin{table}[htbp]
\centering
\setlength{\tabcolsep}{0pt}
\caption{\textbf{Full per-collection reranking results for the Qwen3-4B student and its baselines.}
Higher nDCG@10 and lower $\tau$-PSI are better.
BM25 has zero $\tau$-PSI by construction; GPT-5.4 has no reported $\tau$-PSI because tied grades leave its relative ordering to the tie-break, which is document identifier for nDCG@10, as noted in \Cref{tab:levels}.
Basis: seed 42 and $M=10$ random permutations; \emph{Public 16} excludes Legal-A and Legal-B, whose candidate lists are precomputed. Body \Cref{tab:levels} reports the mean over all 18 collections.}
\label{tab:public-full}
\scriptsize
\begin{tabular*}{\textwidth}{l@{\extracolsep{\fill}}ccccccccccccccccccc}
\toprule
Method & \rothead{Public 16} & \rothead{DL19} & \rothead{DL20} & \rothead{DL21} & \rothead{DL22} & \rothead{DL23} & \rothead{Touch\'e} & \rothead{FiQA} & \rothead{NFCorpus} & \rothead{ArguAna} & \rothead{Climate} & \rothead{T-COVID} & \rothead{DBPedia} & \rothead{SciFact} & \rothead{Signal1M} & \rothead{T-NEWS} & \rothead{Robust04} & \rothead{Legal-A} & \rothead{Legal-B} \\[6pt]
\midrule
\multicolumn{20}{l}{\emph{nDCG@10, higher is better}} \\
\multicolumn{20}{l}{\emph{First-stage retrieval without reranking}} \\
BM25 retrieval & 0.39 & 0.51 & 0.49 & 0.45 & 0.27 & 0.26 & 0.44 & 0.24 & 0.34 & 0.40 & 0.17 & 0.59 & 0.32 & 0.68 & 0.33 & 0.40 & 0.41 & 0.30 & 0.32 \\
\multicolumn{20}{l}{\emph{Prompted closed model ($B=20$)}} \\
GPT-5.4 & 0.49 & 0.68 & 0.65 & 0.65 & 0.45 & 0.42 & 0.23 & 0.37 & 0.38 & 0.40 & 0.19 & 0.85 & 0.39 & 0.75 & 0.27 & 0.50 & 0.57 & 0.36 & 0.31 \\
\multicolumn{20}{l}{\emph{Qwen3-4B}} \\
Off the shelf & 0.38 & 0.60 & 0.55 & 0.54 & 0.39 & 0.33 & 0.22 & 0.25 & 0.31 & 0.21 & 0.16 & 0.73 & 0.33 & 0.51 & 0.22 & 0.33 & 0.45 & 0.29 & 0.24 \\
CapCal & 0.38 & 0.62 & 0.55 & 0.55 & 0.39 & 0.34 & 0.22 & 0.25 & 0.31 & 0.20 & 0.16 & 0.72 & 0.34 & 0.50 & 0.22 & 0.34 & 0.45 & 0.29 & 0.24 \\
Single-order & 0.46 & 0.72 & 0.67 & 0.67 & 0.45 & 0.43 & 0.26 & 0.35 & 0.35 & 0.26 & 0.19 & 0.75 & 0.40 & 0.70 & 0.29 & 0.42 & 0.52 & 0.34 & 0.29 \\
Order-avg. & 0.47 & 0.73 & 0.68 & 0.67 & 0.45 & 0.44 & 0.29 & 0.35 & 0.36 & 0.28 & 0.21 & 0.75 & 0.41 & 0.71 & 0.29 & 0.44 & 0.52 & 0.36 & 0.31 \\
Perm.\ aug. & 0.47 & 0.73 & 0.67 & 0.67 & 0.46 & 0.45 & 0.27 & 0.35 & 0.36 & 0.26 & 0.20 & 0.75 & 0.41 & 0.71 & 0.28 & 0.42 & 0.52 & 0.35 & 0.31 \\
DebiasFirst & 0.47 & 0.72 & 0.68 & 0.67 & 0.46 & 0.44 & 0.27 & 0.35 & 0.36 & 0.24 & 0.20 & 0.75 & 0.41 & 0.71 & 0.29 & 0.43 & 0.52 & 0.35 & 0.31 \\
OC-SFT & 0.48 & 0.73 & 0.68 & 0.68 & 0.47 & 0.45 & 0.28 & 0.37 & 0.36 & 0.31 & 0.19 & 0.74 & 0.42 & 0.71 & 0.30 & 0.42 & 0.52 & 0.35 & 0.31 \\
\midrule
\multicolumn{20}{l}{\emph{$\tau$-PSI, lower is better}} \\
\multicolumn{20}{l}{\emph{Qwen3-4B}} \\
Off the shelf & 0.29 & 0.30 & 0.30 & 0.31 & 0.32 & 0.33 & 0.29 & 0.26 & 0.28 & 0.27 & 0.27 & 0.28 & 0.32 & 0.30 & 0.33 & 0.27 & 0.26 & 0.34 & 0.34 \\
CapCal & 0.29 & 0.30 & 0.30 & 0.31 & 0.32 & 0.33 & 0.29 & 0.26 & 0.28 & 0.27 & 0.27 & 0.28 & 0.32 & 0.30 & 0.33 & 0.27 & 0.27 & 0.34 & 0.34 \\
Single-order & 0.20 & 0.19 & 0.18 & 0.18 & 0.21 & 0.21 & 0.18 & 0.18 & 0.21 & 0.22 & 0.19 & 0.18 & 0.22 & 0.23 & 0.23 & 0.17 & 0.19 & 0.23 & 0.23 \\
Order-avg. & 0.14 & 0.12 & 0.12 & 0.12 & 0.14 & 0.14 & 0.11 & 0.12 & 0.15 & 0.17 & 0.14 & 0.11 & 0.17 & 0.18 & 0.17 & 0.12 & 0.14 & 0.14 & 0.14 \\
Perm.\ aug. & 0.12 & 0.10 & 0.11 & 0.10 & 0.12 & 0.12 & 0.10 & 0.10 & 0.14 & 0.16 & 0.13 & 0.09 & 0.14 & 0.16 & 0.15 & 0.11 & 0.14 & 0.12 & 0.12 \\
DebiasFirst & 0.13 & 0.11 & 0.11 & 0.10 & 0.12 & 0.12 & 0.10 & 0.11 & 0.15 & 0.17 & 0.14 & 0.09 & 0.14 & 0.18 & 0.16 & 0.12 & 0.15 & 0.12 & 0.12 \\
OC-SFT & \textbf{0.09} & \textbf{0.07} & \textbf{0.07} & \textbf{0.06} & \textbf{0.08} & \textbf{0.08} & \textbf{0.07} & \textbf{0.07} & \textbf{0.09} & \textbf{0.13} & \textbf{0.09} & \textbf{0.06} & \textbf{0.10} & \textbf{0.10} & \textbf{0.11} & \textbf{0.08} & \textbf{0.10} & \textbf{0.08} & \textbf{0.08} \\
\bottomrule
\end{tabular*}
\end{table}

\FloatBarrier
\subsection{Alternative first stages}
\label{app:altfirststage}

\begin{table}[htbp]
\centering
\scriptsize
\caption{\textbf{Per-collection results under dense and learned-sparse first stages.}
Each variant has separate nDCG@10 and $\tau$-PSI columns.
The three trained variants use BM25 candidates during training; the BGE and SPLADE results therefore measure transfer to new first stages without retraining.
Higher quality and lower instability are better.
Basis: Qwen3-4B, seed 42 and first-stage-specific judged sets; the \textbf{Mean (13)} row reproduces the BGE and SPLADE rows in \Cref{tab:firststage}, whose first row reports BM25.}
\label{tab:firststage-perdataset}
\setlength{\tabcolsep}{1.5pt}
\resizebox{\textwidth}{!}{%
\begin{tabular}{lrrrrrrrrrrrrrrrr}
\toprule
& \multicolumn{8}{c}{BGE (dense)} & \multicolumn{8}{c}{SPLADE++ ED (learned-sparse)} \\
\cmidrule(lr){2-9}\cmidrule(lr){10-17}
& \multicolumn{2}{c}{Off the shelf} & \multicolumn{2}{c}{Single-order} & \multicolumn{2}{c}{Order-avg.} & \multicolumn{2}{c}{OC-SFT}
& \multicolumn{2}{c}{Off the shelf} & \multicolumn{2}{c}{Single-order} & \multicolumn{2}{c}{Order-avg.} & \multicolumn{2}{c}{OC-SFT} \\
\cmidrule(lr){2-3}\cmidrule(lr){4-5}\cmidrule(lr){6-7}\cmidrule(lr){8-9}
\cmidrule(lr){10-11}\cmidrule(lr){12-13}\cmidrule(lr){14-15}\cmidrule(lr){16-17}
Collection & nDCG & $\tau$-PSI & nDCG & $\tau$-PSI & nDCG & $\tau$-PSI & nDCG & $\tau$-PSI
& nDCG & $\tau$-PSI & nDCG & $\tau$-PSI & nDCG & $\tau$-PSI & nDCG & $\tau$-PSI \\
\midrule
DL19 & 0.573 & 0.336 & 0.740 & 0.202 & 0.758 & 0.131 & 0.740 & 0.073 & 0.586 & 0.333 & 0.739 & 0.209 & 0.727 & 0.134 & 0.741 & 0.075 \\
DL20 & 0.533 & 0.328 & 0.720 & 0.193 & 0.729 & 0.130 & 0.723 & 0.074 & 0.513 & 0.331 & 0.719 & 0.197 & 0.718 & 0.129 & 0.713 & 0.074 \\
Touch\'e-2020 & 0.219 & 0.300 & 0.262 & 0.187 & 0.267 & 0.113 & 0.273 & 0.070 & 0.193 & 0.287 & 0.247 & 0.178 & 0.241 & 0.114 & 0.246 & 0.079 \\
FiQA & 0.256 & 0.304 & 0.360 & 0.210 & 0.377 & 0.131 & 0.382 & 0.070 & 0.250 & 0.292 & 0.354 & 0.202 & 0.356 & 0.128 & 0.365 & 0.070 \\
NFCorpus & 0.281 & 0.324 & 0.355 & 0.245 & 0.369 & 0.190 & 0.371 & 0.118 & 0.266 & 0.317 & 0.331 & 0.235 & 0.347 & 0.186 & 0.344 & 0.117 \\
ArguAna & 0.211 & 0.308 & 0.250 & 0.239 & 0.292 & 0.175 & 0.316 & 0.127 & 0.205 & 0.295 & 0.239 & 0.230 & 0.280 & 0.172 & 0.310 & 0.127 \\
Climate-FEVER & 0.160 & 0.284 & 0.179 & 0.200 & 0.211 & 0.143 & 0.188 & 0.079 & 0.152 & 0.280 & 0.181 & 0.201 & 0.205 & 0.147 & 0.184 & 0.086 \\
DBPedia & 0.370 & 0.314 & 0.459 & 0.196 & 0.473 & 0.147 & 0.463 & 0.084 & 0.360 & 0.321 & 0.443 & 0.213 & 0.460 & 0.153 & 0.453 & 0.088 \\
SciFact & 0.529 & 0.319 & 0.703 & 0.244 & 0.713 & 0.180 & 0.718 & 0.103 & 0.511 & 0.311 & 0.693 & 0.235 & 0.706 & 0.176 & 0.711 & 0.104 \\
Signal-1M & 0.192 & 0.346 & 0.250 & 0.246 & 0.267 & 0.174 & 0.262 & 0.115 & 0.197 & 0.340 & 0.277 & 0.241 & 0.272 & 0.172 & 0.274 & 0.113 \\
TREC-COVID & 0.730 & 0.323 & 0.758 & 0.210 & 0.763 & 0.116 & 0.757 & 0.070 & 0.668 & 0.305 & 0.729 & 0.200 & 0.719 & 0.112 & 0.724 & 0.068 \\
TREC-NEWS & 0.335 & 0.310 & 0.412 & 0.199 & 0.426 & 0.137 & 0.420 & 0.080 & 0.298 & 0.296 & 0.400 & 0.194 & 0.412 & 0.138 & 0.407 & 0.084 \\
Robust04 & 0.462 & 0.293 & 0.557 & 0.201 & 0.567 & 0.142 & 0.555 & 0.091 & 0.460 & 0.285 & 0.563 & 0.193 & 0.571 & 0.139 & 0.568 & 0.089 \\
\midrule
\textbf{Mean (13)} & \textbf{0.373} & \textbf{0.315} & \textbf{0.462} & \textbf{0.213} & \textbf{0.478} & \textbf{0.147} & \textbf{0.474} & \textbf{0.089}
& \textbf{0.358} & \textbf{0.307} & \textbf{0.455} & \textbf{0.210} & \textbf{0.463} & \textbf{0.146} & \textbf{0.465} & \textbf{0.090} \\
\bottomrule
\end{tabular}}
\end{table}

\FloatBarrier
\subsection{Specialized and external rerankers}
\label{app:specialized}

\textbf{On Qwen3-Reranker-4B, OC-SFT matches order-averaged distillation in quality and lowers mean instability (\Cref{tab:specialized}).}
Serving \texttt{jina-reranker-v3} at matched $B=20$ lowers its quality and raises its $\tau$-PSI relative to its native single-context configuration.

\begin{table}[htbp]
\centering
\setlength{\tabcolsep}{0pt}
\caption{\textbf{Purpose-built rerankers and training variants on Qwen3-Reranker-4B.} 
Higher nDCG@10 and lower $\tau$-PSI are better.
RankZephyr and jina are listwise rerankers, whereas mxbai is pointwise and has zero $\tau$-PSI by construction.
The native jina row uses one context; the matched row uses the students' $B=20$ geometry.
Purpose-built systems use large hard-negative-mined reranking corpora, whereas OC-SFT uses about 30K MS MARCO queries, so they are external reference systems.
Basis: \emph{Public 16} excludes Legal-A and Legal-B.}
\label{tab:specialized}
\scriptsize
\begin{tabular*}{\textwidth}{l@{\extracolsep{\fill}}ccccccccccccccccccc}
\toprule
Method & \rothead{Public 16} & \rothead{DL19} & \rothead{DL20} & \rothead{DL21} & \rothead{DL22} & \rothead{DL23} & \rothead{Touch\'e} & \rothead{FiQA} & \rothead{NFCorpus} & \rothead{ArguAna} & \rothead{Climate} & \rothead{T-COVID} & \rothead{DBPedia} & \rothead{SciFact} & \rothead{Signal1M} & \rothead{T-NEWS} & \rothead{Robust04} & \rothead{Legal-A} & \rothead{Legal-B} \\[6pt]
\midrule
\multicolumn{20}{l}{\emph{nDCG@10, higher is better}} \\
RankZephyr-7B & 0.50 & 0.74 & 0.70 & 0.67 & 0.51 & 0.44 & 0.32 & 0.34 & 0.37 & 0.44 & 0.22 & 0.78 & 0.43 & 0.74 & 0.31 & 0.49 & 0.52 & 0.35 & 0.33 \\
jina-reranker-v3 & 0.50 & 0.72 & 0.70 & 0.69 & 0.50 & 0.45 & 0.30 & 0.38 & 0.37 & 0.68 & 0.24 & 0.74 & 0.43 & 0.68 & 0.31 & 0.40 & 0.49 & 0.35 & 0.32 \\
jina-reranker-v3 ($B=20$) & 0.47 & 0.69 & 0.65 & 0.63 & 0.48 & 0.42 & 0.25 & 0.35 & 0.34 & 0.65 & 0.23 & 0.73 & 0.39 & 0.64 & 0.25 & 0.35 & 0.45 & 0.27 & 0.26 \\
mxbai-base-v2 & 0.50 & 0.73 & 0.67 & 0.66 & 0.51 & 0.43 & 0.35 & 0.36 & 0.35 & 0.59 & 0.31 & 0.70 & 0.44 & 0.67 & 0.33 & 0.42 & 0.49 & 0.31 & 0.27 \\
mxbai-large-v2 & 0.52 & 0.73 & 0.69 & 0.69 & 0.50 & 0.46 & 0.33 & 0.39 & 0.36 & 0.67 & 0.35 & 0.74 & 0.45 & 0.73 & 0.31 & 0.47 & 0.51 & 0.33 & 0.32 \\
\midrule
\multicolumn{20}{l}{\emph{Qwen3-Reranker-4B}} \\
\quad Off the shelf & 0.46 & 0.65 & 0.59 & 0.53 & 0.40 & 0.34 & 0.29 & 0.37 & 0.37 & 0.52 & 0.29 & 0.76 & 0.41 & 0.73 & 0.24 & 0.38 & 0.49 & 0.29 & 0.24 \\
\quad Order-avg. & 0.51 & 0.75 & 0.69 & 0.66 & 0.49 & 0.46 & 0.30 & 0.40 & 0.39 & 0.53 & 0.29 & 0.79 & 0.46 & 0.73 & 0.30 & 0.44 & 0.53 & 0.37 & 0.32 \\
\quad OC-SFT & 0.51 & 0.74 & 0.69 & 0.67 & 0.50 & 0.46 & 0.32 & 0.40 & 0.39 & 0.47 & 0.28 & 0.79 & 0.47 & 0.72 & 0.32 & 0.44 & 0.52 & 0.36 & 0.31 \\
\midrule
\multicolumn{20}{l}{\emph{$\tau$-PSI, lower is better}} \\
RankZephyr-7B & 0.40 & 0.39 & 0.39 & 0.40 & 0.40 & 0.40 & 0.41 & 0.41 & 0.38 & 0.40 & 0.41 & 0.40 & 0.41 & 0.41 & 0.43 & 0.40 & 0.40 & 0.41 & 0.41 \\
jina-reranker-v3 & 0.16 & 0.15 & 0.16 & 0.13 & 0.15 & 0.16 & 0.13 & 0.15 & 0.17 & 0.16 & 0.16 & 0.14 & 0.18 & 0.20 & 0.18 & 0.15 & 0.17 & 0.16 & 0.17 \\
jina-reranker-v3 ($B=20$) & 0.18 & 0.17 & 0.17 & 0.16 & 0.18 & 0.18 & 0.15 & 0.17 & 0.18 & 0.18 & 0.18 & 0.15 & 0.20 & 0.20 & 0.20 & 0.17 & 0.18 & 0.18 & 0.18 \\
\midrule
\multicolumn{20}{l}{\emph{Qwen3-Reranker-4B}} \\
\quad Off the shelf & 0.29 & 0.29 & 0.29 & 0.31 & 0.30 & 0.32 & 0.29 & 0.26 & 0.31 & 0.30 & 0.25 & 0.27 & 0.30 & 0.30 & 0.33 & 0.27 & 0.29 & 0.32 & 0.31 \\
\quad Order-avg. & 0.12 & 0.11 & 0.11 & 0.10 & 0.11 & 0.12 & 0.10 & 0.10 & 0.15 & 0.14 & 0.11 & 0.09 & 0.14 & 0.14 & 0.16 & 0.11 & 0.13 & 0.13 & 0.12 \\
\quad OC-SFT & 0.08 & 0.07 & 0.08 & 0.07 & 0.08 & 0.08 & 0.06 & 0.06 & 0.10 & 0.13 & 0.07 & 0.06 & 0.09 & 0.09 & 0.12 & 0.07 & 0.09 & 0.08 & 0.08 \\
\bottomrule
\end{tabular*}
\end{table}

\FloatBarrier
\subsection{Cross-base summaries}
\label{app:crossbase}

\begin{figure}[htbp]
\centering
\includegraphics[width=\linewidth]{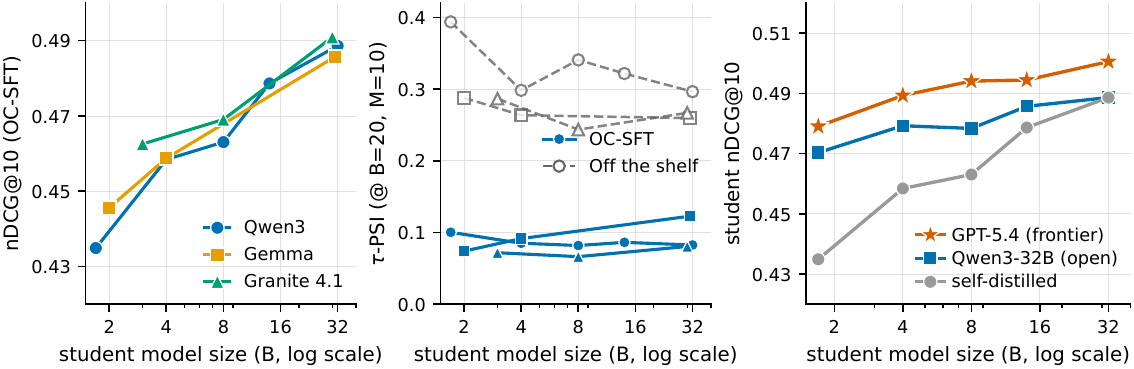}
\caption{\textbf{Model size, family and teacher identity.}
Panels (a) and (b) show mean nDCG@10 and $\tau$-PSI over the 18 reranking collections, respectively, by model size for Qwen3, Gemma and Granite.
Gemma E2B and E4B are plotted at their effective sizes, 2B and 4B.
Panel (c) compares nDCG@10 across the evaluated Qwen3 student sizes for frontier, open and self-distilled teachers.}
\label{fig:sizefam}
\end{figure}

\begin{figure}[htbp]
\centering
\includegraphics[width=0.78\linewidth]{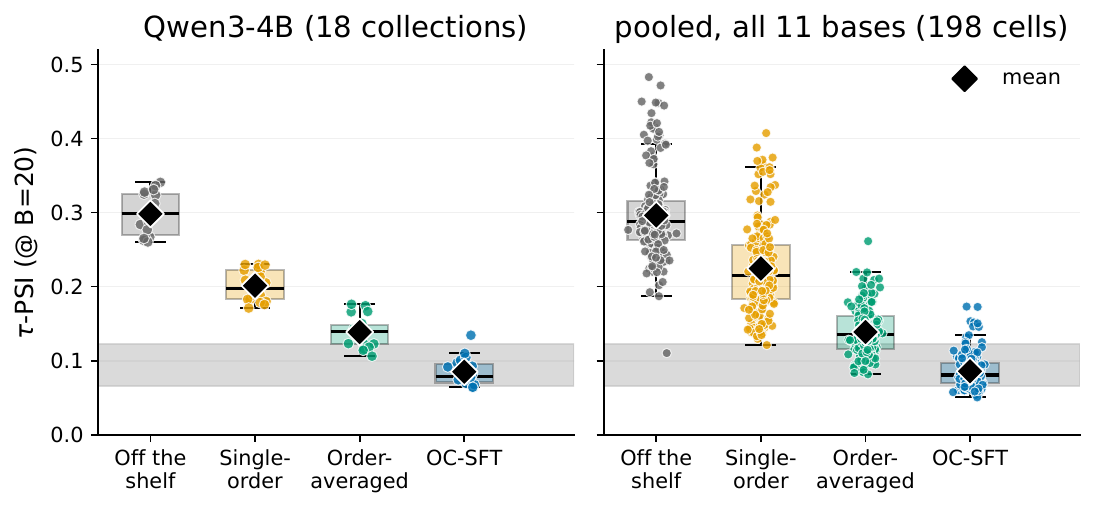}
\caption{\textbf{Per-variant $\tau$-PSI spread across collections.}
Panel (a) shows Qwen3-4B across the 18 reranking collections.
Panel (b) pools the same collections across all 11 dense bases, for 198 model-collection cells.
Diamonds mark means; the shaded region spans the range of model-level mean OC-SFT $\tau$-PSI values.
The sparse MoE is excluded from this figure and reported in \Cref{tab:grids-tau}.
}
\label{fig:taupsibox}
\end{figure}

\textbf{Across all twelve bases, OC-SFT trained from single-order labels is more stable than order-averaged distillation (\Cref{fig:sizefam,fig:taupsibox,tab:grids-tau}).}
OC-SFT trained from order-averaged labels reduces instability further on ten of the eleven dense bases and on the sparse MoE.
Qwen3-0.6B is omitted because its off-the-shelf output is degenerate and does not follow the required format.

\textbf{An open teacher approaches GPT-5.4 across student sizes (\Cref{fig:sizefam}, panel c).}
Using Qwen3-32B as the teacher keeps student nDCG@10 within 0.016 of GPT-5.4 and matches or exceeds self-distillation at every evaluated Qwen3 student size.
Compared with self-distillation, the open teacher improves nDCG@10 by about 0.04 at Qwen3-1.7B, and the difference narrows as student size increases.
GPT-5.4 remains highest at every size, but the open teacher closes most of that quality gap without using GPT-5.4 labels.
Therefore, changing the teacher affects nDCG@10 more than $\tau$-PSI.

\begin{figure}[htbp]
\centering
\includegraphics[width=0.42\linewidth]{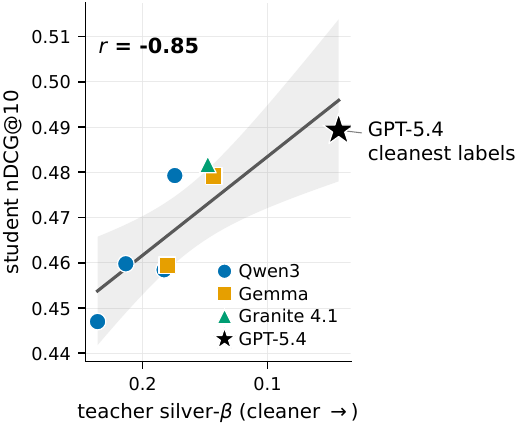}
\caption{\textbf{Student quality correlates with label order-consistency across teachers.}
Student nDCG@10 plotted against teacher silver-$\beta$ for eight teachers. %
The correlation with student nDCG@10 is $-0.852$ for silver-$\beta$ and $+0.812$ for silver quality; they correlate at $-0.528$.
Shading marks the 95\% confidence region for the fitted relationship between silver-$\beta$ and student nDCG@10.
}
\label{fig:teacherclean}
\end{figure}

\FloatBarrier
\enlargethispage{2\baselineskip}
\textbf{Teacher selection should consider label order-consistency as well as label quality (\Cref{fig:teacherclean}).}
Silver-$\beta$ measures how much a document's silver grade shifts when the teacher sees shuffled candidate orders; lower values denote more order-consistent labels.
Formally, let $\mu_t(q,d)$ be a document's mean raw grade over $T{=}10$ teacher permutations and $\bar\delta_p$ the mean of $s^{(k)}_{q,d}-\mu_t(q,d)$ among its occurrences at slot $p$ of the grading prompt; then
\begingroup
\abovedisplayskip=3pt
\belowdisplayskip=3pt
\[
  \text{silver-}\beta = \sqrt{\tfrac{1}{B}\textstyle\sum_{p=0}^{B-1}\bar\delta_p^{\,2}},
\]
\endgroup
which is 0 when grades do not depend on order. It is a position-conditioned residual in grade units, not a rank correlation.
Across the 8 teachers, silver-$\beta$ and silver quality each correlate with student nDCG@10, but they also correlate with each other, so we cannot isolate either effect alone.
The within-teacher control in \Cref{app:silver} varies label order-consistency while holding silver quality nearly fixed and produces a more stable student.
Together, these results show why teacher evaluation should report both quality and order-consistency of its silver labels.\looseness=-1

\vspace{-6pt}

\FloatBarrier

\begin{landscape}
\scriptsize
\setlength{\tabcolsep}{1pt}
\begin{longtable}{lrrrrrrrrrrrrrrrrrrrr}
\caption{\textbf{Per-collection nDCG@10 for every base and training variant.} Twelve bases, eleven dense and one sparse Mixture-of-Experts, each with its four training variants; higher is better. The final columns report means over the 16 public and all 18 reranking collections. Every block carries a fifth row, OC-SFT trained on order-averaged rather than single-order teacher labels. The first stage is BM25 on the 16 public collections. Legal-A and Legal-B use the precomputed candidate lists packaged with their internal fixtures; the retriever identity is not recorded. Basis: all 18 reranking collections, seed 42, one independently selected checkpoint per base and variant; off-the-shelf rows evaluate fixed base checkpoints. Order instability for the same cells is in \Cref{tab:grids-tau}.}\label{tab:grids-ndcg}\\
\toprule
Variant & DL19 & DL20 & DL21 & DL22 & DL23 & Touch\'e-2020 & FiQA & NFCorpus & ArguAna & Climate-FEVER & TREC-COVID & DBPedia & SciFact & Signal-1M & TREC-NEWS & Robust04 & Legal-A & Legal-B & Public 16 & All 18 \\
\midrule
\endfirsthead
\multicolumn{21}{l}{\emph{\Cref{tab:grids-ndcg}, continued.}}\\
\toprule
Variant & DL19 & DL20 & DL21 & DL22 & DL23 & Touch\'e-2020 & FiQA & NFCorpus & ArguAna & Climate-FEVER & TREC-COVID & DBPedia & SciFact & Signal-1M & TREC-NEWS & Robust04 & Legal-A & Legal-B & Public 16 & All 18 \\
\midrule
\endhead
\midrule
\multicolumn{21}{r}{\emph{continued on the next page}}\\
\endfoot
\bottomrule
\endlastfoot
\multicolumn{21}{l}{\textbf{Qwen3-1.7B}}\\
\quad Off the shelf & 0.479 & 0.435 & 0.452 & 0.288 & 0.249 & 0.208 & 0.213 & 0.301 & 0.318 & 0.142 & 0.675 & 0.258 & 0.526 & 0.210 & 0.307 & 0.399 & 0.233 & 0.209 & 0.341 & 0.328 \\
\quad Single-order & 0.586 & 0.547 & 0.545 & 0.347 & 0.317 & 0.246 & 0.215 & 0.289 & 0.262 & 0.165 & 0.703 & 0.315 & 0.525 & 0.232 & 0.352 & 0.430 & 0.180 & 0.174 & 0.380 & 0.357 \\
\quad Order-avg. & 0.678 & 0.615 & 0.630 & 0.405 & 0.367 & 0.282 & 0.288 & 0.336 & 0.344 & 0.177 & 0.736 & 0.379 & 0.633 & 0.296 & 0.405 & 0.474 & 0.287 & 0.279 & 0.440 & 0.423 \\
\quad OC-SFT & 0.695 & 0.633 & 0.662 & 0.427 & 0.391 & 0.290 & 0.296 & 0.347 & 0.286 & 0.202 & 0.728 & 0.417 & 0.638 & 0.292 & 0.432 & 0.495 & 0.305 & 0.291 & 0.452 & 0.435 \\
\quad OC-SFT (avg. labels) & 0.678 & 0.622 & 0.646 & 0.421 & 0.375 & 0.279 & 0.301 & 0.349 & 0.306 & 0.190 & 0.751 & 0.393 & 0.657 & 0.296 & 0.417 & 0.477 & 0.277 & 0.260 & 0.447 & 0.427 \\
\midrule
\multicolumn{21}{l}{\textbf{Qwen3-4B}}\\
\quad Off the shelf & 0.596 & 0.552 & 0.542 & 0.389 & 0.331 & 0.221 & 0.251 & 0.312 & 0.206 & 0.157 & 0.724 & 0.334 & 0.512 & 0.217 & 0.330 & 0.450 & 0.286 & 0.243 & 0.383 & 0.370 \\
\quad Single-order & 0.719 & 0.675 & 0.670 & 0.448 & 0.430 & 0.258 & 0.346 & 0.353 & 0.257 & 0.189 & 0.753 & 0.403 & 0.696 & 0.293 & 0.423 & 0.516 & 0.338 & 0.290 & 0.464 & 0.448 \\
\quad Order-avg. & 0.727 & 0.675 & 0.674 & 0.453 & 0.437 & 0.286 & 0.349 & 0.364 & 0.283 & 0.208 & 0.747 & 0.413 & 0.708 & 0.294 & 0.443 & 0.520 & 0.359 & 0.310 & 0.474 & 0.458 \\
\quad OC-SFT & 0.732 & 0.681 & 0.676 & 0.466 & 0.448 & 0.276 & 0.365 & 0.357 & 0.311 & 0.189 & 0.740 & 0.417 & 0.714 & 0.295 & 0.423 & 0.515 & 0.352 & 0.306 & 0.475 & 0.459 \\
\quad OC-SFT (avg. labels) & 0.728 & 0.681 & 0.670 & 0.464 & 0.451 & 0.274 & 0.359 & 0.356 & 0.302 & 0.187 & 0.750 & 0.419 & 0.714 & 0.298 & 0.427 & 0.517 & 0.332 & 0.294 & 0.475 & 0.457 \\
\midrule
\multicolumn{21}{l}{\textbf{Qwen3-8B}}\\
\quad Off the shelf & 0.611 & 0.533 & 0.505 & 0.362 & 0.327 & 0.229 & 0.310 & 0.318 & 0.235 & 0.141 & 0.718 & 0.329 & 0.602 & 0.208 & 0.327 & 0.453 & 0.265 & 0.241 & 0.388 & 0.373 \\
\quad Single-order & 0.705 & 0.659 & 0.640 & 0.444 & 0.424 & 0.261 & 0.357 & 0.341 & 0.325 & 0.162 & 0.738 & 0.397 & 0.705 & 0.271 & 0.403 & 0.507 & 0.320 & 0.261 & 0.459 & 0.440 \\
\quad Order-avg. & 0.713 & 0.689 & 0.668 & 0.466 & 0.449 & 0.253 & 0.365 & 0.363 & 0.299 & 0.162 & 0.749 & 0.409 & 0.722 & 0.281 & 0.419 & 0.522 & 0.329 & 0.289 & 0.471 & 0.453 \\
\quad OC-SFT & 0.729 & 0.702 & 0.683 & 0.482 & 0.462 & 0.278 & 0.380 & 0.360 & 0.320 & 0.178 & 0.750 & 0.416 & 0.723 & 0.282 & 0.418 & 0.528 & 0.353 & 0.291 & 0.481 & 0.463 \\
\quad OC-SFT (avg. labels) & 0.727 & 0.687 & 0.673 & 0.477 & 0.448 & 0.262 & 0.374 & 0.362 & 0.335 & 0.166 & 0.762 & 0.413 & 0.725 & 0.282 & 0.421 & 0.523 & 0.353 & 0.278 & 0.477 & 0.459 \\
\midrule
\multicolumn{21}{l}{\textbf{Qwen3-14B}}\\
\quad Off the shelf & 0.642 & 0.587 & 0.587 & 0.405 & 0.370 & 0.214 & 0.357 & 0.338 & 0.340 & 0.143 & 0.748 & 0.342 & 0.712 & 0.228 & 0.354 & 0.491 & 0.278 & 0.237 & 0.429 & 0.410 \\
\quad Single-order & 0.718 & 0.694 & 0.676 & 0.463 & 0.440 & 0.271 & 0.400 & 0.362 & 0.455 & 0.172 & 0.752 & 0.407 & 0.729 & 0.304 & 0.438 & 0.533 & 0.342 & 0.280 & 0.488 & 0.469 \\
\quad Order-avg. & 0.733 & 0.702 & 0.699 & 0.482 & 0.447 & 0.260 & 0.399 & 0.375 & 0.445 & 0.156 & 0.780 & 0.415 & 0.744 & 0.298 & 0.465 & 0.553 & 0.339 & 0.286 & 0.497 & 0.477 \\
\quad OC-SFT & 0.749 & 0.709 & 0.702 & 0.490 & 0.462 & 0.282 & 0.405 & 0.376 & 0.358 & 0.171 & 0.780 & 0.430 & 0.740 & 0.301 & 0.465 & 0.557 & 0.342 & 0.298 & 0.499 & 0.479 \\
\quad OC-SFT (avg. labels) & 0.736 & 0.694 & 0.703 & 0.487 & 0.451 & 0.267 & 0.404 & 0.381 & 0.347 & 0.171 & 0.794 & 0.420 & 0.739 & 0.299 & 0.462 & 0.557 & 0.342 & 0.291 & 0.495 & 0.475 \\
\midrule
\multicolumn{21}{l}{\textbf{Qwen3-32B}}\\
\quad Off the shelf & 0.628 & 0.581 & 0.569 & 0.401 & 0.334 & 0.253 & 0.319 & 0.331 & 0.332 & 0.149 & 0.741 & 0.356 & 0.708 & 0.222 & 0.348 & 0.486 & 0.278 & 0.248 & 0.422 & 0.405 \\
\quad Single-order & 0.706 & 0.674 & 0.660 & 0.469 & 0.425 & 0.259 & 0.392 & 0.342 & 0.430 & 0.158 & 0.741 & 0.407 & 0.731 & 0.283 & 0.444 & 0.530 & 0.329 & 0.284 & 0.478 & 0.459 \\
\quad Order-avg. & 0.741 & 0.702 & 0.690 & 0.493 & 0.444 & 0.265 & 0.406 & 0.377 & 0.460 & 0.172 & 0.788 & 0.425 & 0.745 & 0.285 & 0.469 & 0.555 & 0.338 & 0.299 & 0.501 & 0.481 \\
\quad OC-SFT & 0.748 & 0.712 & 0.694 & 0.507 & 0.460 & 0.276 & 0.417 & 0.380 & 0.464 & 0.177 & 0.776 & 0.435 & 0.743 & 0.288 & 0.493 & 0.570 & 0.350 & 0.302 & 0.509 & 0.488 \\
\quad OC-SFT (avg. labels) & 0.742 & 0.702 & 0.694 & 0.502 & 0.451 & 0.267 & 0.413 & 0.385 & 0.449 & 0.174 & 0.783 & 0.431 & 0.739 & 0.290 & 0.467 & 0.560 & 0.338 & 0.293 & 0.503 & 0.482 \\
\midrule
\multicolumn{21}{l}{\textbf{Gemma-E2B}}\\
\quad Off the shelf & 0.586 & 0.518 & 0.581 & 0.365 & 0.312 & 0.224 & 0.235 & 0.296 & 0.296 & 0.143 & 0.699 & 0.316 & 0.526 & 0.213 & 0.307 & 0.443 & 0.280 & 0.225 & 0.379 & 0.365 \\
\quad Single-order & 0.715 & 0.644 & 0.641 & 0.412 & 0.415 & 0.260 & 0.306 & 0.336 & 0.363 & 0.182 & 0.752 & 0.388 & 0.629 & 0.293 & 0.408 & 0.498 & 0.332 & 0.284 & 0.453 & 0.437 \\
\quad Order-avg. & 0.709 & 0.644 & 0.656 & 0.433 & 0.429 & 0.255 & 0.311 & 0.338 & 0.404 & 0.186 & 0.748 & 0.385 & 0.654 & 0.309 & 0.397 & 0.496 & 0.355 & 0.301 & 0.460 & 0.445 \\
\quad OC-SFT & 0.721 & 0.638 & 0.670 & 0.433 & 0.431 & 0.263 & 0.323 & 0.340 & 0.356 & 0.183 & 0.757 & 0.392 & 0.648 & 0.316 & 0.403 & 0.496 & 0.361 & 0.291 & 0.461 & 0.446 \\
\quad OC-SFT (avg. labels) & 0.727 & 0.634 & 0.659 & 0.442 & 0.427 & 0.277 & 0.317 & 0.340 & 0.358 & 0.185 & 0.751 & 0.400 & 0.661 & 0.311 & 0.417 & 0.502 & 0.356 & 0.293 & 0.463 & 0.448 \\
\midrule
\multicolumn{21}{l}{\textbf{Gemma-E4B}}\\
\quad Off the shelf & 0.632 & 0.593 & 0.616 & 0.403 & 0.369 & 0.215 & 0.303 & 0.327 & 0.221 & 0.124 & 0.747 & 0.343 & 0.595 & 0.219 & 0.353 & 0.499 & 0.305 & 0.240 & 0.410 & 0.395 \\
\quad Single-order & 0.719 & 0.679 & 0.688 & 0.458 & 0.437 & 0.242 & 0.344 & 0.354 & 0.342 & 0.138 & 0.768 & 0.393 & 0.702 & 0.309 & 0.433 & 0.540 & 0.366 & 0.295 & 0.472 & 0.456 \\
\quad Order-avg. & 0.727 & 0.680 & 0.688 & 0.472 & 0.446 & 0.256 & 0.353 & 0.364 & 0.392 & 0.152 & 0.764 & 0.404 & 0.709 & 0.310 & 0.448 & 0.541 & 0.362 & 0.298 & 0.482 & 0.465 \\
\quad OC-SFT & 0.695 & 0.665 & 0.669 & 0.463 & 0.439 & 0.252 & 0.357 & 0.364 & 0.365 & 0.178 & 0.771 & 0.406 & 0.706 & 0.293 & 0.446 & 0.551 & 0.351 & 0.287 & 0.476 & 0.459 \\
\quad OC-SFT (avg. labels) & 0.727 & 0.688 & 0.683 & 0.473 & 0.439 & 0.265 & 0.364 & 0.358 & 0.400 & 0.153 & 0.776 & 0.413 & 0.707 & 0.306 & 0.425 & 0.541 & 0.345 & 0.295 & 0.482 & 0.464 \\
\midrule
\multicolumn{21}{l}{\textbf{Gemma-31B}}\\
\quad Off the shelf & 0.642 & 0.632 & 0.624 & 0.465 & 0.416 & 0.242 & 0.413 & 0.355 & 0.413 & 0.163 & 0.756 & 0.368 & 0.720 & 0.226 & 0.371 & 0.543 & 0.354 & 0.281 & 0.459 & 0.444 \\
\quad Single-order & 0.702 & 0.698 & 0.684 & 0.497 & 0.468 & 0.270 & 0.419 & 0.370 & 0.435 & 0.168 & 0.750 & 0.407 & 0.746 & 0.292 & 0.442 & 0.575 & 0.371 & 0.300 & 0.495 & 0.477 \\
\quad Order-avg. & 0.722 & 0.711 & 0.706 & 0.506 & 0.480 & 0.277 & 0.420 & 0.377 & 0.464 & 0.191 & 0.751 & 0.419 & 0.741 & 0.311 & 0.463 & 0.572 & 0.386 & 0.319 & 0.507 & 0.490 \\
\quad OC-SFT & 0.733 & 0.710 & 0.704 & 0.521 & 0.482 & 0.283 & 0.430 & 0.384 & 0.340 & 0.184 & 0.752 & 0.427 & 0.749 & 0.295 & 0.473 & 0.578 & 0.381 & 0.316 & 0.503 & 0.486 \\
\quad OC-SFT (avg. labels) & 0.730 & 0.708 & 0.706 & 0.526 & 0.488 & 0.279 & 0.424 & 0.380 & 0.479 & 0.181 & 0.758 & 0.421 & 0.741 & 0.286 & 0.457 & 0.578 & 0.378 & 0.310 & 0.509 & 0.491 \\
\midrule
\multicolumn{21}{l}{\textbf{Granite-3B}}\\
\quad Off the shelf & 0.586 & 0.510 & 0.539 & 0.380 & 0.311 & 0.221 & 0.276 & 0.323 & 0.177 & 0.176 & 0.708 & 0.322 & 0.573 & 0.231 & 0.358 & 0.443 & 0.279 & 0.246 & 0.383 & 0.370 \\
\quad Single-order & 0.715 & 0.667 & 0.665 & 0.449 & 0.419 & 0.273 & 0.319 & 0.347 & 0.255 & 0.211 & 0.753 & 0.407 & 0.667 & 0.321 & 0.430 & 0.506 & 0.320 & 0.283 & 0.463 & 0.445 \\
\quad Order-avg. & 0.728 & 0.662 & 0.676 & 0.460 & 0.435 & 0.262 & 0.332 & 0.357 & 0.268 & 0.196 & 0.747 & 0.401 & 0.687 & 0.307 & 0.438 & 0.511 & 0.326 & 0.302 & 0.467 & 0.450 \\
\quad OC-SFT & 0.738 & 0.677 & 0.680 & 0.464 & 0.428 & 0.288 & 0.338 & 0.360 & 0.349 & 0.219 & 0.756 & 0.422 & 0.693 & 0.326 & 0.445 & 0.527 & 0.317 & 0.298 & 0.482 & 0.462 \\
\quad OC-SFT (avg. labels) & 0.731 & 0.665 & 0.676 & 0.464 & 0.422 & 0.263 & 0.339 & 0.365 & 0.317 & 0.208 & 0.759 & 0.412 & 0.692 & 0.322 & 0.440 & 0.520 & 0.320 & 0.302 & 0.475 & 0.456 \\
\midrule
\multicolumn{21}{l}{\textbf{Granite-8B}}\\
\quad Off the shelf & 0.637 & 0.587 & 0.584 & 0.402 & 0.376 & 0.240 & 0.325 & 0.354 & 0.277 & 0.153 & 0.752 & 0.364 & 0.683 & 0.240 & 0.397 & 0.483 & 0.296 & 0.258 & 0.428 & 0.412 \\
\quad Single-order & 0.721 & 0.678 & 0.681 & 0.462 & 0.447 & 0.263 & 0.356 & 0.372 & 0.278 & 0.174 & 0.764 & 0.410 & 0.720 & 0.306 & 0.467 & 0.549 & 0.310 & 0.264 & 0.478 & 0.457 \\
\quad Order-avg. & 0.720 & 0.675 & 0.682 & 0.477 & 0.454 & 0.264 & 0.355 & 0.374 & 0.309 & 0.175 & 0.764 & 0.414 & 0.719 & 0.301 & 0.468 & 0.552 & 0.334 & 0.295 & 0.481 & 0.463 \\
\quad OC-SFT & 0.720 & 0.671 & 0.688 & 0.488 & 0.467 & 0.273 & 0.369 & 0.373 & 0.332 & 0.178 & 0.768 & 0.419 & 0.727 & 0.319 & 0.475 & 0.553 & 0.329 & 0.294 & 0.489 & 0.469 \\
\quad OC-SFT (avg. labels) & 0.724 & 0.676 & 0.692 & 0.474 & 0.464 & 0.267 & 0.371 & 0.375 & 0.342 & 0.172 & 0.772 & 0.419 & 0.722 & 0.316 & 0.475 & 0.551 & 0.333 & 0.298 & 0.488 & 0.469 \\
\midrule
\multicolumn{21}{l}{\textbf{Granite-30B}}\\
\quad Off the shelf & 0.659 & 0.611 & 0.595 & 0.429 & 0.391 & 0.241 & 0.378 & 0.345 & 0.283 & 0.145 & 0.761 & 0.373 & 0.702 & 0.231 & 0.382 & 0.517 & 0.310 & 0.253 & 0.440 & 0.423 \\
\quad Single-order & 0.722 & 0.681 & 0.673 & 0.488 & 0.457 & 0.271 & 0.388 & 0.361 & 0.393 & 0.155 & 0.766 & 0.428 & 0.730 & 0.299 & 0.455 & 0.551 & 0.351 & 0.300 & 0.489 & 0.471 \\
\quad Order-avg. & 0.745 & 0.702 & 0.712 & 0.500 & 0.476 & 0.296 & 0.406 & 0.380 & 0.387 & 0.167 & 0.786 & 0.431 & 0.748 & 0.306 & 0.458 & 0.570 & 0.359 & 0.309 & 0.504 & 0.485 \\
\quad OC-SFT & 0.746 & 0.717 & 0.707 & 0.510 & 0.480 & 0.304 & 0.413 & 0.380 & 0.410 & 0.180 & 0.782 & 0.444 & 0.747 & 0.311 & 0.479 & 0.574 & 0.349 & 0.300 & 0.511 & 0.491 \\
\quad OC-SFT (avg. labels) & 0.745 & 0.710 & 0.711 & 0.503 & 0.473 & 0.311 & 0.410 & 0.383 & 0.439 & 0.173 & 0.774 & 0.438 & 0.740 & 0.312 & 0.474 & 0.570 & 0.342 & 0.296 & 0.510 & 0.489 \\
\midrule
\multicolumn{21}{l}{\textbf{Gemma-4 26B-A4B (sparse MoE)}}\\
\quad Off the shelf & 0.674 & 0.658 & 0.628 & 0.471 & 0.415 & 0.246 & 0.398 & 0.346 & 0.353 & 0.147 & 0.749 & 0.361 & 0.692 & 0.239 & 0.379 & 0.547 & 0.339 & 0.276 & 0.456 & 0.440 \\
\quad Single-order & 0.715 & 0.702 & 0.682 & 0.503 & 0.461 & 0.264 & 0.414 & 0.369 & 0.434 & 0.169 & 0.752 & 0.405 & 0.741 & 0.300 & 0.443 & 0.569 & 0.379 & 0.304 & 0.495 & 0.478 \\
\quad Order-avg. & 0.720 & 0.709 & 0.690 & 0.508 & 0.469 & 0.279 & 0.415 & 0.376 & 0.457 & 0.166 & 0.761 & 0.416 & 0.744 & 0.314 & 0.441 & 0.574 & 0.375 & 0.315 & 0.502 & 0.485 \\
\quad OC-SFT ($\lambda=3$) & 0.720 & 0.703 & 0.693 & 0.515 & 0.483 & 0.278 & 0.419 & 0.380 & 0.478 & 0.183 & 0.748 & 0.406 & 0.744 & 0.303 & 0.442 & 0.575 & 0.379 & 0.316 & 0.504 & 0.487 \\
\quad OC-SFT (avg. labels, $\lambda=5$) & 0.728 & 0.702 & 0.690 & 0.518 & 0.489 & 0.283 & 0.418 & 0.378 & 0.461 & 0.173 & 0.758 & 0.410 & 0.743 & 0.306 & 0.451 & 0.577 & 0.372 & 0.303 & 0.505 & 0.487 \\
\end{longtable}
\end{landscape}

\begin{landscape}
\scriptsize
\setlength{\tabcolsep}{1pt}
\begin{longtable}{lrrrrrrrrrrrrrrrrrrr}
\caption{\textbf{Per-collection $\tau$-PSI for every base and training variant.} The same twelve bases and 18 per-collection cells as \Cref{tab:grids-ndcg}; lower is more stable. The \emph{All 18} column averages all 18 reranking collections. Every block also includes OC-SFT trained on order-averaged labels. Basis: all 18 reranking collections, seed 42, $M=10$ random permutations, one independently selected checkpoint per base and variant.}\label{tab:grids-tau}\\
\toprule
Variant & DL19 & DL20 & DL21 & DL22 & DL23 & Touch\'e-2020 & FiQA & NFCorpus & ArguAna & Climate-FEVER & TREC-COVID & DBPedia & SciFact & Signal-1M & TREC-NEWS & Robust04 & Legal-A & Legal-B & All 18 \\
\midrule
\endfirsthead
\multicolumn{20}{l}{\emph{\Cref{tab:grids-tau}, continued.}}\\
\toprule
Variant & DL19 & DL20 & DL21 & DL22 & DL23 & Touch\'e-2020 & FiQA & NFCorpus & ArguAna & Climate-FEVER & TREC-COVID & DBPedia & SciFact & Signal-1M & TREC-NEWS & Robust04 & Legal-A & Legal-B & All 18 \\
\midrule
\endhead
\midrule
\multicolumn{20}{r}{\emph{continued on the next page}}\\
\endfoot
\bottomrule
\endlastfoot
\multicolumn{20}{l}{\textbf{Qwen3-1.7B}}\\
\quad Off the shelf & 0.393 & 0.387 & 0.405 & 0.404 & 0.403 & 0.413 & 0.415 & 0.383 & 0.376 & 0.342 & 0.392 & 0.399 & 0.377 & 0.373 & 0.367 & 0.365 & 0.448 & 0.445 & 0.394 \\
\quad Single-order & 0.329 & 0.324 & 0.336 & 0.352 & 0.352 & 0.315 & 0.361 & 0.355 & 0.365 & 0.388 & 0.337 & 0.361 & 0.407 & 0.357 & 0.334 & 0.370 & 0.371 & 0.374 & 0.355 \\
\quad Order-avg. & 0.148 & 0.148 & 0.149 & 0.163 & 0.163 & 0.128 & 0.144 & 0.179 & 0.198 & 0.174 & 0.137 & 0.180 & 0.211 & 0.181 & 0.135 & 0.177 & 0.172 & 0.173 & 0.164 \\
\quad OC-SFT & 0.094 & 0.099 & 0.084 & 0.097 & 0.105 & 0.074 & 0.081 & 0.101 & 0.153 & 0.096 & 0.079 & 0.104 & 0.110 & 0.132 & 0.087 & 0.105 & 0.106 & 0.101 & 0.100 \\
\quad OC-SFT (avg. labels) & 0.065 & 0.067 & 0.062 & 0.071 & 0.071 & 0.057 & 0.063 & 0.080 & 0.114 & 0.077 & 0.062 & 0.085 & 0.081 & 0.096 & 0.064 & 0.082 & 0.072 & 0.071 & 0.074 \\
\midrule
\multicolumn{20}{l}{\textbf{Qwen3-4B}}\\
\quad Off the shelf & 0.302 & 0.296 & 0.312 & 0.325 & 0.328 & 0.288 & 0.263 & 0.284 & 0.267 & 0.265 & 0.277 & 0.324 & 0.305 & 0.331 & 0.265 & 0.260 & 0.341 & 0.336 & 0.298 \\
\quad Single-order & 0.188 & 0.180 & 0.183 & 0.209 & 0.205 & 0.183 & 0.179 & 0.213 & 0.221 & 0.189 & 0.176 & 0.222 & 0.230 & 0.229 & 0.171 & 0.189 & 0.230 & 0.226 & 0.201 \\
\quad Order-avg. & 0.123 & 0.123 & 0.116 & 0.139 & 0.137 & 0.114 & 0.118 & 0.150 & 0.174 & 0.143 & 0.106 & 0.166 & 0.176 & 0.166 & 0.123 & 0.141 & 0.142 & 0.139 & 0.139 \\
\quad OC-SFT & 0.069 & 0.074 & 0.064 & 0.076 & 0.081 & 0.068 & 0.068 & 0.093 & 0.134 & 0.092 & 0.064 & 0.097 & 0.104 & 0.110 & 0.078 & 0.101 & 0.079 & 0.079 & 0.085 \\
\quad OC-SFT (avg. labels) & 0.062 & 0.066 & 0.059 & 0.070 & 0.070 & 0.061 & 0.060 & 0.079 & 0.125 & 0.080 & 0.058 & 0.088 & 0.095 & 0.097 & 0.065 & 0.082 & 0.066 & 0.065 & 0.075 \\
\midrule
\multicolumn{20}{l}{\textbf{Qwen3-8B}}\\
\quad Off the shelf & 0.316 & 0.304 & 0.326 & 0.323 & 0.325 & 0.303 & 0.262 & 0.289 & 0.240 & 0.244 & 0.301 & 0.315 & 0.269 & 0.321 & 0.259 & 0.257 & 0.318 & 0.321 & 0.294 \\
\quad Single-order & 0.256 & 0.240 & 0.240 & 0.257 & 0.252 & 0.246 & 0.253 & 0.296 & 0.267 & 0.282 & 0.240 & 0.288 & 0.316 & 0.278 & 0.249 & 0.268 & 0.270 & 0.270 & 0.265 \\
\quad Order-avg. & 0.160 & 0.150 & 0.140 & 0.167 & 0.159 & 0.132 & 0.145 & 0.189 & 0.187 & 0.167 & 0.130 & 0.190 & 0.201 & 0.185 & 0.140 & 0.162 & 0.174 & 0.169 & 0.164 \\
\quad OC-SFT & 0.073 & 0.076 & 0.067 & 0.078 & 0.079 & 0.062 & 0.065 & 0.090 & 0.125 & 0.085 & 0.060 & 0.097 & 0.103 & 0.108 & 0.069 & 0.091 & 0.075 & 0.072 & 0.082 \\
\quad OC-SFT (avg. labels) & 0.065 & 0.066 & 0.062 & 0.068 & 0.070 & 0.058 & 0.056 & 0.075 & 0.106 & 0.070 & 0.056 & 0.084 & 0.080 & 0.089 & 0.061 & 0.076 & 0.064 & 0.063 & 0.071 \\
\midrule
\multicolumn{20}{l}{\textbf{Qwen3-14B}}\\
\quad Off the shelf & 0.284 & 0.280 & 0.286 & 0.288 & 0.297 & 0.301 & 0.238 & 0.272 & 0.212 & 0.238 & 0.267 & 0.296 & 0.255 & 0.309 & 0.245 & 0.240 & 0.285 & 0.294 & 0.272 \\
\quad Single-order & 0.252 & 0.246 & 0.232 & 0.256 & 0.260 & 0.265 & 0.266 & 0.294 & 0.284 & 0.281 & 0.236 & 0.284 & 0.328 & 0.277 & 0.268 & 0.268 & 0.283 & 0.277 & 0.270 \\
\quad Order-avg. & 0.161 & 0.159 & 0.137 & 0.167 & 0.168 & 0.147 & 0.164 & 0.199 & 0.179 & 0.209 & 0.126 & 0.219 & 0.261 & 0.192 & 0.175 & 0.188 & 0.179 & 0.167 & 0.178 \\
\quad OC-SFT & 0.079 & 0.085 & 0.062 & 0.077 & 0.081 & 0.075 & 0.078 & 0.099 & 0.125 & 0.094 & 0.060 & 0.100 & 0.095 & 0.104 & 0.076 & 0.096 & 0.086 & 0.082 & 0.086 \\
\quad OC-SFT (avg. labels) & 0.075 & 0.076 & 0.065 & 0.079 & 0.082 & 0.065 & 0.070 & 0.096 & 0.130 & 0.099 & 0.061 & 0.110 & 0.118 & 0.104 & 0.083 & 0.102 & 0.073 & 0.071 & 0.087 \\
\midrule
\multicolumn{20}{l}{\textbf{Qwen3-32B}}\\
\quad Off the shelf & 0.315 & 0.305 & 0.318 & 0.317 & 0.322 & 0.340 & 0.260 & 0.295 & 0.234 & 0.249 & 0.330 & 0.312 & 0.252 & 0.330 & 0.264 & 0.258 & 0.310 & 0.324 & 0.296 \\
\quad Single-order & 0.248 & 0.228 & 0.229 & 0.242 & 0.241 & 0.285 & 0.234 & 0.318 & 0.217 & 0.249 & 0.249 & 0.277 & 0.290 & 0.253 & 0.235 & 0.265 & 0.258 & 0.265 & 0.255 \\
\quad Order-avg. & 0.135 & 0.129 & 0.123 & 0.139 & 0.140 & 0.143 & 0.112 & 0.163 & 0.113 & 0.129 & 0.117 & 0.171 & 0.148 & 0.161 & 0.120 & 0.135 & 0.143 & 0.142 & 0.137 \\
\quad OC-SFT & 0.083 & 0.085 & 0.073 & 0.086 & 0.089 & 0.082 & 0.064 & 0.080 & 0.070 & 0.079 & 0.066 & 0.101 & 0.095 & 0.108 & 0.076 & 0.086 & 0.082 & 0.080 & 0.083 \\
\quad OC-SFT (avg. labels) & 0.063 & 0.066 & 0.059 & 0.068 & 0.071 & 0.059 & 0.054 & 0.073 & 0.069 & 0.069 & 0.053 & 0.090 & 0.075 & 0.092 & 0.060 & 0.074 & 0.062 & 0.059 & 0.068 \\
\midrule
\multicolumn{20}{l}{\textbf{Gemma-E2B}}\\
\quad Off the shelf & 0.301 & 0.306 & 0.298 & 0.308 & 0.311 & 0.269 & 0.268 & 0.249 & 0.279 & 0.248 & 0.273 & 0.301 & 0.276 & 0.324 & 0.263 & 0.237 & 0.335 & 0.332 & 0.288 \\
\quad Single-order & 0.192 & 0.203 & 0.187 & 0.211 & 0.218 & 0.177 & 0.196 & 0.203 & 0.204 & 0.199 & 0.177 & 0.232 & 0.247 & 0.272 & 0.199 & 0.212 & 0.234 & 0.227 & 0.211 \\
\quad Order-avg. & 0.116 & 0.127 & 0.111 & 0.131 & 0.134 & 0.104 & 0.111 & 0.127 & 0.113 & 0.119 & 0.102 & 0.154 & 0.134 & 0.167 & 0.115 & 0.123 & 0.143 & 0.136 & 0.126 \\
\quad OC-SFT & 0.070 & 0.081 & 0.067 & 0.078 & 0.081 & 0.060 & 0.060 & 0.068 & 0.060 & 0.072 & 0.064 & 0.095 & 0.073 & 0.106 & 0.069 & 0.075 & 0.080 & 0.077 & 0.074 \\
\quad OC-SFT (avg. labels) & 0.064 & 0.075 & 0.060 & 0.072 & 0.074 & 0.057 & 0.059 & 0.062 & 0.054 & 0.068 & 0.057 & 0.088 & 0.066 & 0.095 & 0.063 & 0.070 & 0.073 & 0.071 & 0.068 \\
\midrule
\multicolumn{20}{l}{\textbf{Gemma-E4B}}\\
\quad Off the shelf & 0.306 & 0.297 & 0.302 & 0.301 & 0.305 & 0.265 & 0.220 & 0.236 & 0.220 & 0.224 & 0.275 & 0.276 & 0.220 & 0.312 & 0.235 & 0.202 & 0.277 & 0.272 & 0.264 \\
\quad Single-order & 0.185 & 0.179 & 0.175 & 0.197 & 0.198 & 0.150 & 0.141 & 0.169 & 0.158 & 0.169 & 0.149 & 0.200 & 0.186 & 0.213 & 0.144 & 0.138 & 0.208 & 0.199 & 0.175 \\
\quad Order-avg. & 0.115 & 0.123 & 0.111 & 0.130 & 0.134 & 0.096 & 0.094 & 0.118 & 0.110 & 0.113 & 0.092 & 0.143 & 0.126 & 0.152 & 0.100 & 0.100 & 0.132 & 0.122 & 0.117 \\
\quad OC-SFT & 0.087 & 0.097 & 0.081 & 0.095 & 0.099 & 0.070 & 0.071 & 0.094 & 0.097 & 0.089 & 0.065 & 0.117 & 0.112 & 0.118 & 0.080 & 0.092 & 0.096 & 0.090 & 0.092 \\
\quad OC-SFT (avg. labels) & 0.072 & 0.078 & 0.068 & 0.077 & 0.080 & 0.060 & 0.057 & 0.076 & 0.056 & 0.069 & 0.056 & 0.094 & 0.077 & 0.093 & 0.063 & 0.069 & 0.075 & 0.068 & 0.072 \\
\midrule
\multicolumn{20}{l}{\textbf{Gemma-31B}}\\
\quad Off the shelf & 0.284 & 0.273 & 0.294 & 0.270 & 0.274 & 0.298 & 0.206 & 0.246 & 0.225 & 0.233 & 0.273 & 0.280 & 0.243 & 0.297 & 0.243 & 0.227 & 0.247 & 0.261 & 0.260 \\
\quad Single-order & 0.222 & 0.200 & 0.210 & 0.216 & 0.220 & 0.226 & 0.178 & 0.236 & 0.192 & 0.209 & 0.189 & 0.247 & 0.249 & 0.246 & 0.193 & 0.208 & 0.220 & 0.221 & 0.216 \\
\quad Order-avg. & 0.148 & 0.142 & 0.130 & 0.149 & 0.154 & 0.137 & 0.135 & 0.179 & 0.131 & 0.156 & 0.108 & 0.191 & 0.211 & 0.202 & 0.148 & 0.183 & 0.141 & 0.134 & 0.154 \\
\quad OC-SFT & 0.112 & 0.109 & 0.095 & 0.110 & 0.116 & 0.093 & 0.108 & 0.146 & 0.173 & 0.123 & 0.079 & 0.152 & 0.173 & 0.150 & 0.114 & 0.146 & 0.111 & 0.102 & 0.123 \\
\quad OC-SFT (avg. labels) & 0.087 & 0.087 & 0.077 & 0.090 & 0.095 & 0.071 & 0.070 & 0.094 & 0.071 & 0.092 & 0.063 & 0.121 & 0.113 & 0.129 & 0.084 & 0.098 & 0.085 & 0.078 & 0.089 \\
\midrule
\multicolumn{20}{l}{\textbf{Granite-3B}}\\
\quad Off the shelf & 0.291 & 0.285 & 0.290 & 0.300 & 0.299 & 0.264 & 0.301 & 0.272 & 0.277 & 0.268 & 0.267 & 0.292 & 0.314 & 0.296 & 0.262 & 0.272 & 0.303 & 0.306 & 0.287 \\
\quad Single-order & 0.146 & 0.141 & 0.145 & 0.158 & 0.157 & 0.133 & 0.138 & 0.164 & 0.173 & 0.144 & 0.137 & 0.160 & 0.161 & 0.162 & 0.121 & 0.130 & 0.185 & 0.185 & 0.152 \\
\quad Order-avg. & 0.098 & 0.100 & 0.092 & 0.106 & 0.108 & 0.086 & 0.095 & 0.116 & 0.138 & 0.106 & 0.086 & 0.116 & 0.123 & 0.127 & 0.091 & 0.103 & 0.114 & 0.117 & 0.107 \\
\quad OC-SFT & 0.064 & 0.068 & 0.061 & 0.071 & 0.076 & 0.058 & 0.061 & 0.080 & 0.067 & 0.074 & 0.060 & 0.081 & 0.083 & 0.098 & 0.061 & 0.072 & 0.080 & 0.078 & 0.072 \\
\quad OC-SFT (avg. labels) & 0.061 & 0.062 & 0.054 & 0.063 & 0.065 & 0.055 & 0.062 & 0.075 & 0.073 & 0.076 & 0.054 & 0.076 & 0.086 & 0.089 & 0.061 & 0.073 & 0.064 & 0.065 & 0.067 \\
\midrule
\multicolumn{20}{l}{\textbf{Granite-8B}}\\
\quad Off the shelf & 0.193 & 0.187 & 0.286 & 0.282 & 0.287 & 0.218 & 0.248 & 0.243 & 0.110 & 0.225 & 0.257 & 0.274 & 0.252 & 0.288 & 0.228 & 0.220 & 0.295 & 0.293 & 0.244 \\
\quad Single-order & 0.147 & 0.146 & 0.139 & 0.148 & 0.154 & 0.150 & 0.142 & 0.175 & 0.194 & 0.169 & 0.135 & 0.182 & 0.195 & 0.182 & 0.143 & 0.163 & 0.186 & 0.186 & 0.163 \\
\quad Order-avg. & 0.096 & 0.097 & 0.089 & 0.101 & 0.103 & 0.084 & 0.083 & 0.109 & 0.139 & 0.103 & 0.082 & 0.122 & 0.117 & 0.124 & 0.088 & 0.099 & 0.109 & 0.108 & 0.103 \\
\quad OC-SFT & 0.058 & 0.063 & 0.055 & 0.064 & 0.068 & 0.054 & 0.053 & 0.069 & 0.085 & 0.067 & 0.050 & 0.082 & 0.075 & 0.085 & 0.059 & 0.075 & 0.067 & 0.065 & 0.066 \\
\quad OC-SFT (avg. labels) & 0.054 & 0.058 & 0.052 & 0.059 & 0.062 & 0.049 & 0.048 & 0.063 & 0.068 & 0.061 & 0.048 & 0.077 & 0.067 & 0.081 & 0.053 & 0.064 & 0.059 & 0.059 & 0.060 \\
\midrule
\multicolumn{20}{l}{\textbf{Granite-30B}}\\
\quad Off the shelf & 0.274 & 0.271 & 0.273 & 0.276 & 0.283 & 0.276 & 0.248 & 0.249 & 0.235 & 0.236 & 0.270 & 0.288 & 0.271 & 0.300 & 0.239 & 0.241 & 0.282 & 0.289 & 0.267 \\
\quad Single-order & 0.210 & 0.195 & 0.189 & 0.194 & 0.191 & 0.225 & 0.189 & 0.235 & 0.169 & 0.212 & 0.195 & 0.237 & 0.253 & 0.202 & 0.194 & 0.207 & 0.210 & 0.219 & 0.207 \\
\quad Order-avg. & 0.128 & 0.125 & 0.110 & 0.128 & 0.128 & 0.115 & 0.112 & 0.153 & 0.120 & 0.157 & 0.100 & 0.173 & 0.220 & 0.155 & 0.141 & 0.153 & 0.130 & 0.127 & 0.137 \\
\quad OC-SFT & 0.076 & 0.083 & 0.065 & 0.076 & 0.080 & 0.065 & 0.062 & 0.091 & 0.073 & 0.087 & 0.058 & 0.103 & 0.115 & 0.103 & 0.076 & 0.096 & 0.072 & 0.069 & 0.081 \\
\quad OC-SFT (avg. labels) & 0.066 & 0.071 & 0.060 & 0.068 & 0.069 & 0.057 & 0.056 & 0.081 & 0.059 & 0.085 & 0.054 & 0.096 & 0.101 & 0.094 & 0.069 & 0.086 & 0.061 & 0.060 & 0.072 \\
\midrule
\multicolumn{20}{l}{\textbf{Gemma-4 26B-A4B (sparse MoE)}}\\
\quad Off the shelf & 0.267 & 0.251 & 0.277 & 0.254 & 0.263 & 0.288 & 0.193 & 0.233 & 0.197 & 0.207 & 0.256 & 0.268 & 0.225 & 0.279 & 0.219 & 0.201 & 0.249 & 0.268 & 0.244 \\
\quad Single-order & 0.191 & 0.176 & 0.171 & 0.183 & 0.190 & 0.193 & 0.157 & 0.215 & 0.158 & 0.184 & 0.146 & 0.228 & 0.229 & 0.230 & 0.170 & 0.185 & 0.186 & 0.186 & 0.188 \\
\quad Order-avg. & 0.134 & 0.131 & 0.117 & 0.134 & 0.139 & 0.134 & 0.108 & 0.152 & 0.120 & 0.129 & 0.097 & 0.176 & 0.156 & 0.171 & 0.123 & 0.134 & 0.132 & 0.128 & 0.134 \\
\quad OC-SFT ($\lambda=3$) & 0.113 & 0.114 & 0.096 & 0.112 & 0.120 & 0.091 & 0.090 & 0.129 & 0.098 & 0.120 & 0.076 & 0.161 & 0.154 & 0.154 & 0.109 & 0.125 & 0.106 & 0.101 & 0.115 \\
\quad OC-SFT (avg. labels, $\lambda=5$) & 0.088 & 0.087 & 0.076 & 0.087 & 0.095 & 0.073 & 0.066 & 0.092 & 0.066 & 0.084 & 0.060 & 0.120 & 0.098 & 0.114 & 0.076 & 0.085 & 0.081 & 0.076 & 0.085 \\
\end{longtable}
\end{landscape}

\FloatBarrier

\end{document}